\documentclass[11pt,a4paper]{article}
\pdfoutput=1

\usepackage[margin=1.0in]{geometry}
\usepackage{times}
\usepackage{microtype}
\usepackage{graphicx}
\usepackage{booktabs}
\usepackage{amsmath,amssymb,amsfonts,bm}
\usepackage{csquotes}
\usepackage{xcolor}
\usepackage{hyperref}
\usepackage{caption}
\usepackage{subcaption}
\usepackage{fontawesome5}

\usepackage{tikz}
\usetikzlibrary{arrows.meta, positioning, shapes.geometric, calc, fit, backgrounds}
\definecolor{figblue}{HTML}{1E293B}
\definecolor{figborderblue}{HTML}{3B82F6}
\definecolor{figgreen}{HTML}{064E3B}
\definecolor{figbordergreen}{HTML}{10B981}
\definecolor{figred}{HTML}{450A0A}
\definecolor{figborderred}{HTML}{EF4444}
\definecolor{figpurple}{HTML}{2E1065}
\definecolor{figborderpurple}{HTML}{8B5CF6}
\definecolor{figyellow}{HTML}{451A03}
\definecolor{figborderyellow}{HTML}{F59E0B}
\definecolor{figindigo}{HTML}{4B0082}
\definecolor{figamber}{HTML}{FFBF00}

\pgfdeclarelayer{background}
\pgfdeclarelayer{foreground}
\pgfsetlayers{background,main,foreground}

\hypersetup{
    colorlinks=true,
    linkcolor=blue!80!black,
    citecolor=blue!80!black,
    urlcolor=blue!80!black
}

\usepackage[backend=biber,
            style=numeric,
            backref=true,
            isbn=false,
            sortcites,
            maxbibnames=6,
            minbibnames=1]{biblatex}
\title{\textbf{\LARGE{Asuka-FM: A System-Architectural Co-Design Framework for Efficient Text-to-Image Pretraining on Consumer Hardware}}}

\author{
    \textbf{Víctor Rodrigo González}$^{1}$ \quad
    \textbf{Pablo M. Olmos}$^{1}$ \quad
    \textbf{Ignacio Peis}$^{2,3}$ \quad
    \textbf{Alejandro Lancho}$^{1}$ \\[0.45em]
    {\footnotesize $^1$Department of Signal Theory and Communications, Universidad Carlos III de Madrid, Spain} \\[-0.1em]
    {\footnotesize $^2$Department of Applied Mathematics and Computer Science, Technical University of Denmark (DTU), Denmark} \\[-0.1em]
    {\footnotesize $^3$Math Institute, Universidad de Granada, Spain} \\[0.25em]
    {\footnotesize\ttfamily victorro@pa.uc3m.es \quad pamartin@ing.uc3m.es \quad ipeis@ugr.es \quad alancho@ing.uc3m.es}
}
\date{}

\begin{document}

\maketitle
\vspace{-2.6em}
\begin{center}
    \small
    \href{https://deeplearningcafe.github.io/projects/asuka-fm}{\textbf{\color{figborderblue}\faGlobe\ \underline{Project Page}}} \quad\quad
    \href{https://github.com/deeplearningcafe/asuka-fm}{\textbf{\color{figblue}\faGithub\ \underline{Code}}} \quad\quad
    \href{https://huggingface.co/aipracticecafe/asuka-fm-air}{\textbf{\color{figborderyellow}\faCube\ \underline{Model}}} \quad\quad
    \href{https://huggingface.co/datasets/aipracticecafe/i1-dataset}{\textbf{\color{figbordergreen}\faDatabase\ \underline{Data}}}
\end{center}
\vspace{-0.4em}

\begin{center}
    \captionsetup{type=figure, font=footnotesize, labelfont=bf, skip=2pt}
    \begin{minipage}[c]{0.53\textwidth}
        \centering
        \includegraphics[width=\linewidth]{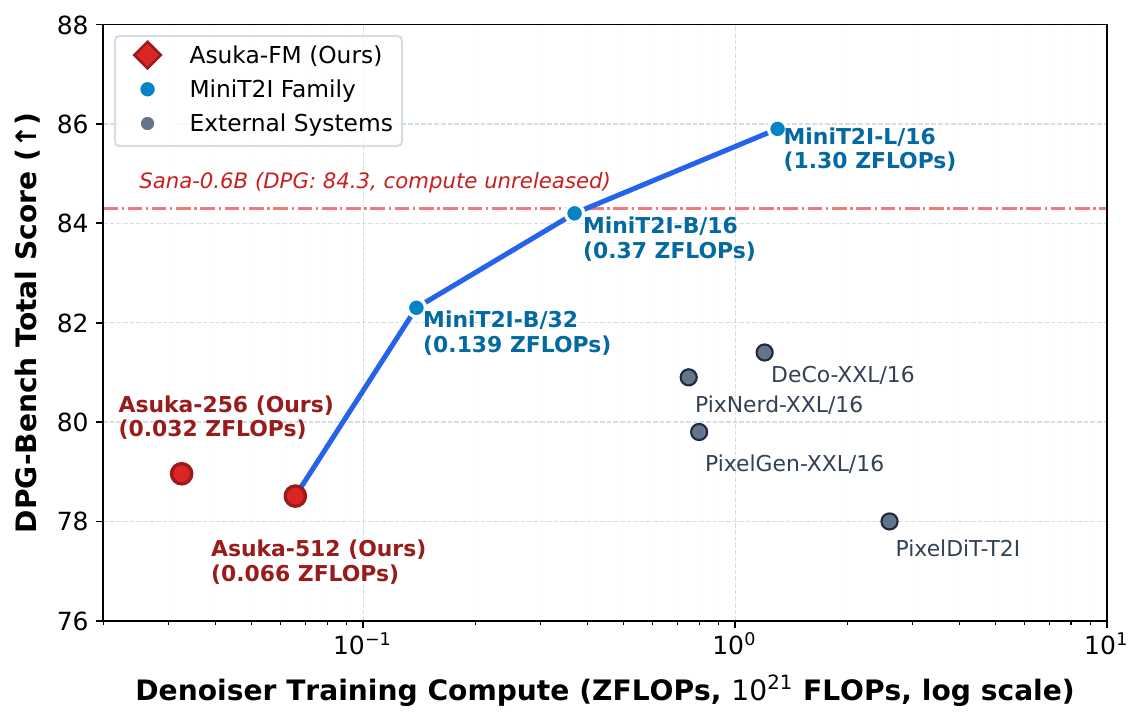}
        \vspace{-7pt}
        \centerline{\footnotesize (a) Compute Pareto Frontier: Training FLOPs vs. DPG-Bench}
    \end{minipage}\hfill
    \begin{minipage}[c]{0.45\textwidth}
        \centering
        \begin{minipage}{\linewidth}
            \centering
            \includegraphics[width=0.48\linewidth]{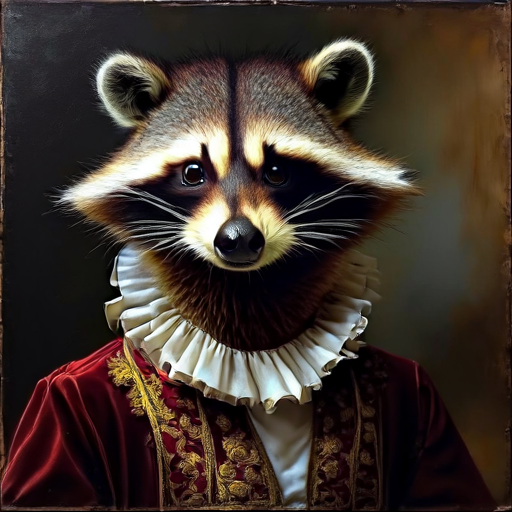}\hfill
            \includegraphics[width=0.48\linewidth]{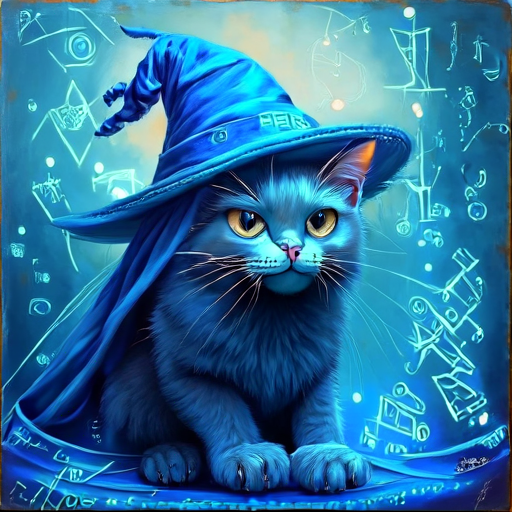}
            \vspace{1pt}
            \centerline{\footnotesize (b) Asuka-FM Samples}
        \end{minipage}
        
        \vspace{2pt}
        
        \begin{minipage}{\linewidth}
            \centering
            \setlength{\tabcolsep}{3.5pt}
            \renewcommand{\arraystretch}{1.02}
            \resizebox{\linewidth}{!}{%
            \begin{tabular}{lcccc}
            \toprule
            \textbf{Model} & \textbf{Params} & \textbf{Hardware Budget} & \textbf{GenEval $\uparrow$} & \textbf{DPG $\uparrow$} \\
            \midrule
            \multicolumn{5}{l}{\textit{\textbf{Industrial Foundation Models}}} \\
            SD3 Medium~\cite{esser2024scaling} & 2.0B & Massive Cluster & 0.620 & 84.08 \\
            \midrule
            \multicolumn{5}{l}{\textit{\textbf{Accessible Academic Baselines}}} \\
            MiniT2I-L/16~\cite{minit2i2026} & 0.91B & 8$\times$ H100 (27 Days) & 0.883$^{\dagger}$ & 84.80 \\
            \midrule
            \multicolumn{5}{l}{\textit{\textbf{Asuka-FM (Consumer GPU, Open Data)}}} \\
            Asuka-512 (Ours) & 0.21B & 1$\times$ RTX 4090 (81h) & 0.609 & 78.51 \\
            \bottomrule
            \end{tabular}%
            }
            \vspace{1pt}
            \centerline{\footnotesize (c) Benchmark Summary}
        \end{minipage}
    \end{minipage}
    \vspace{2pt}
    \captionof{figure}{\textbf{Asuka-FM framework overview.} \textbf{(a)} Denoiser training compute vs. compositional alignment (DPG-Bench). \textbf{(b)} High-fidelity visual synthesis from our 206M dual-stream backbone. \textbf{(c)} Representative performance summary across industrial models, academic baselines, and Asuka-FM ($^\dagger$denotes finetuning on \texttt{BLIP3o-60K}).}
    \label{fig:teaser_overview}
\end{center}
\vspace{-0.2em}

\begin{abstract}
Frontier text-to-image (T2I) foundation models predominantly rely on multi-billion parameter Diffusion Transformers (DiTs) pretrained across massive compute clusters, creating an accessibility barrier for academic laboratories. When training DiTs from scratch on single consumer GPUs (e.g., an NVIDIA RTX 4090 with 24\,GB VRAM), practitioners encounter three primary bottlenecks: quadratic attention sequence costs, Just-In-Time (JIT) compilation graph breaks induced by variable aspect-ratio bucketing, and an autoencoder throughput wall where frozen Variational Autoencoder (VAE) operations dominate step latency. In this work, we introduce \textbf{Asuka-FM}, an end-to-end system and architectural framework demonstrating that competitive dual-stream DiTs ($\sim$206M parameters) can be pretrained from scratch on a single consumer GPU in $\sim$160 GPU hours for under \$55. Asuka-FM integrates four core designs: (i) an AdaLN-free dual-stream DiT backbone reducing parameter count; (ii) structured sequence sparsification via Sparse-Dense Residual Fusion (SPRINT) to cut attention FLOPs by over 50\%; (iii) continuous aspect-normalized Rotary Position Embeddings (RoPE) coupled with shifted square crop training, guaranteeing static tensor shapes while unlocking zero-shot virtual camera control; and (iv) a pixel adaptation phase utilizing a lightweight (10M parameter) Patch Detailer Head operating under velocity regression ($v$-prediction). Evaluated on open-domain benchmarks without benchmark-specific fine-tuning, Asuka-FM achieves GenEval 0.588 / DPG 78.96 scores at 256px and GenEval 0.609 / DPG 78.51 scores at 512px, setting a new Pareto-optimal efficiency standard for democratic foundation modeling.
\end{abstract}

\clearpage
\begin{figure}[p!]
    \centering
    \captionsetup{font=small, labelfont=bf, skip=6pt}
    \includegraphics[width=\textwidth, height=0.86\textheight, keepaspectratio]{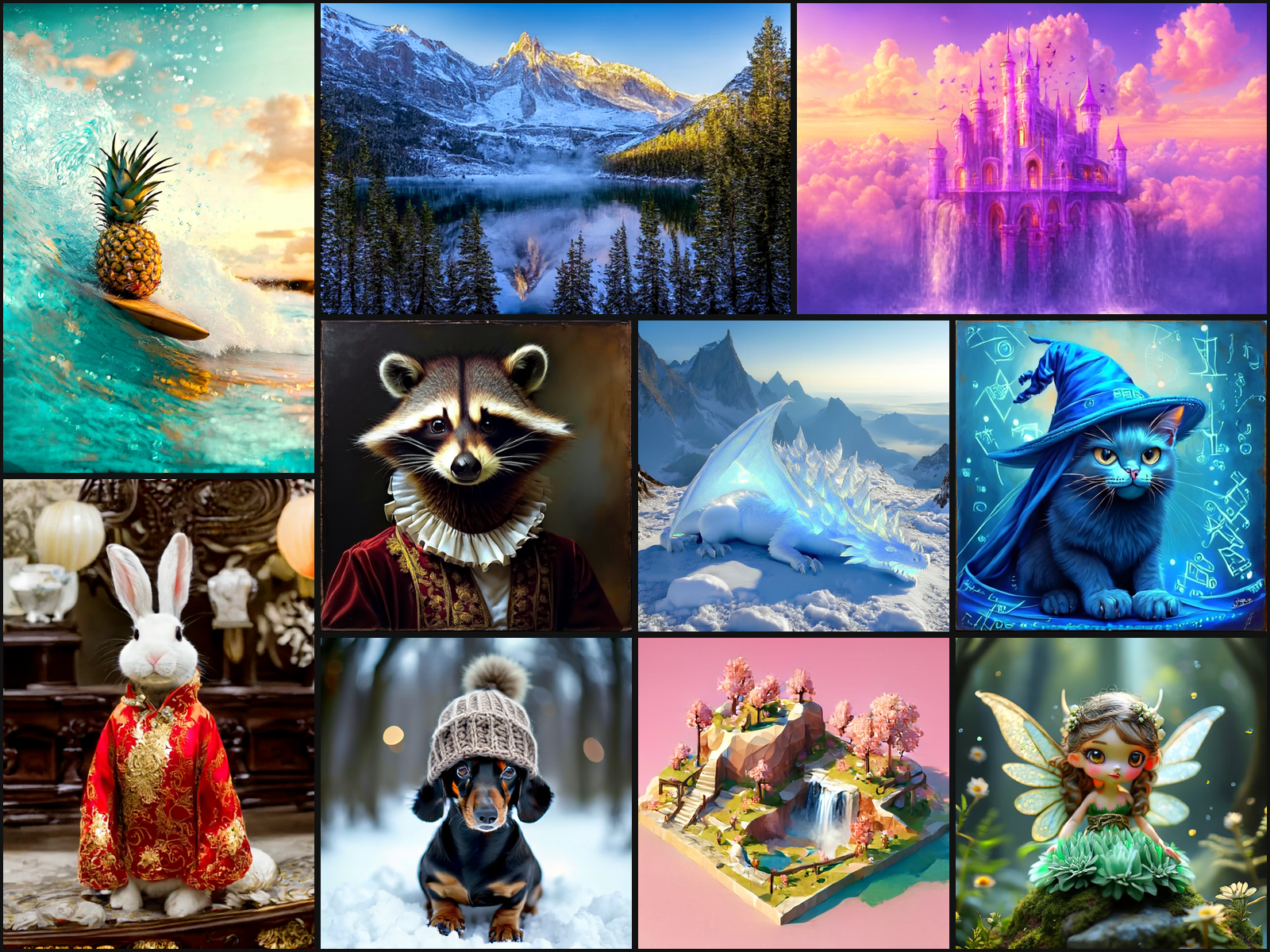}
    \caption{\textbf{High-fidelity zero-shot visual synthesis from scratch on a single consumer GPU.} Curated open-domain samples synthesized by Asuka-FM (206M backbone) following two-stage pretraining ($256^2 \to 512^2$) on a single NVIDIA RTX 4090 GPU in 149.0 hours (\$49.17).}
    \label{fig:teaser_showcase}
\end{figure}
\clearpage

\section{Introduction}
\label{sec:introduction}

The emergence of Diffusion Transformers (DiTs)~\cite{peebles2023scalable} trained via flow matching~\cite{lipman2023flow} has established a new state of the art in high-fidelity text-to-image (T2I) synthesis. However, state-of-the-art models, such as FLUX.1 (12B)~\cite{flux2024}, SD3 (8B)~\cite{esser2024scaling}, and Qwen-Image (20B)~\cite{wu2025qwenimage}, have scaled parameter counts and training datasets to regimes that require industrial-scale data centers. This trajectory has widened the accessibility gap, preventing independent researchers and academic groups with limited resources from training generative models from scratch.

When attempting to pretrain generative diffusion models under compute-constrained environments (e.g., a single workstation with an NVIDIA RTX 4090 with 24\,GB GDDR6X), the standard pretraining paradigm encounters three fundamental systems bottlenecks:
\begin{itemize}
    \item (i)~\textbf{Quadratic Attention Sequence Redundancy:} Standard homogeneous DiTs allocate identical full-token sequence lengths across every transformer block. However, recent representational analyses~\cite{sprint2026} reveal that early layers primarily capture local high-frequency textures, whereas deep middle blocks model global macro-semantics. Processing full-length sequences across deep layers wastes limited floating-point operations (FLOPs) on redundant spatial details, slowing convergence.
    \item (ii)~\textbf{Just-In-Time (JIT) Compilation Graph Breaks vs. Aspect-Ratio Bucketing:} Real-world images feature diverse aspect ratios. Conventional pipelines handle this via dynamic aspect-ratio bucketing, grouping images of varying shapes into specialized mini-batches. Under modern tracing and JIT compiler runtimes (e.g., \texttt{torch.compile} with TorchInductor), changing spatial dimensions dynamically triggers continuous kernel recompilations, invalidates static CUDA graphs, and severely degrades actual Model FLOPs Utilization ($\text{MFU}_{\text{act}}$).
    \item (iii)~\textbf{The Autoencoder Throughput Wall:} In conventional Latent Diffusion Models (LDMs)~\cite{rombach2022high}, an autoencoder (such as the FLUX.2 VAE~\cite{flux2report}) compresses images into a regularized latent space. At high resolutions ($1024 \times 1024$), storing pre-computed 32-channel float16 latents for hundreds of thousands of images requires multiple terabytes of high-speed storage, exceeding consumer NVMe capacities. Consequently, latents must be computed on-the-fly. On consumer silicon during the training stage, the frozen VAE forward pass consumes $\approx 60\%$ of total wall-time, whereas a compiled 200M DiT forward-backward pass requires only $\approx 35\%$ (Figure~\ref{fig:vae_wall_breakdown_appendix} in Appendix~\ref{sec:appendix_full_tables}). Furthermore, at inference, VAE decoding accounts for over 70\% of wall-clock latency under few-step regimes, creating a computational bottleneck for downstream reinforcement learning (e.g., Flow-GRPO~\cite{liu2025flowgrpo}) and distillation.
\end{itemize}

\begin{figure*}[t]
\centering
\resizebox{\textwidth}{!}{%
\begin{tikzpicture}[
    every node/.style={font=\sffamily\small},
    stage/.style={
        rectangle, draw=#1!80!black, fill=#1!12, rounded corners=4pt,
        line width=1.1pt, align=center, inner sep=4pt, minimum height=1.35cm
    },
    ckpt/.style={
        rectangle, draw=orange!80!black, fill=orange!20, rounded corners=4pt,
        line width=1.2pt, align=center, inner sep=4pt, minimum height=1.35cm
    },
    groupbox/.style={
        draw=gray!50, dashed, rounded corners=6pt, fill=gray!3, line width=0.9pt
    },
    headertab/.style={
        rectangle, draw=none, rounded corners=2pt, font=\footnotesize\bfseries,
        inner sep=3pt
    },
    flowarr/.style={
        -{Stealth[length=5pt, width=3.5pt]}, draw=gray!70, line width=1.1pt
    },
    interarr/.style={
        -{Stealth[length=6pt, width=4pt]}, draw=blue!70!black, line width=1.3pt
    }
]
    \def\iconChart{
        \begin{tikzpicture}[scale=0.8, baseline=-2pt]
            \fill[blue!70!black] (0, 0) rectangle (0.08, 0.16);
            \fill[blue!70!black] (0.12, 0) rectangle (0.20, 0.28);
            \fill[blue!70!black] (0.24, 0) rectangle (0.32, 0.40);
        \end{tikzpicture}
    }
    \def\iconGrid{
        \begin{tikzpicture}[scale=0.75, baseline=-2pt]
            \draw[green!60!black, line width=0.8pt, step=0.14] 
                (0, 0) grid (0.42, 0.42);
        \end{tikzpicture}
    }
    \def\iconDecay{
        \begin{tikzpicture}[scale=0.8, baseline=-2pt]
            \draw[purple!80!black, line width=1.0pt] 
                (0, 0.32) to[out=0, in=120] (0.2, 0.2) to[out=-60, in=180] (0.4, 0.04);
        \end{tikzpicture}
    }
    \def\iconCkpt{
        \begin{tikzpicture}[scale=0.75, baseline=-2pt]
            \draw[orange!90!black, fill=orange!40, line width=0.8pt] 
                (0.2, 0.38) -- (0.38, 0.28) -- (0.38, 0.10) -- 
                (0.2, 0.0) -- (0.02, 0.10) -- (0.02, 0.28) -- cycle;
            \draw[orange!90!black, line width=0.6pt] 
                (0.2, 0.38) -- (0.2, 0.19) -- (0.38, 0.10);
            \draw[orange!90!black, line width=0.6pt] 
                (0.2, 0.19) -- (0.02, 0.10);
        \end{tikzpicture}
    }
    \def\iconFilter{
        \begin{tikzpicture}[scale=0.75, baseline=-2pt]
            \draw[teal!80!black, fill=teal!20, line width=0.8pt] 
                (0, 0) rectangle (0.38, 0.38);
            \draw[teal!80!black, line width=0.6pt] (0.19, 0) -- (0.19, 0.38);
            \draw[teal!80!black, line width=0.6pt] (0, 0.19) -- (0.38, 0.19);
        \end{tikzpicture}
    }

    \node[stage=blue, minimum width=3.3cm] (s1a) at (0.0, 6.0) {
        \iconChart\\[-1pt]
        \textbf{1A: SPRINT Sparse}\\
        \scriptsize $r=0.75$ Drop $\cdot$ 75 Ep ($\sim$47.0h)
    };
    \node[stage=green!80!black, minimum width=3.3cm, right=0.65cm of s1a] (s1b) {
        \iconGrid\\[-1pt]
        \textbf{1B: Dense Refine}\\
        \scriptsize $r=0.00$ Full $\cdot$ 20 Ep ($\sim$21.0h)
    };
    \node[stage=purple, minimum width=3.3cm, right=0.65cm of s1b] (s1c) {
        \iconDecay\\[-1pt]
        \textbf{1C: WSD Cooldown}\\
        \scriptsize $\eta \to 10^{-5}$ $\cdot$ Linear Decay
    };
    \node[ckpt, minimum width=3.3cm, right=0.65cm of s1c] (ckpt1) {
        \iconCkpt\\[-1pt]
        \textbf{Asuka-256 Base}\\
        \scriptsize Concept \& RoPE Anchored (\$22.44)
    };
    \draw[flowarr] (s1a) -- (s1b);
    \draw[flowarr] (s1b) -- (s1c);
    \draw[flowarr] (s1c) -- (ckpt1);

    \node[stage=blue, minimum width=3.3cm] (s2a) at (0.0, 3.0) {
        \iconChart\\[-1pt]
        \textbf{2A: High-Res Sparse}\\
        \scriptsize $r=0.75$ Drop $\cdot$ 25 Ep ($\sim$40.0h)
    };
    \node[stage=green!80!black, minimum width=3.3cm, right=0.65cm of s2a] (s2b) {
        \iconGrid\\[-1pt]
        \textbf{2B: Dense Refine}\\
        \scriptsize $r=0.00$ Full $\cdot$ 15 Ep ($\sim$41.0h)
    };
    \node[stage=purple, minimum width=3.3cm, right=0.65cm of s2b] (s2c) {
        \iconDecay\\[-1pt]
        \textbf{2C: WSD Cooldown}\\
        \scriptsize $\eta \to 10^{-5}$ $\cdot$ Linear Decay
    };
    \node[ckpt, minimum width=3.3cm, right=0.65cm of s2c] (ckpt2) {
        \iconCkpt\\[-1pt]
        \textbf{Asuka-512 High-Res}\\
        \scriptsize Textures \& Macro Layout (\$26.73)
    };
    \draw[flowarr] (s2a) -- (s2b);
    \draw[flowarr] (s2b) -- (s2c);
    \draw[flowarr] (s2c) -- (ckpt2);

    \draw[interarr] (ckpt1.south) 
        |- ($(s2a.north) + (-0.8, 0.42)$) 
        -- ($(s2a.north) + (-0.8, 0)$);
    \node[font=\scriptsize\sffamily, text=blue!85!black, fill=white, inner sep=2pt, rounded corners=2pt] 
        at ($(s2a.north) + (2.8, 0.42)$) 
        {$2\times$ Canvas Scaling ($256^2 \to 512^2$)};

    \node[stage=teal, minimum width=3.3cm] (p1) at (0.0, 0.0) {
        \iconFilter\\[-1pt]
        \textbf{Stage 1: PDH Warmup}\\
        \scriptsize Frozen DiT $\cdot$ 5 Ep ($\sim$2.3h)
    };
    \node[stage=blue!70!teal, minimum width=3.3cm, right=0.65cm of p1] (p2) {
        \iconChart\\[-1pt]
        \textbf{Stage 2: Joint SPRINT}\\
        \scriptsize Synth Data $\cdot$ $r=0.75 \to 0$ ($\sim$21.4h)
    };
    \node[stage=green!80!black, minimum width=3.3cm, right=0.65cm of p2] (p3) {
        \iconGrid\\[-1pt]
        \textbf{Stage 3: Real Data SFT}\\
        \scriptsize 396k Curated $\cdot$ 13 Ep ($\sim$34.0h)
    };
    \node[ckpt, minimum width=3.3cm, right=0.65cm of p3] (ckpt3) {
        \iconCkpt\\[-1pt]
        \textbf{Final Pixel DiT}\\
        \scriptsize Zero-Decode RGB Engine (\$19.04)
    };
    \draw[flowarr] (p1) -- (p2);
    \draw[flowarr] (p2) -- (p3);
    \draw[flowarr] (p3) -- (ckpt3);

    \draw[interarr] (ckpt2.south) 
        |- ($(p1.north) + (-0.8, 0.42)$) 
        -- ($(p1.north) + (-0.8, 0)$);
    \node[font=\scriptsize\sffamily, text=teal!90!black, fill=white, inner sep=2pt, rounded corners=2pt] 
        at ($(p1.north) + (3.0, 0.42)$) 
        {Latent Foundation Weights Transferred to Pixel Space};

    \begin{pgfonlayer}{background}
        \draw[groupbox] 
            ($(s1a.north west) + (-0.25, 0.50)$) rectangle 
            ($(ckpt1.south east) + (0.25, -0.22)$);

        \draw[groupbox] 
            ($(s2a.north west) + (-0.25, 0.85)$) rectangle 
            ($(ckpt2.south east) + (0.25, -0.22)$);

        \draw[groupbox] 
            ($(p1.north west) + (-0.25, 0.85)$) rectangle 
            ($(ckpt3.south east) + (0.25, -0.22)$);
    \end{pgfonlayer}

    \begin{pgfonlayer}{foreground}
        \node[headertab, text=blue!90!black, fill=blue!10] 
            at ($(s1a.north west) + (3.3, 0.50)$) {
            Stage 1: 256px Base Pretraining ($\approx$68.0h on 1$\times$ RTX 4090)
        };

        \node[headertab, text=green!80!black, fill=green!10] 
            at ($(s2a.north west) + (3.3, 0.85)$) {
            Stage 2: 512px High-Res Adaptation ($\approx$81.0h on 1$\times$ RTX 4090)
        };

        \node[headertab, text=teal!95!black, fill=teal!10] 
            at ($(p1.north west) + (3.4, 0.85)$) {
            Stage 3: Latent-to-Pixel Adaptation ($\approx$57.7h on RTX 4090)
        };
    \end{pgfonlayer}
\end{tikzpicture}%
}
\vspace{-4pt}
\caption{\textbf{Comprehensive multi-stage training roadmap of Asuka-FM.} Pretraining advances through two latent resolution tiers ($256^2 \to 512^2$) coupling 75\% SPRINT sequence sparsification with full-token dense refinement and Warmup-Stable-Decay (WSD). Stage 3 transitions the converged 512px foundation prior to pixel space using a 10M convolutional Patch Detailer Head (PDH) trained under native $v$-prediction across synthetic and real datasets, permanently removing the VAE decode barrier during few-step inference.}
\label{fig:global_training_roadmap}
\end{figure*}
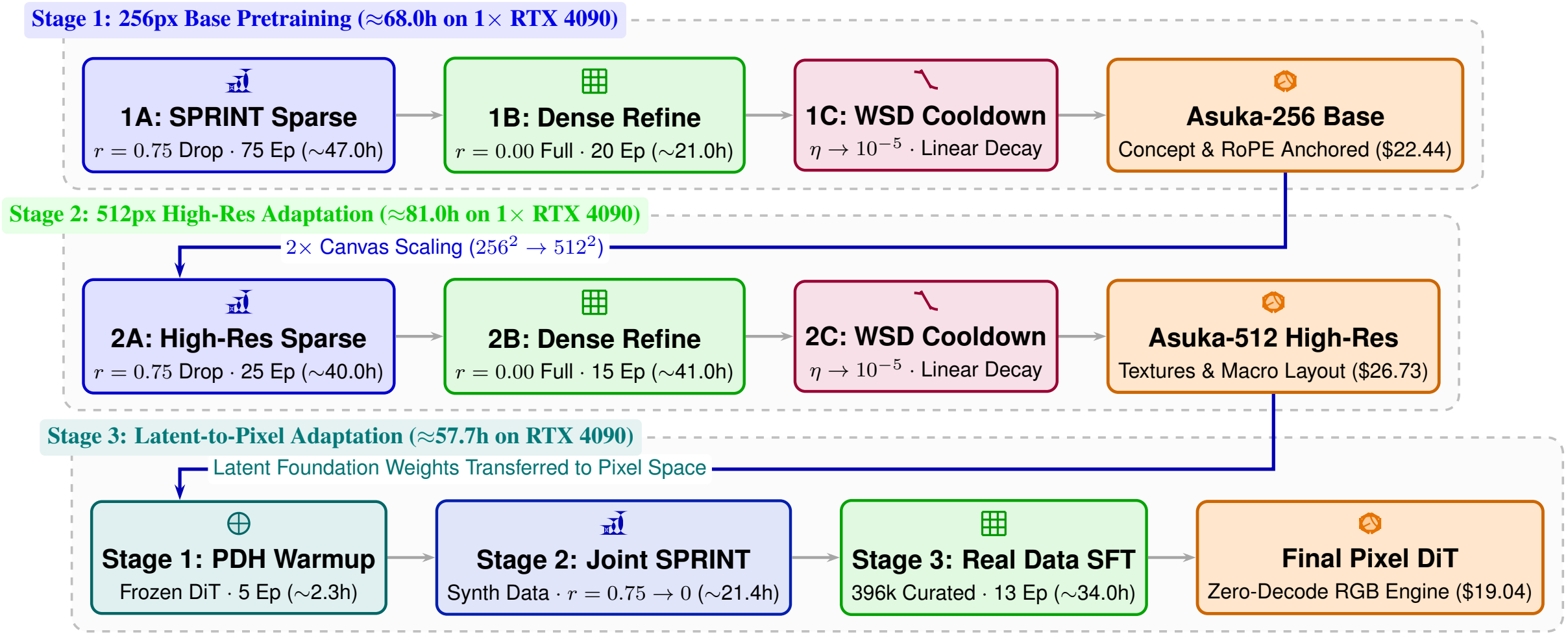

To address these challenges, we introduce \textbf{Asuka-FM}, an end-to-end system and architectural co-design framework proving that a competitive $\sim$206M-parameter Dual-Stream DiT can be pretrained from scratch on a single desktop GPU in $\sim$160 GPU hours for under \$55\footnote{Compute costs are benchmarked against RunPod Community Cloud on-demand rates as of October 2026: \$0.33/GPU-hour for consumer NVIDIA RTX 4090 (24GB GDDR6X) and \$0.40/GPU-hour for datacenter NVIDIA A40 (48GB GDDR6).}. As showcased in Figure~\ref{fig:teaser_showcase}, our open-domain model achieves sharp textural coherence, rich lighting, and diverse multi-concept synthesis (additional domain-specialized anime samples are shown in Figure~\ref{fig:anime_teaser_showcase} in Appendix~\ref{sec:qualitative_gallery}). To achieve this efficiency, we propose a multi-stage training roadmap (Figure~\ref{fig:global_training_roadmap}) coupling structured sequence sparsification, canvas scaling, and latent-to-pixel adaptation. Our primary contributions are:
\begin{itemize}
    \item \textbf{System-Architectural Pretraining Recipe:} We establish an engineering pipeline integrating in-RAM latent caching for low-resolution stages, network streaming for intermediate scales, and FlashAttention-2 kernel fusion within a fully static \texttt{torch.compile} execution graph. This setup sustains an average 75.6\% physical $\text{MFU}_{\text{act}}$ on a single consumer RTX 4090 GPU.
    \item \textbf{High-Efficiency Dual-Stream Architecture:} We couple an AdaLN-free Dual-Stream DiT backbone with three synergistic mechanisms: (i) SPRINT sequence sparsification~\cite{sprint2026}, routing 75\%-sparse token sequences through 9 heavy middle blocks to slash attention FLOPs by over 50\%; (ii) Continuous Aspect-Normalized RoPE and shifted square cropping~\cite{hdm2025,wang2023patch}, supported by a mathematical derivation proving that uncalibrated continuous RoPE undergoes spatial phase collapse, and a full-spectrum frequency calibration that ensures isotropic decay while conferring zero-shot virtual camera control; and (iii) Decoupled Latent-to-Pixel Adaptation, which attaches a lightweight (10M parameter) convolutional Patch Detailer Head (PDH)~\cite{dip2026} under native velocity regression ($v$-prediction) to eliminate the online VAE wall.
    \item \textbf{Empirical Ablations and Profiling with Systems Analysis:} We discover an \textit{Amortization Asymmetry}: while pixel adaptation eliminates the online VAE encoding wall during pretraining, its recurrent convolutional evaluation makes standard multi-step sampling ($N \ge 13$) slower than latent ODE integration, delivering superior efficiency exclusively in few-step regimes ($N \le 8$, yielding up to $3.34\times$ speedups at $N = 1$ step).
    \item \textbf{Open-Domain Validation:} Trained on a balanced 625K multi-source open corpus without benchmark-specific fine-tuning (avoiding BLIP3o-60K overfitting), Asuka-FM achieves a GenEval score of 0.588 and DPG-Bench score of 78.96 at 256px (68h), converging to 0.609 GenEval and 78.51 DPG-Bench at 512px (81h).
\end{itemize}

\section{Related Work}
\label{sec:related_work}
\textbf{Diffusion Transformers and Conditioning Paradigms.} The transition from U-Nets to Diffusion Transformers (DiTs)~\cite{peebles2023scalable} established transformers as the standard generative backbone. Multi-Modal DiTs (MMDiT) in SD3~\cite{esser2024scaling}, FLUX.1~\cite{flux2024} utilize separate weights per modality with shared self-attention, while single-stream models (Lumina-T2X~\cite{gao2024lumina}, Z-Image~\cite{cai2025zimage}) concatenate modalities directly. To eliminate parameter overhead from per-block Adaptive Layer Normalization (AdaLN-Zero), DiT-Air~\cite{chen2025ditair} shared AdaLN parameters globally. While Peebles \& Xie~\cite{peebles2023scalable} initially explored concatenating class tokens in single-stream models, $i1$~\cite{i1arxiv2026} showed that explicit modulation can be entirely omitted when stabilized by Sandwich RMSNorm~\cite{ding2021cogview} and SwiGLU activations~\cite{shazeer2020glu}. Asuka-FM adopts this principle by replacing AdaLN with in-context temporal conditioning, prepending a continuous sinusoidal timestep token directly to the multimodal sequence.

\textbf{Sequence Sparsification and Token Dropping.} Because self-attention complexity scales as $\mathcal{O}(N^2)$ with sequence length $N$, token dropping provides significant acceleration. MaskDiT~\cite{zheng2024maskdit} and MDTv2~\cite{gao2023mdtv2} introduced masked autoencoding objectives, while TREAD~\cite{krause2025tread} bypassed subsets of tokens through inner layers. SPRINT~\cite{sprint2026} demonstrated that structural specialization—combining a dense shallow encoder ($f_\theta$), a 75\%-sparse middle path ($g_\theta$), and a dense decoder ($h_\theta$) fused via linear skips—permits aggressive token dropping without fidelity loss. We extend SPRINT to open-domain multimodal conditioning and evaluate its scaling limits.

\textbf{Looped Transformers: Parameter-Efficient Depth Scaling.} Looped transformers~\cite{dehghani2018universal,geiping2026scaling,chng2026looped} achieve parameter efficiency by repeatedly executing a shared group of middle transformer blocks. In Looped-DiT~\cite{chng2026looped}, cycling through shared blocks keeps weights resident in on-chip SRAM and L2 cache, mitigating memory bandwidth bottlenecks. While looped models compress parameter count across recurrent depth, Asuka-FM focuses on sequence-length compression via SPRINT, providing complementary architectural paths toward efficient pretraining.

\textbf{Micro-Budget Foundation Modeling and Pixel Diffusion.} Micro-budget training has gained significant attention. \textit{Stretching Each Dollar}~\cite{sehwag2025stretching} demonstrated T2I pretraining from scratch for \$1,890 on 8$\times$ H100s using mixture-of-experts and deferred masking. Concurrently, HDM~\cite{hdm2025} trained a 343M model for \$535--620 on 4$\times$ RTX 5090 GPUs using shifted square cropping and aspect-normalized RoPE with learnable frequency parameters. In pixel-space diffusion, early works suffered from slow convergence~\cite{ho2020denoising,chen2025pixelflow}. JiT~\cite{li2026basicsletdenoisinggenerative} and Jiang et al.~\cite{jiang2026empirical} advocated clean data prediction ($x_0$-prediction), but Iris-3B~\cite{cai2026iris3bgoinglatentpixelspace} showed that velocity regression ($v$-prediction) provides equal stability. DiP~\cite{dip2026} introduced the Patch Detailer Head (PDH) to bridge latent efficiency with pixel generation. Asuka-FM synthesizes these insights into an integrated low-cost training pipeline.

\section{Preliminaries}
\label{sec:preliminaries}

\subsection{Continuous Normalizing Flows and Flow Matching}
We formulate image generation through Continuous Normalizing Flows (CNF) and Conditional Flow Matching (CFM)~\cite{lipman2023flow,liu2023flow}. Let $x_0 \sim p_0(x)$ denote an empirical sample from data distribution $p_0(x)$ on a low-dimensional manifold $\mathcal{M} \subset \mathbb{R}^{d_{\text{data}}}$ of ambient dimension $d_{\text{data}}$, and let $x_1 \sim p_1(x) = \mathcal{N}(\mathbf{0}, \mathbf{I})$ denote noise from a standard Gaussian prior with zero mean $\mathbf{0}$ and identity covariance $\mathbf{I}$. Under the optimal-transport linear probability path across continuous time $t \in [0, 1]$, intermediate noisy states $x_t$ are constructed via:
\begin{equation}
    x_t = (1 - t)x_0 + t x_1, \quad t \in [0, 1].
    \label{eq:cfm_interpolation}
\end{equation}
Differentiating Eq.~\eqref{eq:cfm_interpolation} yields the conditional velocity field $u_t(x_t \mid x_0, x_1) = x_1 - x_0$. A neural vector field regressor $v_\theta(x_t, t, c)$, conditioned on prompt context $c$, is trained via the simulation-free objective:
\begin{equation}
    \mathcal{L}_{\text{CFM}}(\theta) = \mathbb{E}_{t, x_0, x_1} \left[ \left\| v_\theta(x_t, t, c) - (x_1 - x_0) \right\|_2^2 \right].
    \label{eq:cfm_loss}
\end{equation}
At inference, novel samples are generated by drawing $x_1 \sim \mathcal{N}(\mathbf{0}, \mathbf{I})$ and integrating the probability flow ODE: $d x_t = v_\theta(x_t, t, c) \, dt$ from $t = 1$ to $t = 0$.

\subsection{Latent Diffusion and Dual-Stream Tokenization}
Following the Latent Diffusion Model (LDM) framework~\cite{rombach2022high}, an autoencoder compresses raw pixel images of target training resolution $H_0 \times W_0$ (e.g., $H_0 = W_0 = 256$) into a latent space $\mathcal{Z} = \mathbb{R}^{C \times H \times W}$ with $C$ channels and spatial dimensions $H \times W$ via spatial downsampling factor $f = H_0 / H = W_0 / W$ (e.g., $f=8, C=32$ for the FLUX.2 VAE). Following Diffusion Transformers (DiT)~\cite{peebles2023scalable}, noisy latent tensors $z_t \in \mathbb{R}^{C \times H \times W}$ at timestep $t$ are divided into non-overlapping spatial patches of size $p \times p$ (giving effective pixel patch size $P = f \cdot p = 16$), producing a flattened sequence of image tokens $z_{\text{img}} \in \mathbb{R}^{N \times D}$, where $N = (H/p)(W/p)$ is the image token count and $D=768$ is the transformer hidden dimension.

In Dual-Stream DiTs~\cite{esser2024scaling}, image tokens $z_{\text{img}}$ and text tokens $c_{\text{text}} \in \mathbb{R}^{L_{\text{text}} \times D}$ (of text sequence length $L_{\text{text}}$) are mapped through dedicated linear projections into image and context queries, keys, and values ($Q_{\text{img}}, K_{\text{img}}, V_{\text{img}}$ and $Q_{\text{ctx}}, K_{\text{ctx}}, V_{\text{ctx}}$) while sharing joint attention blocks. These projections are concatenated along the token sequence dimension ($\|$) to form the joint matrices $Q, K, V$:
\begin{equation}
    Q = [Q_{\text{img}} \,\|\, Q_{\text{ctx}}], \,\, 
    K = [K_{\text{img}} \,\|\, K_{\text{ctx}}], \,\, 
    V = [V_{\text{img}} \,\|\, V_{\text{ctx}}].
\end{equation}
Spatial layout is anchored via 2D axial Rotary Position Embeddings (RoPE)~\cite{su2024roformer}, which apply a block-diagonal rotation matrix $\mathbf{R}_{\mathbf{p}}$ parameterized by continuous coordinates $\mathbf{p} = (y, x)$ to yield rotated image queries and keys $\tilde{Q}_{\text{img}} = \mathbf{R}_{\mathbf{p}} Q_{\text{img}}$ and $\tilde{K}_{\text{img}} = \mathbf{R}_{\mathbf{p}} K_{\text{img}}$. Here, $(y, x)$ denotes the vertical and horizontal center coordinates of each patch in matrix row-major convention.

\subsection{Timestep Sampling and Time-SNR Shifting}
To counteract signal-to-noise ratio (SNR) compression when scaling image resolution, we introduce a resolution-dependent shift factor $s_{\text{shift}} \in \{1.0, 2.0\}$ and sample an unconstrained log-odds variable $s \sim \mathcal{N}(\mu, \sigma_s^2)$ from a normal distribution with mean $\mu = -\ln(s_{\text{shift}})$ and standard deviation $\sigma_s$, mapping it to $t = \text{sigmoid}(s)$~\cite{esser2024scaling}. We then compute the shifted timestep $t_{\text{shifted}} = (t \cdot s_{\text{shift}}) / [1 + (s_{\text{shift}} - 1)t]$ for Eq.~\eqref{eq:cfm_interpolation}, biasing sampling toward the noise-dominant regime ($t \to 1$).

\section{Asuka-FM Framework}
\label{sec:asuka_framework}

Asuka-FM establishes an end-to-end system and architectural framework designed to train high-capacity Diffusion Transformers (DiTs) from scratch within limited consumer-compute budgets. Our work extends the Micro-Budget Foundation Modeling research to an even more accessible boundary: pretraining a 206M Dual-Stream DiT on a \textit{single} consumer workstation equipped with an RTX 4090 GPU in 161.2 hours. Furthermore, the random square crops training \cite{wang2023patch} and the multi-captioning strategy of $i1$~\cite{i1arxiv2026} demonstrate that a capable foundation model can be trained from scratch on fewer than one million images.

Rather than relying on massive parameter scaling or distributed multi-node infrastructure, the framework co-designs the model architecture, memory hierarchy, and training curriculum around four complementary pillars:
\begin{enumerate}
    \item \textbf{Minimalist Dual-Stream Backbone (Section~\ref{sec:dual_stream}):} Completely eliminates per-block modulation overhead reducing compute and parameters.
    \item \textbf{Structured Sequence Sparsification via SPRINT (Section~\ref{sec:sprint}):} Reduces attention FLOPs by over 50\% while providing strong regularization.
    \item \textbf{Continuous Aspect-Normalized RoPE \& Shifted Cropping (Section~\ref{sec:calibrated_rope}):} Prevents rotary phase collapse, enforces static tensor shapes to eliminate JIT compilation breaks, and provides zero-shot virtual camera controls (pan and zoom).
    \item \textbf{Decoupled Latent-to-Pixel Adaptation (Section~\ref{sec:latent_to_pixel_decoupling}):} Transfers converged latent priors directly into raw pixel space, permanently bypassing the autoencoder decoding latency wall during downstream inference and reinforcement learning rollouts.
\end{enumerate}

\subsection{Minimalist Conditioning and Dual-Stream Backbone}
\label{sec:dual_stream}

\begin{table}[t]
\centering
\small
\caption{\textbf{Asuka-FM architecture specifications.} 206M-parameter Dual-Stream SPRINT DiT backbone.}
\label{tab:model_architecture}
\setlength{\tabcolsep}{6pt}
\renewcommand{\arraystretch}{1.05}
\begin{tabular}{lc}
\toprule
\textbf{Architectural Hyperparameter} & \textbf{Specification / Value} \\
\midrule
Latent Channel Dimension ($C$)        & 32 (FLUX.2-dev VAE, $f=8$) \\
Latent Patch Dimension ($p$)          & $2 \times 2$ (Effective $P=16$) \\
Hidden Dimension ($D$)                & 768 \\
Number of Attention Heads ($H$)       & 12 \\
Attention Head Dimension ($d_h$)      & 64 \\
Feedforward Network (FFN)             & SwiGLU ($D_{\text{ffn}} = 2048$) \\
Total Transformer Blocks ($L$)        & 13 Blocks ($2\text{ Dense} + 9\text{ Sparse} + 2\text{ Dense}$) \\
Positional Embedding                  & Calibrated Aspect RoPE ($\omega \in [\pi/4, \, 2\pi]$) \\
Text Encoder                          & LiquidAI/LFM2.5-230M (Frozen, $D_{\text{ctx}}=640$) \\
Text Adapter                          & 2-Layer Transformer ($D=768$, 6.09M params) \\
Patch Detailer Head (DiP)             & 4-Stage Conv U-Net (+10.09M params) \\
\midrule
\textbf{Total Pretraining Parameters} & \textbf{206.12M} (200.03M Backbone + 6.09M Adapter) \\
\bottomrule
\end{tabular}
\end{table}

Asuka-FM employs a 206M-parameter Dual-Stream DiT backbone optimized for single-GPU execution (Table~\ref{tab:model_architecture}). Standard AdaLN-Zero requires projecting conditioning embeddings into six independent affine modulation parameters per residual stream across every layer, introducing non-compute parameter overhead and intermediate memory traffic. Following $i1$~\cite{i1arxiv2026}, we eliminate per-block AdaLN modulation layers. To retain temporal guidance, the continuous scalar timestep $t \in [0, 1]$ is mapped via a sinusoidal MLP into a single temporal token $z_{\text{time}} \in \mathbb{R}^{1 \times D}$ and directly prepended to the text sequence $c_{\text{text}} \in \mathbb{R}^{L_{\text{text}} \times D}$, forming the combined context sequence $c = [z_{\text{time}} \,\|\, c_{\text{text}}] \in \mathbb{R}^{(1 + L_{\text{text}}) \times D}$. Time conditioning is broadcast through shared attention at near-zero parameter cost.

Text representations are extracted from a frozen language model (\texttt{LiquidAI/LFM2.5-230M}~\cite{liquidAI2026230M}, output dimension $D_{\text{ctx}}=640$) and mapped into the DiT hidden dimension $D=768$ via a 2-layer bidirectional Transformer Text Adapter ($6.09\text{M}$ params). Following attention mixing, image and context representations are updated via independent SwiGLU feedforward networks~\cite{shazeer2020glu} with intermediate dimension $D_{\text{ffn}} = 2048$. Sandwich RMSNorm~\cite{ding2021cogview} stabilizes bfloat16 mixed precision. In the final decoder block, projections for the text stream are frozen and pruned from the backward pass, avoiding dead gradient calculations.

\subsection{Sparse-Dense Residual Fusion (SPRINT)}
\label{sec:sprint}

To eliminate quadratic attention overhead during pretraining, we integrate SPRINT~\cite{sprint2026}, partitioning our $L=13$-layer transformer into three functional stages: a dense shallow encoder $f_\theta$ ($L_{\text{enc}}=2$), a sparse deep middle path $g_\theta$ ($L_{\text{mid}}=9$), and a dense decoder $h_\theta$ ($L_{\text{dec}}=2$).

\begin{figure*}[t]
\centering
\resizebox{\textwidth}{!}{%
\begin{tikzpicture}[
    every node/.style={font=\sffamily\footnotesize},
    box/.style={
        rectangle, draw=#1!80!black, fill=#1!12, rounded corners=4pt,
        line width=1.0pt, align=center, inner sep=4pt, minimum height=1.20cm
    },
    accentbox/.style={
        rectangle, draw=figamber!85!black, fill=figamber!18, rounded corners=4pt,
        line width=1.1pt, align=center, inner sep=4pt, minimum height=1.20cm
    },
    groupbox/.style={
        draw=gray!40, dashed, rounded corners=6pt, fill=gray!2, line width=0.8pt
    },
    headertab/.style={
        rectangle, draw=none, rounded corners=2pt, font=\footnotesize\bfseries,
        inner sep=3pt
    },
    flowarr/.style={
        -{Stealth[length=5pt, width=3.5pt]}, draw=gray!75, line width=1.0pt
    },
    interarr/.style={
        -{Stealth[length=6pt, width=4pt]}, draw=blue!75!black, line width=1.2pt
    },
    skiparr/.style={
        -{Stealth[length=5pt, width=3.5pt]}, draw=teal!85!black, line width=1.1pt, dashed
    }
]
    \node[box=blue, minimum width=3.1cm] (r1_img) at (0.0, 3.6) {
        \textbf{Input Canvas $I_{\text{orig}}$}\\
        \scriptsize $H_{\text{res}} \times W_{\text{res}} \ge 512\text{px}$\\
        \scriptsize Scale to $\min(H, W) = X$
    };

    \node[box=teal!80!black, minimum width=3.3cm, right=0.55cm of r1_img] (r1_crop) {
        \textbf{Shifted Crop \& RoPE}\\
        \scriptsize Static Window $X \times X$\\
        \scriptsize $[-\!r_H, r_H] \times [-\!r_W, r_W]$
    };

    \node[box=figindigo, minimum width=3.1cm, right=0.55cm of r1_crop] (r1_vae) {
        \textbf{FLUX.2 VAE}\\
        \scriptsize $f=8, C=32$\\
        \scriptsize In-RAM / FIFO Cache
    };

    \node[box=purple, minimum width=3.1cm, right=0.55cm of r1_vae] (r1_txt) {
        \textbf{Text Adapter}\\
        \scriptsize LFM2.5 Frozen LLM\\
        \scriptsize 2-Layer Transf. ($D=768$)
    };

    \node[accentbox, minimum width=3.1cm, right=0.55cm of r1_txt] (r1_ctx) {
        \textbf{Multimodal Context}\\
        \scriptsize Sinusoidal Time Token $z_{\text{time}}$\\
        \scriptsize Context $[z_{\text{time}} \,\|\, c_{\text{text}}]$
    };

    \draw[flowarr] (r1_img) -- (r1_crop);
    \draw[flowarr] (r1_crop) -- (r1_vae);
    \draw[flowarr] (r1_vae) -- (r1_txt);
    \draw[flowarr] (r1_txt) -- (r1_ctx);

    \node[box=blue, minimum width=2.8cm] (r2_patch) at (0.0, 0.0) {
        \textbf{Patchify ($p=2$)}\\
        \scriptsize $N = (H/2)(W/2)$\\
        \scriptsize Effective $P=16$
    };

    \node[box=teal!85!black, minimum width=3.0cm, right=0.5cm of r2_patch] (r2_enc) {
        \textbf{Shallow Core $f_\theta$}\\
        \scriptsize $L_{\text{enc}} = 2$ Dense Blocks\\
        \scriptsize Full Token Seq. $N$
    };

    \node[box=red!80!black, minimum width=3.2cm, right=0.5cm of r2_enc] (r2_mid) {
        \textbf{Middle Path $g_\theta$}\\
        \scriptsize $L_{\text{mid}} = 9$ Sparse Blocks\\
        \scriptsize 75\% Drop ($0.25N$ Active)
    };

    \node[box=teal!85!black, minimum width=3.2cm, right=0.5cm of r2_mid] (r2_dec) {
        \textbf{Dense Decoder $h_\theta$}\\
        \scriptsize Restore $+[\mathbf{M}]$ Tokens\\
        \scriptsize $W_{\text{fuse}}[f_t \,\|\, g_t^{\text{pad}}] + b_{\text{fuse}}$
    };

    \node[accentbox, minimum width=3.1cm, right=0.5cm of r2_dec] (r2_pdh) {
        \textbf{Decoupled PDH}\\
        \scriptsize 10.09M Conv U-Net\\
        \scriptsize Direct $v$-Pred $\to$ RGB
    };

    \draw[flowarr] (r2_patch) -- (r2_enc);
    \draw[flowarr] (r2_enc) -- (r2_mid);
    \draw[flowarr] (r2_mid) -- (r2_dec);
    \draw[flowarr] (r2_dec) -- (r2_pdh);

    \draw[skiparr] (r2_enc.south) -- ++(0, -0.32) 
        -| node[pos=0.25, below, font=\scriptsize, text=teal!90!black] {Dense Shallow Residual Skip $f_t$ ($N$ Tokens)} 
        (r2_dec.south);

    \draw[interarr] (r1_vae.south) -- ++(0, -0.45) -| (r2_patch.north);
    \draw[interarr] (r1_ctx.south) -- ++(0, -0.45) -| ($(r2_enc.north) + (0.5, 0)$);
    \node[font=\scriptsize\sffamily, text=blue!85!black, fill=white, inner sep=2pt, rounded corners=2pt] 
        at ($(r2_patch.north) + (2.5, 0.45)$) 
        {Tokenized Latents + In-Context Time Conditioning $[z_{\text{time}} \,\|\, c_{\text{text}}]$};

    \begin{pgfonlayer}{background}
        \draw[groupbox] ($(r1_img.north west) + (-0.25, 0.48)$) rectangle ($(r1_ctx.south east) + (0.25, -0.22)$);
        \draw[groupbox] ($(r2_patch.north west) + (-0.25, 0.80)$) rectangle ($(r2_pdh.south east) + (0.25, -0.55)$);
    \end{pgfonlayer}

    \begin{pgfonlayer}{foreground}
        \node[headertab, text=blue!90!black, fill=blue!10] at ($(r1_img.north west) + (3.8, 0.48)$) {
            Stage 1: Shifted Crop, Aspect RoPE \& Latent Context Assembly
        };
        \node[headertab, text=figindigo!90!black, fill=figindigo!10] at ($(r2_patch.north west) + (3.8, 0.80)$) {
            Stage 2: SPRINT Sequence Sparsification \& Decoupled Pixel Transfer
        };
    \end{pgfonlayer}
\end{tikzpicture}%
}
\vspace{-4pt}
\caption{\textbf{Asuka-FM system-architectural co-design pipeline.} \textbf{Top Row (Stage 1):} Raw images $I_{\text{orig}}$ undergo shifted square cropping to enforce static tensor dimensions ($X \times X$) and eliminate JIT recompilations, while continuous aspect-normalized RoPE prevents phase collapse. Precomputed FLUX.2 VAE latents $z_t$ receive prepended in-context temporal conditioning ($z_{\text{time}}$) alongside text representations from a 2-layer adapter without per-block AdaLN parameters. \textbf{Bottom Row (Stage 2):} The 13-layer dual-stream DiT executes SPRINT sequence sparsification: shallow blocks ($f_\theta$) supply high-frequency dense skips $f_t$, deep middle blocks ($g_\theta$) process macro-semantics over a 75\%-sparse grid, and dense decoder blocks ($h_\theta$) fuse representations to predict flow trajectories. Finally, the lightweight 10.09M Patch Detailer Head (PDH) transfers representations directly to RGB pixels under native $v$-prediction.}
\label{fig:macro_system_pipeline}
\end{figure*}
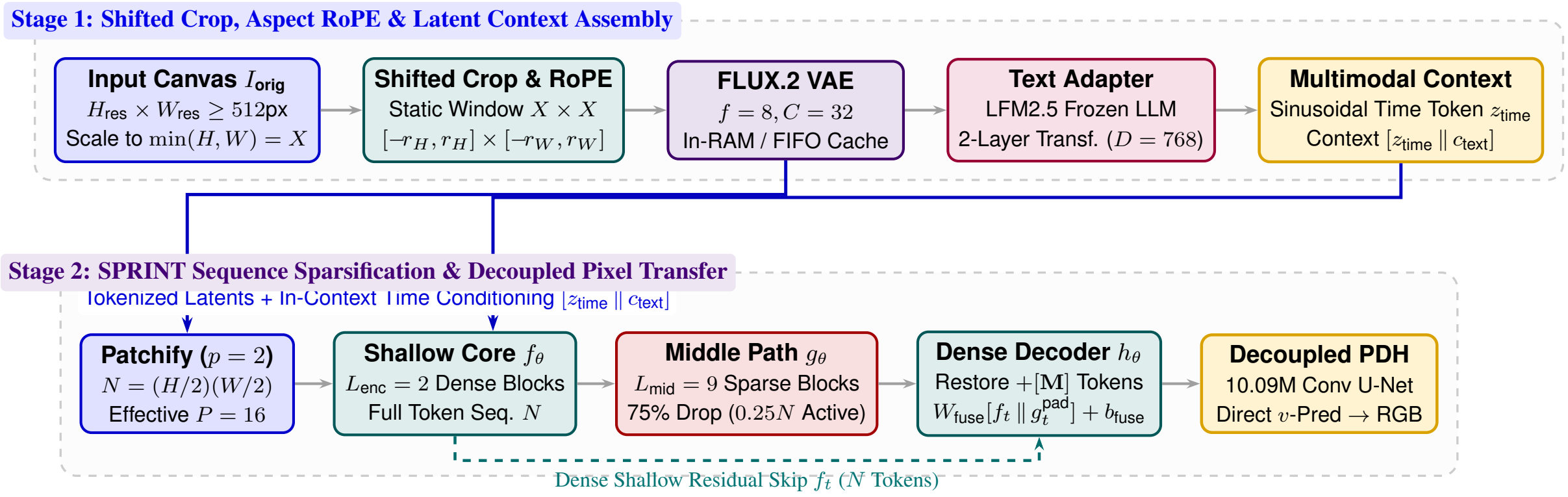

\textbf{Structured Group-Wise Subsampling.} Following SPRINT~\cite{sprint2026}, our 13-layer transformer is partitioned into three functional stages (Figure~\ref{fig:macro_system_pipeline}): 
\begin{enumerate}
    \item (i)~a \textit{dense shallow encoder} $f_\theta$ ($L_{\text{enc}}=2$) processing the batched full token sequence $x_t \in \mathbb{R}^{B \times N \times D}$ (for batch size $B$ and $N$ image tokens) to preserve local edge representations $f_t \in \mathbb{R}^{B \times N \times D}$. 
    \item (ii)~a \textit{sparse deep middle path} $g_\theta$ ($L_{\text{mid}}=9$, $78.4\%$ of backbone weights) processing only the active subsampled tokens $g_t^{\text{drop}} \in \mathbb{R}^{B \times N_{\text{sp}} \times D}$ by deterministically retaining $k=1$ token per $n \times n = 2 \times 2$ spatial group ($75\%$ drop ratio $r=0.75$, leaving $N_{\text{sp}} = (1-r)N = 0.25N$ active tokens) to route macro-semantics at low FLOP cost. 
    \item (iii)~a \textit{dense decoder} $h_\theta$ ($L_{\text{dec}}=2$) that restores sequence length by inserting a learnable mask vector $\mathbf{M} \in \mathbb{R}^D$ at dropped positions to yield $g_t^{\text{pad}} \in \mathbb{R}^{B \times N \times D}$, fuses shallow and deep paths via a linear projection with weight $W_{\text{fuse}} \in \mathbb{R}^{D \times 2D}$ and bias $b_{\text{fuse}} \in \mathbb{R}^D$ into $h_t = W_{\text{fuse}}[f_t \,\|\, g_t^{\text{pad}}] + b_{\text{fuse}}$, and predicts the velocity field $\hat{v}_t$.
\end{enumerate}

\textbf{Analytical Speedup Model.} At $512 \times 512$ ($N=1024, N_{\text{sp}}=0.25N=256, D=768, D_{\text{ffn}}=2048$), neglecting the short text sequence, the theoretical FLOP complexity $\mathcal{C}_{\text{SPRINT}}$ under SPRINT is:
\begin{equation}
\begin{aligned}
    \mathcal{C}_{\text{SPRINT}} &= (L_{\text{enc}} + L_{\text{dec}}) \Big( 2ND \big(2D + N + 4D_{\text{ffn}}\big) \Big) \\
    &\quad + L_{\text{mid}} \Big( 2N_{\text{sp}}D \big(2D + N_{\text{sp}} + 4D_{\text{ffn}}\big) \Big) \\
    &\quad + 2ND^2.
\end{aligned}
\label{eq:sprint_complexity}
\end{equation}
Compared to a homogeneous dense DiT of complexity $\mathcal{C}_{\text{dense}}$ (evaluating all $L=13$ blocks with $N$ tokens and no fusion overhead), Eq.~\eqref{eq:sprint_complexity} yields a theoretical FLOP ratio of $\alpha_{\text{FLOP}} = \mathcal{C}_{\text{SPRINT}} / \mathcal{C}_{\text{dense}} \approx 0.474$ ($2.11\times$ arithmetic acceleration, with self-attention FLOPs reduced by $64.9\%$). Accounting for memory bus traffic, text-stream compute, and kernel launch overheads on physical silicon (RTX 4090), the model predicts an achievable speedup of $1.65\times\text{--}1.95\times$, closely matching our empirical measurement of $1.67\times$ ($108.5\,\text{img/s}$ sparse vs. $64.8\,\text{img/s}$ dense at $512^2$, Table~\ref{tab:training_stages}).

\subsection{Calibrated Continuous Aspect-Normalized RoPE}
\label{sec:calibrated_rope}
Standard discrete integer positional encodings fail when generalizing across variable aspect ratios, inducing artifacts. To resolve this, previous work~\cite{hdm2025} treats vertical and horizontal spatial frequencies as learnable parameters ($\omega_h, \omega_w \in \mathbb{R}^{N_H \times (d_h/4)}$ for patch-grid height $N_H$ and head dimension $d_h$) initialized via log-spaced heuristics. In contrast, Asuka-FM establishes a parameter-free continuous coordinate formulation. We mathematically diagnose why naive continuous RoPE undergoes spatial phase collapse and derive a closed-form, calibrated frequency spectrum that achieves isotropic decay without optimizer overhead.

\textbf{Continuous Coordinate Mapping.} For a latent patch grid of height $N_H = H/p$ and width $N_W = W/p$, aspect-ratio scale factors $r_H$ and $r_W$ preserving unit canvas area satisfy $r_H \cdot r_W = 1.0$ and $r_H / r_W = N_H / N_W$, yielding $r_H = \sqrt{N_H/N_W}$ and $r_W = \sqrt{N_W/N_H}$. Continuous normalized coordinates $(y_{\text{pos}}, x_{\text{pos}})$ for a patch centered at $(y_{\text{center}}, x_{\text{center}})$ are bounded within $[-r_H, r_H] \times [-r_W, r_W]$ via:
\begin{equation}
    y_{\text{pos}} = \left( \frac{y_{\text{center}}}{H} \right) 2 r_H - r_H, \quad
    x_{\text{pos}} = \left( \frac{x_{\text{center}}}{W} \right) 2 r_W - r_W.
    \label{eq:continuous_coords}
\end{equation}

\begin{figure*}[t]
\centering
\includegraphics[width=0.96\textwidth]{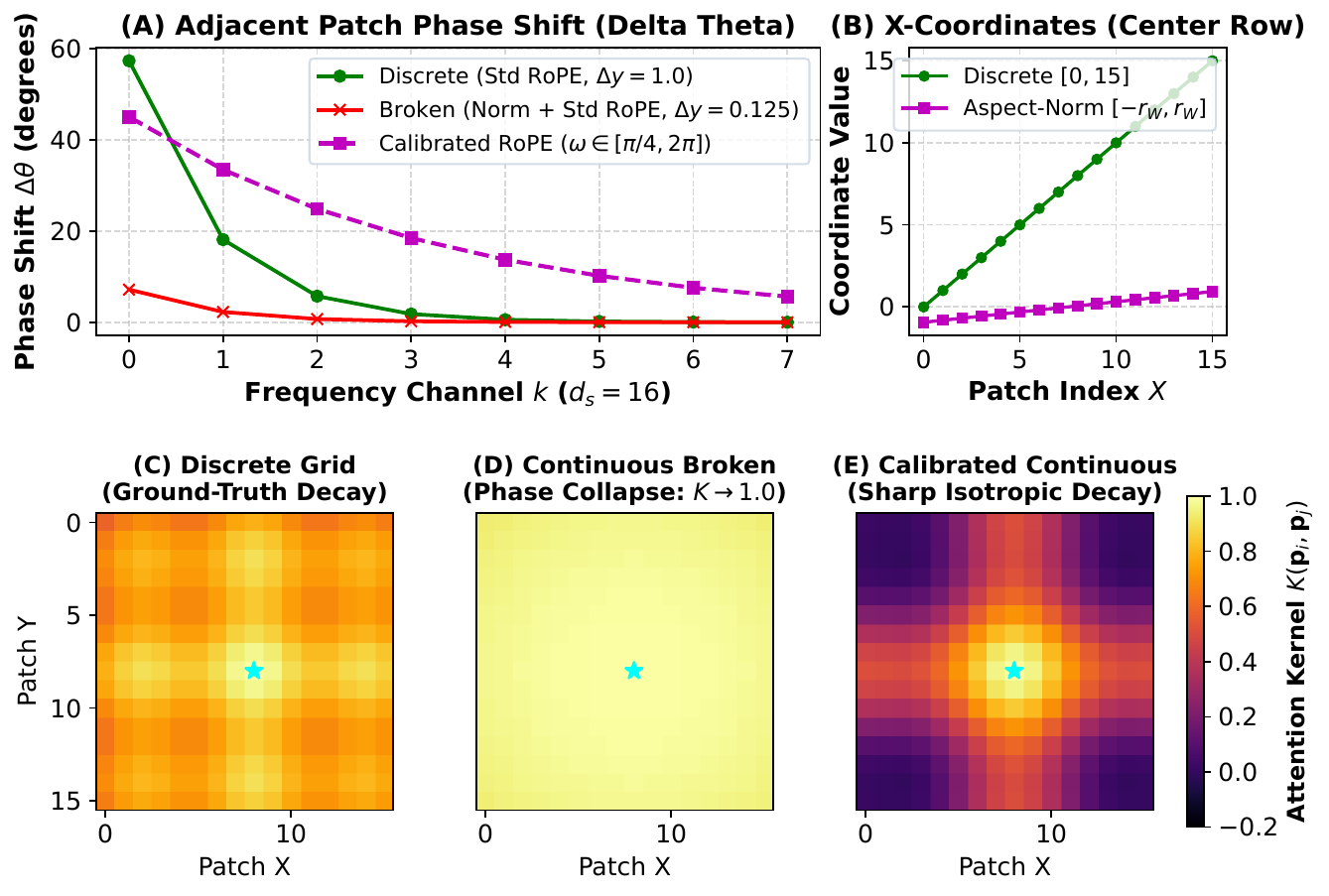}
\caption{\textbf{Continuous RoPE coordinate geometry and frequency calibration analysis.} 
\textbf{Top Row:} (A) Adjacent patch phase step $\Delta\theta_k$ across channels $k \in [0, 7]$: uncalibrated normalized coordinates collapse phase shifts to $<7.16^\circ$ ($K \to 1.0$), while calibrated frequencies ($\omega \in [\pi/4, 2\pi]$) restore angular separation ($22.5^\circ\text{--}45^\circ$). (B) Center-row coordinate trajectories comparing discrete integer indices $[0, 15]$ against continuous coordinates $[-r_W, r_W]$. 
\textbf{Bottom Row:} Spatial attention kernel heatmaps $K(\mathbf{p}_i, \mathbf{p}_j)$ centered at probe patch $(8, 8)$ (cyan star). (C) Discrete integer baseline exhibits natural radial decay. (D) Uncalibrated continuous RoPE undergoes phase collapse ($K \approx 1.0$). (E) Calibrated continuous RoPE restores sharp isotropic radial decay across arbitrary canvas dimensions without learned parameters.}
\label{fig:rope_calibration}
\end{figure*}

\textbf{Continuous RoPE Phase Collapse.} In 2D axial RoPE, the head dimension $d_h$ is partitioned equally across vertical and horizontal axes, yielding $d_s = d_h / 4$ two-dimensional rotation channels per axis. Let $\omega_k$ denote the spatial frequency for channel $k \in \{0, \dots, d_s - 1\}$. The normalized spatial attention kernel $K(\mathbf{p}_i, \mathbf{p}_j)$ between patches at coordinates $\mathbf{p}_i = (y_i, x_i)$ and $\mathbf{p}_j = (y_j, x_j)$ evaluates to:
\begin{equation}
    K(\mathbf{p}_i, \mathbf{p}_j) = \sum_{k=0}^{d_s - 1} \frac{\cos\big((y_i - y_j)\omega_k\big) + \cos\big((x_i - x_j)\omega_k\big)}{2 d_s}.
    \label{eq:rope_kernel}
\end{equation}

In standard discrete RoPE~\cite{su2024roformer}, adjacent vertical patch separation is fixed at $\Delta y = 1.0$. However, when mapped to continuous normalized coordinates $y \in [-r_H, r_H]$ on an $N_H \times N_W$ patch canvas, the normalized inter-patch separation contracts to $\Delta y_{\text{norm}} = 2 r_H / N_H \ll 1.0$. Under standard geometric base frequencies ($\omega_k = 10000^{-k/d_s} \le 1.0$ across $2d_s = d_h/2$ dimensions per axis), the adjacent phase shift $\Delta \theta_k = \Delta y_{\text{norm}} \cdot \omega_k$ contracts to $\Delta \theta_k \le 2 r_H / N_H$. Evaluating the second-order Taylor expansion of the cosine kernel around $\Delta \theta_k \approx 0$ between adjacent patches $\mathbf{p}_i$ and $\mathbf{p}_{i+1}$ (shown for a diagonal step $\Delta x_{\text{norm}} = \Delta y_{\text{norm}}$; an axial step yields prefactor $\frac{1}{4d_s}$) gives:
\begin{equation}
\begin{aligned}
    K(\mathbf{p}_i, \mathbf{p}_{i+1}) &= 1 - \frac{1}{2d_s} \sum_{k=0}^{d_s - 1} (\Delta \theta_k)^2 + \mathcal{O}(N_H^{-4}) \\
    &= 1.0 - \mathcal{O}\left(N_H^{-2}\right).
\end{aligned}
\label{eq:rope_collapse}
\end{equation}
Here, $\mathcal{O}(\cdot)$ denotes Landau's big-$\mathcal{O}$ notation, characterizing the asymptotic upper bound on the Taylor expansion truncation error as grid resolution $N_H \to \infty$. Specifically, since $\cos(u) = 1 - u^2/2 + \mathcal{O}(u^4)$ and adjacent phase shifts scale inversely with resolution ($\Delta \theta_k \propto N_H^{-1}$), the remainder terms vanish at least as fast as $N_H^{-4}$. Summing over all $d_s$ frequency channels proves that the kernel's deviation from identity decays asymptotically as $\mathcal{O}(N_H^{-2})$. 

Consequently, as canvas resolution $N_H$ scales, $K(\mathbf{p}_i, \mathbf{p}_{i+1}) \to 1.0$ uniformly across the grid (Figure~\ref{fig:rope_calibration}D). Adjacent tokens lose spatial discriminability ($\Delta \theta_k \to 0$), causing phase collapse and anatomical tiling artifacts.

\textbf{Full-Spectrum Frequency Calibration.} To guarantee isotropic radial decay without harmonic aliasing ($\Delta x \cdot \omega_{\max} = 2\pi$ for horizontal coordinate separation $\Delta x$ and peak frequency $\omega_{\max} = f_{\max}$), spatial frequencies must satisfy two boundary constraints between upper bound $f_{\max}$ and lower bound $f_{\min}$:
\begin{itemize}
    \item \textit{Local Dispersion Bound ($f_{\max}$):} On the canonical $256^2$ square patch grid ($N_H = 16, r_H = 1.0$), adjacent patches separated by $\Delta p_{\min} = 2r_H / N_H = 2.0/16$ must achieve angular dispersion $\Delta \theta \in [\pi/4, \pi/2]$ to maintain local boundary discriminability without exceeding the Nyquist limit $\pi$, yielding $f_{\max} = \frac{\pi / 4}{2.0 / 16} = 2\pi \approx 6.283$ (Figure~\ref{fig:rope_calibration}A).
    \item \textit{Global Monotonicity Bound ($f_{\min}$):} Across the maximum canvas separation $\Delta p_{\max} = 2 r_H \approx 2.0$, the slowest channel must not complete a half-cycle ($<\pi$), preventing periodic harmonic resonance across the canvas and setting $f_{\min} = \frac{\pi / 2}{2.0} = \frac{\pi}{4} \approx 0.785$.
\end{itemize}
Frequencies $\omega_k$ are interpolated log-linearly from $f_{\max}$ to $f_{\min}$ across channels $k \in \{0, \dots, d_s - 1\}$:
\begin{equation}
    \omega_k = \exp\left( \ln(f_{\max}) - \frac{k}{d_s - 1} \ln\left(\frac{f_{\max}}{f_{\min}}\right) \right).
    \label{eq:calibrated_spectrum}
\end{equation}
This closed-form formulation enforces sharp radial decay across arbitrary aspect ratios without requiring learned parameters or optimizer tracking (Figure~\ref{fig:rope_calibration}E vs. discrete integer baseline in Figure~\ref{fig:rope_calibration}C). These frequencies are registered as static buffers (\texttt{register\_buffer}), avoiding optimizer tracking entirely.

\subsection{Shifted Square Crop and Camera Dynamics}
\label{sec:shifted_crop}
Variable aspect-ratio training typically requires padding or dynamic batching, triggering continuous JIT graph recompilations under \texttt{torch.compile}. Following Patch Diffusion~\cite{wang2023patch} and HDM~\cite{hdm2025}, we train strictly on fixed square crops while conditioning on continuous coordinates:
\begin{enumerate}
    \item An input image $I_{\text{orig}}$ with native resolution $H_{\text{res}} \times W_{\text{res}}$ is isotropically scaled to $H_{\text{scaled}} \times W_{\text{scaled}}$ such that its shorter side matches training dimension $X$ (e.g., $X=256$ or $512$), i.e., $\min(H_{\text{scaled}}, W_{\text{scaled}}) = X$.
    \item Continuous 2D coordinates $\mathbf{P}_{\text{coord}} \in \mathbb{R}^{N_{H,\text{scaled}} \times N_{W,\text{scaled}} \times 2}$ are calculated across the scaled patch grid ($N_{H,\text{scaled}} = H_{\text{scaled}}/P$, $N_{W,\text{scaled}} = W_{\text{scaled}}/P$, with effective patch stride $P=16$) using Eq.~\eqref{eq:continuous_coords}.
    \item A random $X \times X$ window is extracted, yielding the aligned pair of cropped image and cropped position map $(I_{\text{crop}}, \mathbf{P}_{\text{crop}})$. Because tensor shapes remain strictly static ($X \times X$), GPU execution graphs execute without recompilations or CUDA graph invalidations.
\end{enumerate}

\textbf{Zero-Shot Virtual Camera Inversion.} Because the model is conditioned on continuous coordinates rather than discrete token indices, affine transformations parameterized by an optical magnification factor $\text{zoom} \in \mathbb{R}^+$ and normalized pan offsets $(x_{\text{shift}}, y_{\text{shift}})$ can be applied to the evaluation coordinate mesh $(x_{\text{pos}}, y_{\text{pos}})$ at inference time to steer the virtual camera viewport $(x_{\text{pos}}', y_{\text{pos}}')$ without parameter fine-tuning:
\begin{equation}
    y_{\text{pos}}' = \frac{y_{\text{pos}}}{\text{zoom}} + (y_{\text{shift}} \cdot 2 r_H), \quad
    x_{\text{pos}}' = \frac{x_{\text{pos}}}{\text{zoom}} + (x_{\text{shift}} \cdot 2 r_W).
    \label{eq:camera_transforms}
\end{equation}
Mathematically, $\text{zoom} > 1.0$ contracts the continuous coordinate query bounds from $[-r_W, r_W]$ to $[-\frac{r_W}{\text{zoom}}, \frac{r_W}{\text{zoom}}]$. Because the transformer outputs tokens over a fixed $N$-patch grid, querying this contracted coordinate neighborhood forces the model to synthesize a smaller spatial sub-region across the full canvas, creating continuous optical focal magnification (zoom in). Conversely, $\text{zoom} < 1.0$ expands the coordinate field to incorporate wide-angle scene context. Concurrently, $(x_{\text{shift}}, y_{\text{shift}})$ applies a linear translation vector $(\Delta x, \Delta y) = (x_{\text{shift}} \cdot 2r_W, y_{\text{shift}} \cdot 2r_H)$ to the coordinate center, shifting the camera frustum horizontally and vertically across identical noise seeds (Appendix~\ref{sec:camera_dynamics}).

\subsection{Latent-to-Pixel Decoupling via DiP Head}
\label{sec:latent_to_pixel_decoupling}
While latent diffusion provides rapid semantic pretraining, the frozen VAE decoder imposes latency bottlenecks during few-step inference and online reinforcement learning rollouts (e.g., Flow-GRPO~\cite{liu2025flowgrpo}). We bridge this trade-off by pretraining in latent space and transferring to raw pixel space via a lightweight convolutional Patch Detailer Head (PDH) based on DiP~\cite{dip2026}.

\textbf{PDH Architecture.} Patch-level visual tokens $s_i \in \mathbb{R}^D$ ($i \in \{1, \dots, N\}$) emitted by the DiT decoder $h_\theta$ are mapped into raw RGB patches via a 4-stage convolutional U-Net adding only 10.09M parameters ($\approx 5\times$ smaller than standard autoencoder decoders). To prevent high-frequency aliasing from transposed convolutions, Asuka-FM employs nearest-neighbor upsampling coupled with $3 \times 3$ convolutions (\texttt{Upsample} + \texttt{Conv2d}). Reconstructed patches are folded back into the full RGB pixel canvas of shape $[B, 3, H_0, W_0]$ (for batch size $B$ and 3 RGB channels).

\textbf{Direct Velocity Regression vs. Clean Data Prediction.} Previous pixel diffusion studies~\cite{jiang2026empirical,li2026basicsletdenoisinggenerative} advocated clean data prediction ($x_0$-prediction). However, converting predicted clean states $\hat{x}_0$ into velocity targets $v_t$ from noisy pixel states $x_t$ at timestep $t$ via Eq.~\eqref{eq:cfm_interpolation}, $v_t = (x_t - \hat{x}_0)/t$, introduces mathematical instability. As $t \to 0$, this conversion suffers from division-by-zero singularities under bfloat16, requiring artificial clamping and dynamic percentile clipping that flattens dynamic range and induces chromatic blowout. In contrast, our ablations (corroborated by Iris-3B~\cite{cai2026iris3bgoinglatentpixelspace}) demonstrate that direct velocity regression ($v$-prediction) is superior. Retaining native $v$-prediction preserves continuous probability paths, stabilizes optimization, and eliminates the need for dynamic activation clipping.

\begin{table*}[t]
\centering
\captionsetup{font=small, labelfont=bf, skip=4pt}
\caption{\textbf{Pretraining progression and compute metrics of Asuka-FM on a single consumer NVIDIA RTX 4090 GPU (24GB).} All stages use continuous aspect-normalized RoPE, bfloat16, and \texttt{torch.compile}. Pretraining uses an effective batch size of 256. Unabridged configurations and token processing rates are detailed in Table~\ref{tab:training_stages_full} (Appendix~\ref{sec:appendix_full_tables}). Evaluated at \$0.33/GPU-hour.}
\label{tab:training_stages}
\setlength{\tabcolsep}{4.2pt}
\renewcommand{\arraystretch}{1.08}
\resizebox{\linewidth}{!}{%
\begin{tabular}{lccccccccr}
\toprule
\textbf{Stage / Phase} & \textbf{Res.} \textbf{($X^2$)} & \textbf{Drop} \textbf{($r$)} & \textbf{Samples} 
& \shortstack{\textbf{Throughput}\\\textbf{(img/s)}} & \shortstack{\textbf{MFU (\%)}\\\textbf{Theo / Act}$^\dagger$} 
& \shortstack{\textbf{Compute}\\\textbf{(EFLOPs)}$^*$} & \shortstack{\textbf{Wall}\\\textbf{Time}} & \shortstack{\textbf{Cost}\\\textbf{(\$)}} \\
\midrule
\multicolumn{9}{l}{\textit{\textbf{Stage 1: Base Semantic Pretraining ($256 \times 256$, Latent Patch Grid $16 \times 16 = 256$ Image Tokens)}}} \\
Offline VAE Encoding$^\ddagger$ & $256^2$ & --   & 0.625M & 45.2 & --              & --    & 3.8 h  & \$1.27 \\
1A: SPRINT Sparse               & $256^2$ & 0.75 & 46.88M & 282.6 & 74.4\% / 86.8\% & 23.74 & 47.0 h & \$15.51 \\
1B: Dense Refine                & $256^2$ & 0.00 & 12.50M & 172.5 & 62.9\% / 73.4\% & 8.73  & 21.0 h & \$6.93 \\
\midrule
\multicolumn{9}{l}{\textit{\textbf{Stage 2: High-Resolution Adaptation ($512 \times 512$, Latent Patch Grid $32 \times 32 = 1,024$ Image Tokens)}}} \\
Offline VAE Encoding$^\ddagger$ & $512^2$ & --   & 0.625M & 20.6  & --              & --    & 8.4 h  & \$2.78 \\
2A: High-Res Sparse             & $512^2$ & 0.75 & 15.63M & 108.5 & 58.5\% / 68.3\% & 16.24 & 40.0 h & \$13.20 \\
2B: Dense Refine                & $512^2$ & 0.00 & 9.38M  & 64.8  & 61.0\% / 71.2\% & 16.94 & 41.0 h & \$13.53 \\
\midrule
\multicolumn{3}{l}{\textbf{Cumulative DiT Training Total (Stages 1 + 2)}} & \textbf{84.38M} & -- & \textbf{64.8\% / 75.6\%} & \textbf{65.65} & \textbf{149.0 h} & \textbf{\$49.17} \\
\multicolumn{3}{l}{\textbf{Cumulative Grand Total (Including Offline VAE Pre-Encoding)}} & -- & -- & -- & \textbf{65.65} & \textbf{161.2 h} & \textbf{\$53.20} \\
\bottomrule
\end{tabular}%
}
\vspace{2pt}
\begin{flushleft}
\footnotesize
$^\dagger\text{MFU}_{\text{act}}$ evaluates actual tensor ops ($7PD$ per sample, for parameter count $P_{\text{param}}$ and tokens per sample $D_{\text{tok}}$) under SwiGLU recomputation; $\text{MFU}_{\text{theo}}$ denotes standard algorithmic utilization ($6PD$).\\
$^*1\text{ EFLOP} = 10^{18}\text{ FLOPs} = 10^3\text{ PFLOPs} = 10^{-3}\text{ ZFLOPs}$. Cumulative theoretical compute is $56.27\text{ EFLOPs}$.\\
$^\ddagger$One-time offline encoding using FLUX.2 VAE ($C=32, f=8$) permanently removing autoencoder overhead from training.
\end{flushleft}
\end{table*}

\section{Progressive Training and Systems Engineering}
\label{sec:systems_engineering}

\subsection{Curated Multi-Source Dataset and Ingestion Pipeline}

\paragraph{Open-Domain General Corpus (625K).} Following empirical findings from the $i1$~\cite{i1arxiv2026}, training on balanced mixtures with $100\text{k}\text{--}400\text{k}$ high-quality samples per source yields competitive generative performance without massive web scraping. We curate an open-domain dataset of \textbf{625,000 samples} by extracting exactly 125,000 samples across five sources including synthetic and real samples: (i) FLUX-Reason-6M~\cite{fang2025fluxreason6mprismbenchmillionscale} (complex composition); (ii) GPT-Edit-Simpler (paired instructional edits); (iii) Midjourney-v6-Recap (artistic lighting and textures); (iv) Pexels (natural photography and anatomy); and (v) TextAtlas5M~\cite{csu2024textatlas} (text-rich scenes). All images satisfy $\min(H, W) \ge 512\,\text{px}$.

\textbf{VLM Multi-Captioning.} We maintain five synthetic captions per image generated via \textit{Qwen3-VL-30B-A3B} from $i1$~\cite{i1arxiv2026} dataset. At each epoch, a caption is sampled dynamically, providing continuous semantic regularization that prevents text memorization. Training context is partitioned into three discrete tiers ($[77, 152, 227]$ tokens), reducing padding overhead by up to 66\% and triggering TorchInductor recompilation exactly twice during warmup.

\paragraph{Curated Danbooru Corpus (337K) \& Tag-Rarity Weighting.} To investigate domain specialization and high-density concept binding, we curate 337,000 illustrations\footnote{\url{https://huggingface.co/datasets/aipracticecafe/curated-danbooru-2026}} from Danbooru2023~\cite{danbooru2023}. Because community tags often lack structural details, each image is processed through a SwinV2 tagger~\cite{liu2022swinv2} to upsample descriptive metadata. Prompts are formatted into a canonical prefix structure with dynamically shuffled general tags:
\begin{displayquote}
\texttt{[character count], [character name], [series], [artist], [rating], [aesthetic tags], [year], [shuffled general tags]}
\end{displayquote}
To prevent frequent tags (e.g., \texttt{1girl}, \texttt{solo}) from dominating gradient updates while rare artist styles remain undertrained, we scale the flow matching loss $\mathcal{L}_{\text{CFM}}$ by an inverse tag-frequency prompt weight $w_{\text{prompt}}$ to obtain the weighted loss $\mathcal{L}_{\text{weighted}}$. Given the active tag set $\mathcal{T}$ of cardinality $|\mathcal{T}|$, dataset size $N_{\text{total}}$, and corpus frequency $N_k$ of tag $k \in \mathcal{T}$, we define:
\begin{equation}
    w_{\text{prompt}} = 1.0 + \frac{1}{|\mathcal{T}|} \sum_{k \in \mathcal{T}} \ln\left( \frac{N_{\text{total}}}{N_k + 1} \right), \quad \mathcal{L}_{\text{weighted}} = w_{\text{prompt}} \cdot \mathcal{L}_{\text{CFM}}.
    \label{eq:rarity_loss_weight}
\end{equation}
where $\mathcal{T}$ is the active tag set, $N_{\text{total}}$ is the dataset size, and $N_k$ is the corpus frequency of tag $k$.

We use this dataset to train an anime-specialized Asuka-FM model\footnote{\url{https://huggingface.co/aipracticecafe/asuka-fm}} in order to validate our Open-Domain dataset results.

\subsection{Progressive Pretraining Recipe}
Pretraining advances through two core resolution tiers (Table~\ref{tab:training_stages}; see Table~\ref{tab:training_stages_full} in Appendix~\ref{sec:appendix_full_tables} for the complete progression matrix including batch sizes, epoch schedules, and average token rates): base semantic anchoring at $256 \times 256$ and texture synthesis at $512 \times 512$. All stages employ 8-bit AdamW ($\beta_1 = 0.9, \beta_2 = 0.95$, weight decay $\lambda = 0.01$) via \texttt{bitsandbytes}, mixed-precision bfloat16 with TF32 tensor cores, and FlashAttention-2 inside \texttt{torch.compile}. Classifier-Free Guidance dropout is set to $p_{\text{uncond}} = 0.10$.

\textbf{Execution Breakdown.} Cumulatively, \textbf{149.0 hours of DiT latent pretraining (\$49.17 cost)} executed \textbf{65.65 ExaFLOPs ($0.0657\text{ ZFLOPs}$)} at an average sustained utilization of $\text{MFU}_{\text{act}} = 75.6\%$. Including one-time offline VAE pre-encoding (12.2h, \$4.03), the entire latent pretraining pipeline converges in \textbf{161.2 total GPU hours (\$53.20)}.

\subsection{Memory Hierarchy Optimizations}
Pretraining experiments were executed on a single workstation: NVIDIA RTX 4090 GPU (24GB GDDR6X, 165.2 TFLOPS peak), AMD Ryzen 9 7950X CPU, and 128GB DDR5 RAM under Ubuntu 24.04 and PyTorch 2.8.

(i)~\textbf{In-RAM Latent Caching ($256^2$):} Pre-encoded latents for all 625k samples at $256^2$ require $\approx 52\,\text{GB}$. Caching latents directly in system RAM eliminated disk I/O and autoencoder latency, sustaining $282.6\,\text{img/s}$ ($86.8\%\,\text{MFU}_{\text{act}}$). (ii)~\textbf{Network Latent Streaming ($512^2$):} At $512^2$, latents expand to $\approx 203\,\text{GB}$, exceeding system RAM. Latents were pre-encoded offline (8.4h) and streamed asynchronously via dynamic FIFO queues, sustaining $108.5\,\text{img/s}$ ($68.3\%\,\text{MFU}_{\text{act}}$). (iii)~\textbf{Selective Activation Checkpointing:} Checkpointing only the SwiGLU blocks while retaining joint self-attention tensors reduced peak activation memory by $\approx 30\%$, enabling micro-batch size 64 at $512^2$ within 24\,GB VRAM. Recomputing MLP projections adds $+1PD$ work factor ($7PD$ on silicon vs. $6PD$ theoretical, where $P$ and $D$ here denote active model parameters and tokens per sample).

\section{Experiments and Empirical Analysis}
\label{sec:experiments}
We ablated our architectural components and the SPRINT algorithm on a controlled toy T2I benchmark (details in Appendix~\ref{sec:toy_diffusion}); here, we focus on open-domain foundation benchmarks and micro-architectural systems profiling.

\begin{table*}[t]
\centering
\captionsetup{font=small, labelfont=bf, skip=4pt}
\caption{\textbf{Quantitative comparison on representative text-to-image benchmarks (GenEval and DPG-Bench).} Results marked with ($^*$) denote numbers reported in prior literature~\protect\cite{cai2025zimage,i1arxiv2026,chng2026looped,minit2i2026}. Models marked with $^\dagger$ indicate fine-tuning on the \texttt{BLIP3o-60K} dataset, which specifically targets GenEval metrics. Asuka-FM is trained from scratch on open data without benchmark-specific fine-tuning. The full GenEval breakdown is provided in Table~\ref{tab:benchmark_comparison_full} in Appendix~\ref{sec:appendix_full_tables}.}
\label{tab:benchmark_comparison}
\setlength{\tabcolsep}{5.5pt}
\renewcommand{\arraystretch}{1.06}
\resizebox{\linewidth}{!}{%
\begin{tabular}{lccccc}
\toprule
\textbf{Model} & \textbf{Params} & \shortstack{\textbf{Hardware}\\\textbf{Budget}} & \shortstack{\textbf{Training}\\\textbf{Data}} & \shortstack{\textbf{GenEval}\\\textbf{Total $\uparrow$}} & \shortstack{\textbf{DPG-Bench}\\\textbf{Total $\uparrow$}} \\
\midrule
\multicolumn{6}{l}{\textit{\textbf{Industrial Foundation Models (Multi-Billion Parameters, Datacenter Compute)}}} \\
SDXL Base~\cite{sdxl2024}                & 2.6B  & Hundreds of A100s & Millions (Web) & 0.406$^*$          & 72.40$^*$ \\
SD3 Medium~\cite{esser2024scaling}       & 2.0B  & Massive Cluster   & Millions (Web) & 0.620$^*$          & 84.08$^*$ \\
FLUX.1 [Dev]~\cite{flux2024}             & 12.0B & Massive Cluster   & Web-Scale      & 0.660$^*$          & 83.84$^*$ \\
Janus-Pro-7B~\cite{chen2025januspro}     & 7.0B  & Large Cluster     & Multimodal Web & 0.800$^*$          & 84.19$^*$ \\
$i1$ (Open Foundation)~\cite{i1arxiv2026}& 3.0B  & TPU v5p-128 Pod   & Curated Open   & 0.840              & 86.73 \\
\midrule
\multicolumn{6}{l}{\textit{\textbf{Accessible Academic Baselines (Sub-2B Parameters, Finetuned on BLIP3o-60K$^\dagger$)}}} \\
PixNerd-XXL~\cite{wang2026pixnerd}       & 1.2B  & Multi-GPU Cluster & Pretrain + BLIP3o$^\dagger$ & 0.730$^*$          & 80.90$^*$ \\
PixelGen-XXL~\cite{ma2026pixelgen}       & 1.1B  & 8$\times$ H800 (6 Days) & 36M + BLIP3o$^\dagger$ & 0.790$^*$          & 79.36 \\
DeCo-XXL~\cite{ma2026deco}               & 1.1B  & Multi-GPU Cluster & Pretrain + BLIP3o$^\dagger$ & 0.860$^*$          & 81.40$^*$ \\
Looped-DiT B/16~\cite{chng2026looped}    & 0.26B & 8$\times$ H100 (3 Days) & CC12M + BLIP3o$^\dagger$ & \textbf{0.874}$^*$ & 87.00$^*$ \\
MiniT2I-L/16~\cite{minit2i2026}          & 0.91B & 8$\times$ H100 (27 Days)   & CC12M + BLIP3o$^\dagger$ & \textbf{0.883}$^*$ & 84.80$^*$ \\
\midrule
\multicolumn{6}{l}{\textit{\textbf{Asuka-FM (Ours: Single Consumer RTX 4090, 625K Balanced Dataset, No BLIP3o Finetuning)}}} \\
\textbf{Asuka-256 (Dense Final)}        & 0.21B & 1$\times$ RTX 4090 (68h) & 625K Open Curated & \textbf{0.588}     & \textbf{78.96} \\
\textbf{Asuka-512 (75\% Sparse)}        & 0.21B & 1$\times$ RTX 4090 (40h) & 625K Open Curated & 0.480              & 68.40 \\
\textbf{Asuka-512 (Dense FT 20k Steps)} & 0.21B & 1$\times$ RTX 4090 (60h) & 625K Open Curated & 0.575              & 79.49 \\
\textbf{Asuka-512 (Dense Final, Ep 139)}& 0.21B & 1$\times$ RTX 4090 (81h) & 625K Open Curated & \textbf{0.609}     & \textbf{78.51} \\
\bottomrule
\end{tabular}%
}
\end{table*}

\subsection{Open-Domain Benchmark Evaluation}

We evaluate Asuka-FM across two benchmarks: GenEval~\cite{ghosh2024geneval} and DPG-Bench~\cite{dpg2024} (Table~\ref{tab:benchmark_comparison}; the complete GenEval breakdown is provided in Table~\ref{tab:benchmark_comparison_full} of Appendix~\ref{sec:appendix_full_tables}). At $256 \times 256$, Asuka-FM achieves GenEval of \textbf{0.588} and DPG-Bench of \textbf{78.96}. When scaling to $512 \times 512$ under 75\% SPRINT token dropping (Phase 2A), performance temporarily drops to $0.480$ on GenEval and $68.40$ on DPG-Bench due to fine spatial binding degradation (color attribution drops from 41.75\% to 16.25\%). However, full-sequence dense refinement (Phase 2B) rapidly restores high-frequency fidelity: within 20,000 steps, DPG-Bench surges to 79.49, converging at Epoch 139 to \textbf{0.609 GenEval and 78.51 DPG-Bench}. 

\textbf{Evaluation Protocol and Checkpoint selection.} To ensure strict reproducibility, benchmarks are evaluated using a 100-step Euler ODE solver. Classifier-Free Guidance (CFG) scale is set to $w_{\text{cfg}} = 8.0$ at $256^2$ and reduced to $w_{\text{cfg}} = 6.0$ at $512^2$ to prevent contrast blowout. We explicitly select Epoch 139 over the 20k-step checkpoint despite the slight 0.98-point variance in DPG: extended training improves fine-grained multi-object binding, with color attribution rising from 0.358 to \textbf{0.450} and two-object layout rising from 0.788 to \textbf{0.811} (Table~\ref{tab:benchmark_comparison_full}).

\textbf{Contextualizing Benchmark Fine-Tuning.} As documented in prior evaluations~\cite{i1arxiv2026,wu2025openunisimplebaselineunified}, several academic baselines (e.g., Looped-DiT, MiniT2I, DeCo) report scores following fine-tuning on the \texttt{BLIP3o-60K} prompt dataset~\cite{chen2025blip3o}, which directly aligns models to GenEval syntactic patterns. In contrast, Asuka-FM is evaluated strictly as an open-domain foundation model trained on general multi-source data without benchmark-specific fine-tuning. Even without specialized prompt tuning, Asuka-FM's DPG-Bench score ($78.51\text{--}79.49$) is highly competitive with models trained on massive clusters (DeCo at 81.40, SDXL at 72.40), while outperforming SDXL on GenEval ($0.609\text{ vs. } 0.406$).

\subsection{Compute Budget Comparison}
Table~\ref{tab:hardware_budget_comparison} compares Asuka-FM against modern foundation models. While sub-2B models require multi-GPU nodes costing \$2,000--\$5,000, Asuka-FM completes its latent two-stage curriculum in 149.0 hours on a single RTX 4090 GPU (\$49.17), establishing an order-of-magnitude cost reduction. On the Pareto frontier (Figure~\ref{fig:teaser_overview}, Panel A), Asuka-FM executes only \textbf{0.0657 ZFLOPs} (56.27 EFLOPs theoretical), sitting at the extreme low-compute anchor while maintaining high fidelity.

\begin{table}[t]
\centering
\small
\caption{\textbf{Pretraining hardware and financial cost comparison across foundation T2I models.} Cloud rates: \$0.33/hr (RTX 4090), \$1.50/hr (H800), \$2.89/hr (H100), \$537.60/hr (TPU v5p-128 pod).}
\label{tab:hardware_budget_comparison}
\setlength{\tabcolsep}{5.5pt}
\renewcommand{\arraystretch}{1.04}
\begin{tabular}{lccccr}
\toprule
\textbf{Model} & \textbf{Params} & \textbf{Hardware} & \textbf{Wall Time} & \textbf{GPU-h} & \textbf{Cost (\$)} \\
\midrule
FLUX.1 [Dev]  & 12.0B & H100 SXM5 Pod & Weeks       & $>100\text{k}$ & $>\$175\text{k}$ \\
SD3 Medium    & 2.0B  & H100 SXM5 Pod & Weeks       & $>50\text{k}$ & $>\$70\text{k}$ \\
$i1$.   & 3.0B  & TPU v5p-128   & 29.5 Days   & $90.6\text{k}$ & $\approx \$380\text{k}$ \\
PixelGen-XXL  & 1.1B  & 8$\times$ H800 & 6.0 Days    & 1,152         & $\approx \$2,880$ \\
MiniT2I-B/32  & 0.26B & 8$\times$ H100 & 3.0 Days    & 576           & $\approx \$2,016$ \\
Looped-DiT    & 0.26B & 8$\times$ H100 & 3.0 Days    & 576     & $\approx \$2,016$ \\
\midrule
\textbf{Asuka-256} & \textbf{0.21B} & \textbf{1$\times$ RTX 4090} & \textbf{68.0 h} & \textbf{68.0} & \textbf{\$22.44} \\
\textbf{Asuka-512} & \textbf{0.21B} & \textbf{1$\times$ RTX 4090} & \textbf{81.0 h} & \textbf{81.0} & \textbf{\$26.73} \\
\textbf{Asuka-FM DiT Total} & \textbf{0.21B} & \textbf{1$\times$ RTX 4090} & \textbf{149.0 h} & \textbf{149.0} & \textbf{\$49.17} \\
\textbf{Asuka-FM Full (Inc. VAE)} & \textbf{0.21B} & \textbf{1$\times$ RTX 4090} & \textbf{161.2 h} & \textbf{161.2} & \textbf{\$53.20} \\
\bottomrule
\end{tabular}
\end{table}

\subsection{Hardware Profiling: RTX 4090 vs. NVIDIA A40}
\label{sec:hardware_profiling}
We profile kernel execution across consumer and datacenter silicon: RTX 4090 (24GB GDDR6X, Ada Lovelace) and NVIDIA A40 (48GB GDDR6, Ampere) under identical software stacks (Table~\ref{tab:profiling_systems_comparison}; see Table~\ref{tab:profiling_systems_comparison_full} in Appendix~\ref{sec:appendix_full_tables} for complete resolution and token sequence specifications).

\textbf{The VRAM and Peak TFLOPS Analysis.} The A40 provides double the physical VRAM (48GB vs. 24GB) and comparable rated compute ($\sim$150 TFLOPS). However, at batch size 64, the RTX 4090 completes each compiled step in $308.27\,\text{ms}$ ($84.22\,\text{TFLOPS}$, $\text{MFU}_{\text{act}} = 50.98\%$), whereas the A40 requires $607.65\,\text{ms}$ ($44.43\,\text{TFLOPS}$, $\text{MFU}_{\text{act}} = 26.90\%$). The consumer GPU executes identical workloads \textbf{$1.97\times$ faster}. Peak memory allocation reaches only $15.25\,\text{GB}$ on the RTX 4090; the extra 34.75\,GB on the A40 remains completely idle.

\textbf{Micro-Architectural Drivers.} This divergence is driven by two factors:
(i)~\textbf{12$\times$ L2 Cache Differential (72\,MB vs. 6\,MB):} Under \texttt{torch.compile}, TorchInductor fuses projections, norms, and SwiGLU gating into compound Triton kernels. On Ada Lovelace, the 72\,MB L2 cache keeps intermediate activations resident in on-chip SRAM across kernel invocations. On Ampere (6\,MB L2), batched sequence activations ($N_{\text{batch}} = B \cdot N = 16{,}384$ tokens/step) overflow, thrashing memory buses.
(ii)~\textbf{DRAM Bandwidth Starvation (1,008\,GB/s vs. 696\,GB/s):} When cache overflows, the A40 falls back to GDDR6 memory ($31\%$ slower). Memory-bound phases suffer heavy penalties: text encoder forward is $2.57\times$ slower on A40 ($99.98\,\text{ms vs. } 38.94\,\text{ms}$), and AdamW steps are $1.74\times$ slower ($36.33\,\text{ms vs. } 20.87\,\text{ms}$).

\textbf{The Online VAE Wall at 512px and 1MP.} Profiling at $512 \times 512$ on the A40 under online VAE encoding (Table~\ref{tab:profiling_systems_comparison}) shows that in-place VAE encoding consumes $2,263.52\,\text{ms}$ out of $3,808.42\,\text{ms}$ per step ($59.4\%$ of wall-clock time; see Figure~\ref{fig:vae_wall_breakdown_appendix} in Appendix~\ref{sec:appendix_full_tables} for resolution breakdown), dragging end-to-end system MFU down to \textbf{$15.27\%$}. This bottleneck intensifies at native 1-megapixel ($1024 \times 1024, N = 4096$ tokens): caching uncompressed 32-channel float16 latents for 337k samples requires over $1.2\,\text{TB}$ of storage, exceeding consumer NVMe capacities and forcing on-the-fly encoding. Profiling at micro-batch size $B=8$ on a single RTX 4090 reveals that the frozen VAE encoder consumes $164.92\,\text{ms}$ per step (\textbf{62.0\% of total step latency}, $265.44\,\text{ms}$ compiled), whereas the DiT forward and backward passes execute in $23.70\,\text{ms}$ and $44.50\,\text{ms}$ respectively ($<70\,\text{ms}$ combined). Because the frozen autoencoder runs \textbf{$2.42\times$ slower} than the generative backbone, physical hardware MFU is capped at $14.5\%\text{--}26.2\%$—establishing the primary systems motivation for our autoencoder-free latent-to-pixel adaptation strategy (preliminary 1MP SPRINT scaling diagnostics on the Danbooru corpus are deferred to Appendix~\ref{sec:appendix_1mp_scaling}).

\begin{table*}[t]
\centering
\captionsetup{font=small, labelfont=bf, skip=4pt}
\caption{\textbf{Micro-architectural profiling across hardware architectures at $256^2$ and $512^2$.} Profiled across 30 active iterations via PyTorch \texttt{FlopCounterMode} (165.2 TFLOPS peak reference, micro-batch 64 with $16{,}384\text{ tokens/step}$ at $256^2$ and $65{,}536\text{ tokens/step}$ at $512^2$). Canvas dimensions and token lengths are reported in Table~\ref{tab:profiling_systems_comparison_full} (Appendix~\ref{sec:appendix_full_tables}).}
\label{tab:profiling_systems_comparison}
\setlength{\tabcolsep}{3.4pt}
\renewcommand{\arraystretch}{1.08}
\resizebox{\linewidth}{!}{%
\begin{tabular}{lccccccccc}
\toprule
\textbf{Hardware Device} & \shortstack{\textbf{Peak VRAM}\\\textbf{Allocated}} & \multicolumn{5}{c}{\textbf{Step Latency Breakdown (ms)}} & \shortstack{\textbf{Total Step}\\\textbf{Latency}} & \shortstack{\textbf{Backbone}\\\textbf{Throughput}} & \shortstack{\textbf{MFU$_{\text{act}}$ (\%)}\\\textbf{(Isolated / Sys)}} \\
\cmidrule(lr){3-7}
& & \shortstack{\textbf{Data /}\\\textbf{IO}} & \shortstack{\textbf{Online}\\\textbf{VAE}} & \shortstack{\textbf{Text}\\\textbf{Enc.}} & \shortstack{\textbf{Backbone}\\\textbf{(F+B)}} & \shortstack{\textbf{Opt. \&}\\\textbf{Clip}} & & & \\
\midrule
\multicolumn{10}{l}{\textit{\textbf{Production Latent Pretraining: Kernel-Fused Tensor Saturation ($256 \times 256$, In-RAM Latents)}}} \\
NVIDIA A40         & 13,254.5 MB          & 1.44 ms           & --                & 99.98 ms          & 469.88 ms             & 36.33 ms             & 607.65 ms          & 44.43 TFLOPS        & 26.90\% / 31.71\% \\
\textbf{RTX 4090}  & \textbf{15,249.5 MB} & \textbf{0.27 ms}  & \textbf{0.28 ms}  & \textbf{38.94 ms} & \textbf{247.91 ms}    & \textbf{20.87 ms}    & \textbf{308.27 ms} & \textbf{84.22 TFLOPS} & \textbf{50.98\% / 63.25\%} \\
\midrule
\multicolumn{10}{l}{\textit{\textbf{Online Autoencoder Wall ($512 \times 512$, FLUX.2 VAE On-the-Fly Encoding, Dense Tokens)}}} \\
NVIDIA A40         & 35,201.0 MB          & 0.33 ms           & 2,263.52 ms       & 111.34 ms         & 1,396.95 ms           & 36.28 ms             & 3,808.42 ms        & 51.39 TFLOPS        & 31.11\% / \textbf{15.27\%} \\
\bottomrule
\end{tabular}%
}
\end{table*}

\begin{table*}[t]
\centering
\captionsetup{font=small, labelfont=bf, skip=4pt}
\caption{\textbf{Latent-to-Pixel adaptation progression and compute metrics of Asuka-FM on a single consumer NVIDIA RTX 4090 GPU (24GB).} Adapts the converged 512px latent foundation model to direct RGB pixel space using a 10.09M convolutional Patch Detailer Head (PDH) under native $v$-prediction at effective batch size 128. Hardware costs evaluated at \$0.33/GPU-hour for RTX 4090 and \$0.40/GPU-hour for A40 (Phase 0). Unabridged telemetry and hyperparameter matrices are detailed in Table~\ref{tab:unabridged_pixel_lifecycle_appendix} in Appendix~\ref{sec:appendix_pixel_benchmarks}.}
\label{tab:pixel_training_stages}
\setlength{\tabcolsep}{4.2pt}
\renewcommand{\arraystretch}{1.08}
\resizebox{\linewidth}{!}{%
\begin{tabular}{lccccccccr}
\toprule
\textbf{Stage / Phase} & \shortstack{\textbf{Res.}\\\textbf{($X^2$)}} & \shortstack{\textbf{Drop}\\\textbf{($r$)}} & \textbf{Samples} 
& \shortstack{\textbf{Throughput}\\\textbf{(img/s)}} & \shortstack{\textbf{MFU (\%)}\\\textbf{Theo / Act}$^\dagger$} 
& \shortstack{\textbf{Compute}\\\textbf{(EFLOPs)}$^*$} & \shortstack{\textbf{Wall}\\\textbf{Time}} & \shortstack{\textbf{Cost}\\\textbf{(\$)}} \\
\midrule
Phase 0: Offline Synth Generation (A40)$^\ddagger$ & $512^2$ & --   & 0.10M & 1.85 img/s  & --              & --    & 15.0 h & \$6.00 \\
Stage 1: Decoupled PDH Warmup (Synth)              & $512^2$ & 0.75 & 0.50M & 61.8 img/s  & 62.4\% / 72.8\% & 0.52  & 2.25 h & \$0.74 \\
Stage 2A: Joint SPRINT Sparse (Synth)              & $512^2$ & 0.75 & 2.50M & 57.7 img/s  & 58.1\% / 67.8\% & 2.60  & 12.0 h & \$3.98 \\
Stage 2B: Full-Token Dense Refine (Synth)          & $512^2$ & 0.00 & 1.50M & 44.4 img/s  & 59.8\% / 69.8\% & 2.71  & 9.4 h  & \$3.10 \\
Stage 3: Real-Data SFT Refinement (Real)           & $512^2$ & 0.00 & 5.16M & 42.1 img/s  & 56.7\% / 66.2\% & 9.32  & 34.0 h & \$11.22 \\
\midrule
\multicolumn{3}{l}{\textbf{Cumulative Pixel DiT Adaptation Total (Stages 1--3)}} & \textbf{9.66M} & \textbf{46.5 img/s} & \textbf{57.8\% / 67.5\%} & \textbf{15.15} & \textbf{57.65 h} & \textbf{\$19.04} \\
\multicolumn{3}{l}{\textbf{Cumulative Grand Total (Including Offline Synth Dataset Generation)}} & -- & -- & -- & \textbf{15.15} & \textbf{72.65 h} & \textbf{\$25.04} \\
\midrule
\multicolumn{9}{l}{\textit{\textbf{Progressive Token Compression via Large Patch Adaptation ($P=16 \to 32$)}}} \\
ps32-adapt16 (Large Patch Engine)                  & $512^2$ & 0.00 & 9.92M & 38.3 img/s  & 54.2\% / 63.3\% & 17.92 & 72.0 h & \$23.76 \\
\bottomrule
\end{tabular}%
}
\vspace{2pt}
\begin{flushleft}
\footnotesize
$^\dagger\text{MFU}_{\text{act}}$ evaluates actual tensor ops ($7PD$ per sample, for parameter count $P_{\text{param}}$ and tokens per sample $D_{\text{tok}}$) under selective SwiGLU activation checkpointing.\\
$^*1\text{ EFLOP} = 10^{18}\text{ FLOPs}$. Cumulative theoretical algorithmic compute for Stages 1--3 is $12.98\text{ EFLOPs}$.\\
$^\ddagger$One-time offline generation of 100k synthetic anchor samples on an NVIDIA A40 GPU (\$0.40/h).
\end{flushleft}
\end{table*}

\subsection{Latent-to-Pixel Adaptation at 512px Resolution}
\label{sec:pixel_adaptation_experiments}
Following our pretraining recipe, we adapt the converged 512px latent model directly to raw RGB pixel space using the 10.09M convolutional Patch Detailer Head (PDH) across three progressive stages (Table~\ref{tab:pixel_training_stages}):
\begin{itemize}
    \item \textbf{Phase 0 (Offline Synthetic Dataset Generation):} Using our converged Asuka-512 latent foundation checkpoint, we generate a support corpus of 100,000 synthetic images across diverse prompts. On an NVIDIA A40 GPU, this offline generation phase completed in 15.0 hours at a sustained rate of $1.85\,\text{img/s}$ (\$6.00 compute cost).
    \item \textbf{Stage 1 (Decoupled PDH Warmup):} To suppress initial gradient shocks from randomly initialized convolutional filters, the 206M DiT backbone is frozen. Only the input patch embedder (\texttt{x\_embedder}) and the PDH convolutional U-Net are trained on 100k synthetic images for 5 epochs ($0.50\text{M}$ samples total) under 75\% SPRINT token dropping ($r=0.75$, peak lr $10^{-4}$ with 25\% linear warmup). This phase converged in 2.25 hours (2h 14m 45s) on an RTX 4090, sustaining $61.8\,\text{img/s}$ ($\text{MFU}_{\text{act}} = 72.8\%$, \$0.74 cost).
    \item \textbf{Stage 2A (Joint Synthetic SPRINT Adaptation):} The full network is unfrozen. Training proceeds for 25 epochs ($2.50\text{M}$ samples) on synthetic data under $r=0.75$ token dropping (DiT backbone learning-rate multiplier $1.25\times$, base lr $6\times 10^{-5}$ with 10\% warmup), completing in 12.0 hours (12h 02m 42s) at $57.7\,\text{img/s}$ (\$3.98 cost).
    \item \textbf{Stage 2B (Full-Token Dense Refinement):} The network switches to dense tokens ($r=0.00$, micro-batch size 8, effective batch size 128) for 15 epochs ($1.50\text{M}$ samples) with a 10\% WSD linear decay schedule. This phase converged in 9.4 hours (9h 23m 31s) at $44.4\,\text{img/s}$ (\$3.10 cost).
    \item \textbf{Stage 3 (Real-Data SFT Refinement):} Synthetic data is discarded. The model is fine-tuned strictly on full tokens ($r=0.00$) over a curated high-aesthetic real corpus of 396,635 samples ($\min(H, W) \ge 1024\text{px}$) extracted from the $i1$ dataset. Training for 13 epochs ($5.16\text{M}$ samples) completed in 34.0 hours on a single RTX 4090 at $42.1\,\text{img/s}$ (\$11.22 cost).
\end{itemize}

Cumulatively, the three DiT pixel adaptation stages executed 15.15 EFLOPs over 57.65 wall-clock hours (\$19.04 compute cost) at an average sustained utilization of $\text{MFU}_{\text{act}} = 67.5\%$. Including offline synthetic dataset generation, the entire latent-to-pixel adaptation pipeline converges in 72.65 GPU hours (\$25.04).

\begin{table*}[t]
\centering
\captionsetup{font=small, labelfont=bf, skip=4pt}
\caption{\textbf{Quantitative benchmark summary of Latent-to-Pixel adaptation on GenEval and DPG-Bench.} Evaluated at $512^2$ canvas resolution using a 100-step Euler ODE solver with CFG scale $w_{\text{cfg}}=6.0$. Sampling throughput and per-image latency profiled on a single NVIDIA A40 GPU at batch size 32. Full task-by-task GenEval breakdowns are provided in Table~\ref{tab:pixel_adaptation_benchmarks_extended} in Appendix~\ref{sec:appendix_pixel_benchmarks}.}
\label{tab:pixel_benchmarks_summary}
\setlength{\tabcolsep}{5.2pt}
\renewcommand{\arraystretch}{1.08}
\resizebox{\linewidth}{!}{%
\begin{tabular}{lcccccc}
\toprule
\textbf{Configuration / Stage} & \shortstack{\textbf{Patch}\\\textbf{Size ($P$)}} & \shortstack{\textbf{Target}\\\textbf{Objective}} & \shortstack{\textbf{Training}\\\textbf{Schedule}} & \shortstack{\textbf{GenEval}\\\textbf{Total $\uparrow$}} & \shortstack{\textbf{DPG-Bench}\\\textbf{Total $\uparrow$}} & \shortstack{\textbf{Sampling Latency / Img}\\\textbf{(20-step / 30-step)}} \\
\midrule
\multicolumn{7}{l}{\textit{\textbf{Baseline Latent Foundation Model (Pretraining Anchor)}}} \\
Asuka-512 (Base Latent)                   & $P=16$ & $v$-pred   & 139 Epochs (81.0h) & \textbf{0.609} & \textbf{78.51} & 0.250 s / 0.374 s \\
\midrule
\multicolumn{7}{l}{\textit{\textbf{Latent-to-Pixel Adaptation Progression (DiP Architecture, Zero VAE Decode)}}} \\
Stage 2A: Joint SPRINT Sparse ($r=0.75$)   & $P=16$ & $v$-pred   & 25 Epochs (12.0h)  & 0.465          & 69.63          & 0.329 s / 0.493 s \\
Stage 2B: Full-Token Refine (Synth, $v$-pred)& $P=16$ & $v$-pred   & 15 Epochs (9.4h)   & 0.505          & 72.35          & 0.329 s / 0.493 s \\
Stage 2B: Full-Token Refine (Synth, $x_0$-pred)& $P=16$ & $x_0$-pred & 15 Epochs (9.4h)   & 0.477          & 71.49          & 0.329 s / 0.493 s \\
Stage 3 SFT ($x_0$-pred Ablation)         & $P=16$ & $x_0$-pred & 13 Epochs (34.0h)  & 0.545          & \textbf{76.32} & 0.329 s / 0.493 s \\
\textbf{Stage 3 SFT ($v$-pred, Final Pixel DiT)} & \textbf{$P=16$} & \textbf{$v$-pred} & \textbf{13 Epochs (34.0h)} & \textbf{0.568} & \textbf{75.31} & 0.329 s / 0.493 s \\
\midrule
\multicolumn{7}{l}{\textit{\textbf{Progressive Token Compression via Large Patch Adaptation ($P=16 \to 32$)}}} \\
ps32-adapt16 (Large Patch Pixel Engine)   & $P=32$ & $v$-pred   & +25 Epochs (72.0h) & 0.504          & 69.87          & \textbf{0.187 s / 0.277 s} \\
\bottomrule
\end{tabular}%
}
\end{table*}

\subsection{Empirical Analysis of Latent-to-Pixel Mechanics}

\paragraph{Quantitative Convergence and Benchmark Recovery.}
As documented in Table~\ref{tab:pixel_benchmarks_summary}, the final converged Pixel DiT ($P=16$, Stage 3 SFT) achieves \textbf{75.31 DPG-Bench} and \textbf{0.568 GenEval}, recovering over \textbf{95\%} of the generative compositional fidelity of the source latent foundation model (78.51 DPG / 0.609 GenEval) while permanently bypassing the autoencoder decoding wall.

\paragraph{$v$-Prediction vs. $x_0$-Prediction.}
To rigorously evaluate prediction objectives under identical training compute, we benchmarked $x_0$-prediction against $v$-prediction across Stages 2B and 3. As detailed in Table~\ref{tab:pixel_benchmarks_summary}, while $x_0$-prediction reached a marginally higher overall DPG score (76.32 vs. 75.31), $v$-prediction delivered strictly superior performance on GenEval (0.568 vs. 0.545) and fine-grained color attribution (0.420 vs. 0.348; Table~\ref{tab:pixel_adaptation_benchmarks_extended}). 

Critically, converting predicted clean states $\hat{x}_0$ into velocity targets $v_t = (x_t - \hat{x}_0)/t$ introduces a fatal singularity as $t \to 0$ in \texttt{bfloat16}, requiring dynamic percentile clipping that flattens dynamic range and induces chromatic blowout. In contrast, native $v$-prediction maintained smooth optimization trajectories with zero activation clipping, completely avoiding color desaturation.

\paragraph{Progressive Token Compression ($P=16 \to 32$).}
To accelerate multi-step pixel inference, we evaluated progressive patch widening to $P=32$ ($4\times$ token compression, sequence length $N=256$ at $512^2$; 25 epochs, 72.0h). While compositional binding declines under the larger patch stride (DPG-Bench drops to 69.87, GenEval to 0.504), sampling latency at $512^2$ on an NVIDIA A40 drops to $0.187\,\text{s}$ (20 steps) and $0.277\,\text{s}$ (30 steps, a \textbf{$1.78\times$ speedup} over $P=16$), establishing an ultra-fast sampling engine for compute-critical applications.

\subsection{Inference Profiling: The Amortization Asymmetry}
\label{sec:inference_profiling}
A common assumption in pixel diffusion literature~\cite{dip2026} is that eliminating the VAE decoder accelerates inference. Profiling across $N \in \{20, 30\}$ sampling steps on an NVIDIA A40 (Table~\ref{tab:sampling_profiling_appendix}) reveals an \textbf{Amortization Asymmetry}: for standard multi-step sampling ($N \in \{20, 30\}$), pixel diffusion is actually $9\%\text{--}31\%$ slower than latent diffusion. At $1024^2$ (batch size $B=16$), generating 30-step trajectories requires $46.43\,\text{s}$ in latent mode versus $53.28\,\text{s}$ in pixel mode ($14.8\%$ slower).

\textbf{Deconstructing the Asymmetry.} Total sampling latency over $N$ ODE steps decomposes into per-step evaluation latency $t_{\text{step}}$ and fixed terminal decoding overhead $t_{\text{fixed}}$: $\text{Total Time}(N) = N \cdot t_{\text{step}} + t_{\text{fixed}}$. In latent sampling, terminal VAE decoding incurs a one-time fixed penalty ($t_{\text{fixed, latent}} = 4.82\,\text{s}$ at $1024^2$), while per-step latency remains low ($t_{\text{step, latent}} = 1.387\,\text{s}$ per batch step). In pixel sampling, while the VAE overhead is eliminated ($t_{\text{fixed, pixel}} = 0.09\,\text{s} \approx 0$), the convolutional PDH head must evaluate \textit{recurrently at every step} (and twice per step under CFG), increasing per-step latency to $t_{\text{step, pixel}} = 1.773\,\text{s}$ (+27.8\%). Across 30 steps, this recurrent overhead accumulates to $11.58\,\text{s}$, completely eclipsing the $4.82\,\text{s}$ VAE decode savings.

\begin{figure}[t]
\centering
\captionsetup{font=small, labelfont=bf, skip=4pt}
\includegraphics[width=0.60\textwidth]{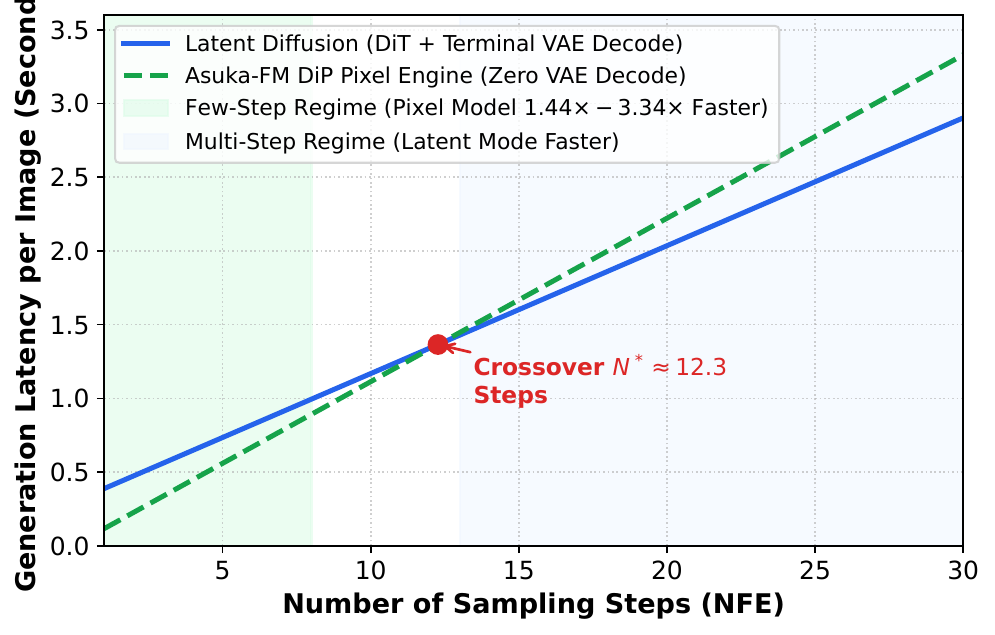}
\caption{\textbf{Sampling latency trade-off at 1MP: Latent DiT vs. DiP Pixel Engine on NVIDIA A40.} Pixel sampling is faster in few-step regimes ($N \le 8$, $3.34\times$ speedup at 1 step), while latent sampling is faster for multi-step ancestral sampling ($N \ge 13$).}
\label{fig:sampling_tradeoff}
\end{figure}

\textbf{Crossover Dynamics and Operational Regimes.} Equating both linear trajectories yields an empirical break-even crossover threshold of $N^* \approx 12.25$ sampling steps (formal derivation in Appendix~\ref{sec:appendix_amortization_math}), cleanly partitioning operational utility (Figure~\ref{fig:sampling_tradeoff}). For \textit{multi-step regimes} ($N \ge 13$ steps), latent sampling is strictly faster because the $O(1)$ VAE decode is amortized across many steps. Conversely, in \textit{few-step regimes} ($N \le 8$ steps), pixel diffusion is decisively superior: under 4-step distilled sampling (DMD2~\cite{yin2024dmd2}), pixel latency is $7.18\,\text{s}$ ($0.449\,\text{s/img}$) versus $10.37\,\text{s}$ ($0.648\,\text{s/img}$) for latent mode (a \textbf{$1.44\times$ speedup}). At $N=1$ step, pixel mode achieves a \textbf{$3.34\times$ acceleration}, establishing pixel models as natural engines for distilled few-step rollout.

\subsection{Manifold Support: Square Synthetics vs. Multi-Aspect Corpora}
\label{sec:manifold_support}
An empirical discovery is the divergence in camera controllability across training distributions. Models trained on synthetic open-domain datasets (e.g., FLUX-Reason-6M, Midjourney-v6) exhibit geometric warping and background collapse on some prompts during coordinate manipulation. Synthetic datasets are mainly generated as centered $1024 \times 1024$ squares. Extracting shifted square crops from square-native images forcibly clips through canonical focal points, teaching the model that off-center coordinate queries correlate with truncated subjects. 

In contrast, our curated Danbooru corpus naturally spans non-square aspect ratios ($1:2$ portrait pin-ups to $21:9$ panoramic spreads). Performing shifted cropping on raw rectangular illustrations forces the continuous RoPE manifold to encounter the entire coordinate spectrum $[-r_H, r_H] \times [-r_W, r_W]$ over diverse, naturally framed focal points and perspective depths. Consequently, the anime-specialized model establishes genuine spatial invariance, unlocking robust zero-shot virtual camera manipulation without auxiliary conditioning tokens.

\section{Discussion, Limitations, and Future Work}
\label{sec:discussion}

\textbf{Limitations.} While Asuka-FM achieves competitive synthesis on a single consumer GPU, several limitations remain: (i) Pretraining was conducted on 625,000 samples; multi-billion parameter models trained on hundreds of millions of images remain superior on complex crowd scenes and long-tail concepts. (ii) Although our corpus includes TextAtlas5M~\cite{csu2024textatlas}, the 206M backbone does not consistently render multi-word text strings. (iii) In pixel space, multi-step sampling ($N \ge 13$ steps) without patch enlargement ($P=16$) exhibits higher latency than latent diffusion due to recurrent PDH evaluation.

\textbf{Future Work.} Immediate research will expand along two directions:
(i) \textbf{Native 1MP Scaling Curriculum:} Building on our preliminary 1MP profiling (Appendix~\ref{sec:appendix_1mp_scaling}).
(ii) \textbf{Zero-Decode Downstream Post-Training (Flow-GRPO):} In Flow-GRPO reinforcement learning~\cite{liu2025flowgrpo}, generating $G=12$ rollouts per prompt requires 12 independent VAE decodes in latent mode ($>60\%$ of GPU memory). By emitting direct RGB tensors, our pixel model eliminates this decoding bottleneck, enabling scalable online RL on consumer workstations.

\section{Conclusion}
\label{sec:conclusion}

Asuka-FM demonstrates that high-resolution text-to-image foundation models do not fundamentally require datacenter-scale compute clusters. By synthesizing an AdaLN-free dual-stream backbone with prepended temporal conditioning, structured sequence sparsification via SPRINT, continuous aspect-normalized RoPE with shifted square crop training, and decoupled latent-to-pixel adaptation via a lightweight Patch Detailer Head, Asuka-FM successfully trains a 206M latent generative foundation model from scratch on a \textbf{single consumer NVIDIA RTX 4090 GPU in 161.2 wall-clock hours, executing 65.65 ExaFLOPs for only \$53.2 in compute}.

Operating on the Pareto frontier of denoiser compute ($0.0657\text{ ZFLOPs}$), Asuka-FM achieves a GenEval score of \textbf{0.609} and DPG-Bench score of \textbf{78.51} at 512px without benchmark-specific fine-tuning. Micro-architectural profiling reveals an Amortization Asymmetry: while pixel adaptation eliminates the online VAE wall during pretraining and delivers up to $3.34\times$ speedups in few-step regimes ($N \le 4$ sampling steps), latent sampling remains faster for multi-step trajectories ($N \ge 13$ steps). Asuka-FM provides an empirical and systems foundation for accessible, low-budget generative foundation modeling across academic and independent research laboratories.

\newpage
\printbibliography
\section*{Acknowledgments}
We gratefully acknowledge the Department of Signal Theory and Communications at \textbf{Universidad Carlos III de Madrid (UC3M)} for providing the dedicated GPU workstations and computing infrastructure that made these extensive pretraining experiments possible.

Víctor Rodrigo González and Alejandro Lancho have received funding from the Ministerio de Ciencia, Innovación y Universidades, Spain, under Grant Agreement No.~PID2023-148856OA-I00 (funded by MICIU/AEI/10.13039/501100011033 and by ERDF/UE). Alejandro Lancho and Pablo M. Olmos have received funding from the Comunidad de Madrid through the IDEA-CM project (Grant Agreement No.~TEC-2024/COM-89).

Additionally, Alejandro Lancho has received funding from the Comunidad de Madrid through the CARE-CM project (Grant Agreement No.~SYG-2024/COM-870). Ignacio Peis acknowledges support from the Danish Data Science Academy, which is funded by the Novo Nordisk Foundation (NNF21SA0069429), and from the “la Caixa” Foundation (ID 100010434) through the Junior Leader Incoming fellowship LCF/BQ/PI26/12120010. Pablo M. Olmos has additionally been supported by the Ministerio de Ciencia, Innovación y Universidades, Spain, under Grant PID2024-157856NB-I00 (CARTESIAN), funded by MCIN/AEI/10.13039/501100011033 and, as appropriate, by “ERDF A way of making Europe”; by the ELLIS Unit Madrid (European Laboratory for Learning and Intelligent Systems); and by the Comunidad de Madrid under grant IND2024/TIC-34728.

Our work stands on the shoulders of the open machine learning research community. In particular, we express our sincere appreciation to the authors of:
\begin{itemize}
    \item \textbf{HDM} (\cite{hdm2025}), for pioneering consumer-scale anime foundation pretraining and demonstrating the viability of shifted square cropping and continuous RoPE.
    \item \textbf{SPRINT} (\cite{sprint2026}), for the sparse-dense residual fusion paradigm that fundamentally slashed our attention sequence complexity.
    \item \textbf{$i1$} (\cite{i1arxiv2026}), for establishing an exemplary open recipe for multi-modal diffusion transformers and demonstrating that explicit AdaLN modulation can be omitted, and for publicly releasing the synthetic captions used for training.
    \item \textbf{DiP} (\cite{dip2026}) and the empirical pixel diffusion studies by \cite{chen2026l2punlockinglatentpotential, jiang2026empirical}, for illuminating the dynamics of the Patch Detailer Head and pixel-space velocity trajectories.
\end{itemize}
We note that while we did not have direct communication or collaboration with the authors of these prior works, our system-architectural framework was made possible through their transparent public papers, released checkpoints, and open codebases.

We also extend our gratitude to the broader anime and open-source community, particularly the crowdsourced annotators and maintainers of the Danbooru platform \cite{danbooru2023} whose decades of detailed image tagging provided the rich semantic metadata necessary for concept learning, as well as the maintainers of \texttt{bitsandbytes}, FlashAttention \cite{dao2023flashattn2}, and PyTorch.

Finally, the project \textit{Asuka-FM} is dedicated to Asuka Souryuu Langley from \textit{Neon Genesis Evangelion}, whose unyielding determination inspired our resolve through countless errors and pretraining milestones.

\newpage
\appendix

\section{Extended Benchmark Breakdowns and Systems Telemetry}
\label{sec:appendix_full_tables}

This section provides the complete, unabridged numerical tables corresponding to the condensed performance summaries in the main text.

\subsection{Full Pretraining Lifecycle and Token Telemetry}
Table~\ref{tab:training_stages_full} provides the unabridged pretraining lifecycle metrics of Asuka-FM across both square training resolutions $X \times X$ ($256^2$ and $512^2$) and SPRINT token drop ratios $r$, detailing per-stage batch sizes, epoch allocations, cumulative token sequences, and physical hardware throughputs. As discussed in Section~\ref{sec:systems_engineering}, the token processing rate demonstrates the arithmetic efficiency of SPRINT: by processing only $25\%$ of visual tokens in the middle blocks ($r=0.75$), the system achieves $99.2\text{k tok/s}$ in Phase 1A and $78.1\text{k tok/s}$ in Phase 2A, substantially mitigating memory bandwidth saturation on consumer-grade GDDR6X buses.

\begin{table*}[!t]
\centering
\small
\caption{\textbf{Comprehensive pretraining progression and compute metrics of Asuka-FM on a single consumer NVIDIA RTX 4090 GPU (24GB).} Unabridged reference for Table~\ref{tab:training_stages}. Sequence lengths account for visual image tokens plus context ($227\text{ text} + 1\text{ time} = 228\text{ tokens}$). $\text{MFU}_{\text{act}}$ evaluates actual tensor operations on silicon under selective SwiGLU activation checkpointing ($7PD$ work factor, where $P$ denotes active model parameters and $D$ denotes tokens per sample); $\text{MFU}_{\text{theo}}$ denotes standard theoretical algorithmic utilization ($6PD$ factor). Hardware costs evaluated at \$0.33/GPU-hour.}
\label{tab:training_stages_full}
\setlength{\tabcolsep}{2.8pt}
\renewcommand{\arraystretch}{1.08}
\resizebox{\textwidth}{!}{%
\begin{tabular}{lcccccccccccr}
\toprule
\shortstack{\textbf{Stage /}\\\textbf{Phase}} & 
\shortstack{\textbf{Res.}\\\textbf{($X^2$)}} & 
\shortstack{\textbf{Batch}\\\textbf{Size}} & 
\shortstack{\textbf{Drop}\\\textbf{($r$)}} & 
\textbf{Epochs} & 
\textbf{Samples} & 
\shortstack{\textbf{Avg}\\\textbf{Tokens}} & 
\shortstack{\textbf{Throughput}\\\textbf{(img/s)}} & 
\shortstack{\textbf{Token}\\\textbf{Rate (tok/s)}} & 
\shortstack{\textbf{MFU (\%)}\\\textbf{Theo / Act}$^\dagger$} & 
\shortstack{\textbf{Compute}\\\textbf{(EFLOPs)}$^*$} & 
\shortstack{\textbf{Wall}\\\textbf{Time}} & 
\shortstack{\textbf{Cost}\\\textbf{(\$)}} \\
\midrule
\multicolumn{13}{l}{\textit{\textbf{Stage 1: Base Semantic Pretraining ($256 \times 256$, Latent Patch Grid $16 \times 16 = 256$ Image Tokens)}}} \\
Offline VAE Encoding$^\ddagger$ & $256^2$ & 64 & -- & -- & 0.625M & -- & 45.2 & -- & -- & -- & 3.8 h & \$1.27 \\
1A: SPRINT Sparse & $256^2$ & 256 & 0.75 & 75 & 46.88M & 351.1 tok & 282.6 & 99.2k & 74.4\% / 86.8\% & 23.74 & 47.0 h & \$15.51 \\
1B: Dense Refine  & $256^2$ & 256 & 0.00 & 25 & 12.50M & 484.0 tok & 172.5 & 83.5k & 62.9\% / 73.4\% & 8.73  & 21.0 h & \$6.93 \\
\midrule
\multicolumn{13}{l}{\textit{\textbf{Stage 2: High-Resolution Adaptation ($512 \times 512$, Latent Patch Grid $32 \times 32 = 1,024$ Image Tokens)}}} \\
Offline VAE Encoding$^\ddagger$ & $512^2$ & 32 & -- & -- & 0.625M & -- & 20.6 & -- & -- & -- & 8.4 h & \$2.78 \\
2A: High-Res Sparse & $512^2$ & 256 & 0.75 & 25 & 15.63M & 720.3 tok & 108.5 & 78.1k & 58.5\% / 68.3\% & 16.24 & 40.0 h & \$13.20 \\
2B: Dense Refine    & $512^2$ & 256 & 0.00 & 15 & 9.38M  & 1,252.0 tok & 64.8  & 81.1k & 61.0\% / 71.2\% & 16.94 & 41.0 h & \$13.53 \\
\midrule
\multicolumn{4}{l}{\textbf{Cumulative DiT Training Total (Stages 1 + 2)}} & \textbf{140} & \textbf{84.38M} & -- & -- & -- & \textbf{64.8\% / 75.6\%} & \textbf{65.65} & \textbf{149.0 h} & \textbf{\$49.17} \\
\multicolumn{4}{l}{\textbf{Cumulative Grand Total (Including Offline VAE Pre-Encoding)}} & -- & -- & -- & -- & -- & -- & \textbf{65.65} & \textbf{161.2 h} & \textbf{\$53.20} \\
\bottomrule
\multicolumn{13}{l}{\small $^\dagger\text{MFU}_{\text{act}}$ evaluates actual tensor ops ($7PD$ per sample, for active parameter count $P$ and tokens per sample $D$) under SwiGLU recomputation; $\text{MFU}_{\text{theo}}$ denotes standard theoretical utilization ($6PD$).} \\
\multicolumn{13}{l}{\small $^*1\text{ EFLOP} = 10^{18}\text{ FLOPs} = 10^3\text{ PFLOPs} = 10^{-3}\text{ ZFLOPs}$. Cumulative theoretical algorithmic compute is $56.27\text{ EFLOPs}$.} \\
\multicolumn{13}{l}{\small $^\ddagger$One-time offline encoding using FLUX.2 VAE ($C=32, f=8$) permanently removing autoencoder overhead from generative pretraining.}
\end{tabular}%
}
\end{table*}

\subsection{Fine-Grained GenEval Task Breakdown}
Table~\ref{tab:benchmark_comparison_full} presents the complete compositional task breakdown on GenEval~\cite{ghosh2024geneval} alongside DPG-Bench~\cite{dpg2024}. The individual columns evaluate basic object synthesis (\textit{Single Object}), relational spatial semantics (\textit{Two Objects}, \textit{Position}), numerical consistency (\textit{Counting}), and chromatic attribute binding (\textit{Colors}, \textit{Color Attribution}). 

Crucially, the breakdown contextualizes the dynamics of SPRINT during pretraining. In Phase 2A ($75\%$ token drop), the model maintains strong single-object recognition ($0.963$) and two-object layout ($0.639$), but suffers a temporary drop in color-attribute binding ($0.163$). Once terminal dense refinement is initiated in Stage 2B, full spatial cross-attention across all $1{,}024$ tokens recovers color attribution to $0.450$, directly proving that macro-semantics can be learned sparsely while fine attribute binding requires dense token interaction.

\begin{table*}[!t]
\centering
\small
\caption{\textbf{Comprehensive quantitative comparison on representative text-to-image benchmarks (GenEval and DPG-Bench) with full task breakdown.} Unabridged reference for Table~\ref{tab:benchmark_comparison}. Results marked with ($^*$) denote numbers reported in prior literature. Models marked with $^\dagger$ indicate fine-tuning on the \texttt{BLIP3o-60K} instruction dataset. Asuka-FM is trained from scratch on open-source data without benchmark-specific fine-tuning.}
\label{tab:benchmark_comparison_full}
\setlength{\tabcolsep}{3.2pt}
\renewcommand{\arraystretch}{1.08}
\resizebox{\textwidth}{!}{%
\begin{tabular}{lccccccccccc}
\toprule
\textbf{Model} & \textbf{Params} & 
\shortstack{\textbf{Hardware}\\\textbf{Budget}} & 
\shortstack{\textbf{Training}\\\textbf{Data}} & 
\multicolumn{6}{c}{\textbf{GenEval Task Breakdown}} & 
\shortstack{\textbf{GenEval}\\\textbf{Overall $\uparrow$}} & 
\shortstack{\textbf{DPG-Bench}\\\textbf{Total Score $\uparrow$}} \\
\cmidrule(lr){5-10}
& & & & \textbf{Single} & \textbf{Two-Obj} & \textbf{Count} & \textbf{Colors} & \textbf{Position} & \shortstack{\textbf{Color-}\\\textbf{Attr}} & & \\
\midrule
\multicolumn{12}{l}{\textit{\textbf{Industrial Foundation Models (Multi-Billion Parameters, Datacenter Compute)}}} \\
SDXL Base~\cite{sdxl2024}                & 2.6B  & Hundreds of A100s & Millions (Web) & --    & --    & --    & --    & --    & --    & 0.406$^*$          & 72.40$^*$ \\
SD3 Medium~\cite{esser2024scaling}       & 2.0B  & Massive Cluster   & Millions (Web) & 0.98  & 0.84  & 0.66  & 0.74  & 0.40  & 0.43  & 0.620$^*$          & 84.08$^*$ \\
FLUX.1 [Dev]~\cite{flux2024}             & 12.0B & Massive Cluster   & Web-Scale      & 0.99  & 0.81  & 0.79  & 0.74  & 0.20  & 0.47  & 0.660$^*$          & 83.84$^*$ \\
Janus-Pro-7B~\cite{chen2025januspro}     & 7.0B  & Large Cluster     & Multimodal Web & --    & --    & --    & --    & --    & --    & 0.800$^*$          & 84.19$^*$ \\
$i1$ (Open Foundation)~\cite{i1arxiv2026}& 3.0B  & TPU v5p-128 Pod   & Curated Open   & --    & --    & --    & --    & --    & --    & 0.840              & 86.73 \\
\midrule
\multicolumn{12}{l}{\textit{\textbf{Accessible Academic Baselines (Sub-2B Parameters, Finetuned on BLIP3o-60K$^\dagger$)}}} \\
PixNerd-XXL~\cite{wang2026pixnerd}       & 1.2B  & Multi-GPU Cluster & Pretrain + BLIP3o$^\dagger$ & 0.97  & 0.86  & 0.44  & 0.83  & 0.71  & 0.53  & 0.730$^*$          & 80.90$^*$ \\
PixelGen-XXL~\cite{ma2026pixelgen}       & 1.1B  & 8$\times$ H800 (6 Days) & 36M + BLIP3o$^\dagger$ & 0.99  & 0.88  & 0.59  & 0.90  & 0.70  & 0.70  & 0.790$^*$          & 79.36 \\
DeCo-XXL~\cite{ma2026deco}               & 1.1B  & Multi-GPU Cluster & Pretrain + BLIP3o$^\dagger$ & --    & --    & --    & --    & --    & --    & 0.860$^*$          & 81.40$^*$ \\
Looped-DiT B/16~\cite{chng2026looped}    & 0.26B & 8$\times$ H100 (3 Days) & CC12M + BLIP3o$^\dagger$ & --    & --    & --    & --    & --    & --    & \textbf{0.874}$^*$ & 87.00$^*$ \\
MiniT2I-L/16~\cite{minit2i2026}          & 0.91B & 8$\times$ H100 (27 Days)   & CC12M + BLIP3o$^\dagger$ & --    & --    & --    & --    & --    & --    & \textbf{0.883}$^*$ & 84.80$^*$ \\
\midrule
\multicolumn{12}{l}{\textit{\textbf{Asuka-FM (Ours: Single Consumer RTX 4090, 625K Balanced Dataset, No BLIP3o Finetuning)}}} \\
\textbf{Asuka-256 (Dense Final)}        & 0.21B & 1$\times$ RTX 4090 (68h) & 625K Open Curated & 0.950 & 0.750 & 0.319 & 0.782 & 0.308 & 0.418 & \textbf{0.588}     & \textbf{78.96} \\
\textbf{Asuka-512 (75\% Sparse)}        & 0.21B & 1$\times$ RTX 4090 (40h) & 625K Open Curated & 0.963 & 0.639 & 0.253 & 0.673 & 0.190 & 0.163 & 0.480              & 68.40 \\
\textbf{Asuka-512 (Dense FT 20k Steps)} & 0.21B & 1$\times$ RTX 4090 (60h) & 625K Open Curated & 0.953 & 0.788 & 0.331 & 0.715 & 0.305 & 0.358 & 0.575              & 79.49 \\
\textbf{Asuka-512 (Dense Final, Ep 139)}& 0.21B & 1$\times$ RTX 4090 (81h) & 625K Open Curated & 0.950 & 0.811 & 0.347 & 0.817 & 0.278 & 0.450 & \textbf{0.609}     & \textbf{78.51} \\
\bottomrule
\end{tabular}%
}
\end{table*}

\subsection{Full Hardware Profiling and Micro-Architectural Telemetry}
Table~\ref{tab:profiling_systems_comparison_full} details the step latency, VRAM allocation, and tensor core utilization—reported both for the isolated DiT backbone (`Iso.`) and end-to-end training step (`Sys`)—across the RTX 4090 and NVIDIA A40 GPUs, retaining the explicit token sequence dimensions per step. 

\begin{figure}[!t]
\centering
\includegraphics[width=0.58\textwidth]{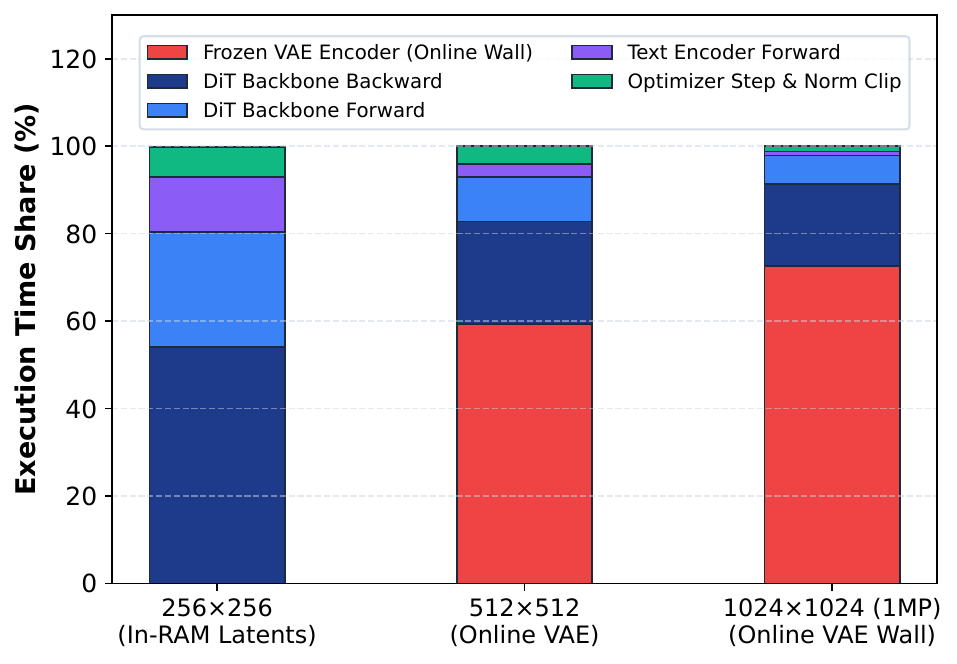}
\caption{\textbf{Per-step execution latency breakdown across resolutions on consumer hardware.} While In-RAM latent caching eliminates online VAE overhead at $256^2$ (devoting $>80\%$ of step latency to the DiT backbone), on-the-fly VAE encoding at $1024^2$ (1MP) consumes \textbf{72.6\%} of wall-clock time ($2{,}548.1\,\text{ms}$ out of $3{,}507.6\,\text{ms}$). Decoupled latent-to-pixel adaptation permanently bypasses this throughput wall.}
\label{fig:vae_wall_breakdown_appendix}
\end{figure}

This full view illustrates the hardware architectural difference: at $256^2$ ($16{,}384\text{ tokens/step}$), the RTX 4090’s large 72MB L2 cache keeps the entire fused Inductor kernel resident on-chip, sustaining $84.22\text{ TFLOPS}$ and cutting total step latency to $308.27\text{ ms}$. In contrast, the A40 overflows its 6MB L2 cache, incurring frequent spills to GDDR6 DRAM ($696\text{ GB/s}$) and doubling total step time to $607.65\text{ ms}$. Figure~\ref{fig:vae_wall_breakdown_appendix} further contrasts In-RAM latent caching at $256^2$ against online VAE encoding at $512^2$ and $1024^2$, highlighting why high-resolution online autoencoding stalls the generative backbone.

\begin{table*}[!t]
\centering
\small
\caption{\textbf{Comprehensive micro-architectural profiling across hardware architectures at $256^2$ and $512^2$.} Unabridged reference for Table~\ref{tab:profiling_systems_comparison}. Profiled across 30 active iterations via PyTorch \texttt{FlopCounterMode} (165.2 TFLOPS peak reference). Reports explicit spatial dimensions, per-step token volumes, and isolated backbone (`Iso.`) vs.\ end-to-end system (`Sys`) $\text{MFU}_{\text{act}}$.}
\label{tab:profiling_systems_comparison_full}
\setlength{\tabcolsep}{2.8pt}
\renewcommand{\arraystretch}{1.08}
\resizebox{\textwidth}{!}{%
\begin{tabular}{lccccccccccc}
\toprule
\shortstack{\textbf{Hardware}\\\textbf{Device}} & 
\shortstack{\textbf{Res.}\\\textbf{Canvas}} & 
\shortstack{\textbf{Batch Size}\\\textbf{(Tokens/Step)}} & 
\shortstack{\textbf{Peak VRAM}\\\textbf{Allocated}} & 
\multicolumn{5}{c}{\textbf{Step Latency Breakdown (ms)}} & 
\shortstack{\textbf{Total Step}\\\textbf{Latency}} & 
\shortstack{\textbf{Backbone}\\\textbf{Throughput}} & 
\shortstack{\textbf{MFU$_{\text{act}}$ (\%)}\\\textbf{(Iso. / Sys)}} \\
\cmidrule(lr){5-9}
& & & & 
\shortstack{\textbf{Data/}\\\textbf{IO}} & 
\shortstack{\textbf{Online}\\\textbf{VAE}} & 
\shortstack{\textbf{Text}\\\textbf{Enc.}} & 
\shortstack{\textbf{Backbone}\\\textbf{(F+B)}} & 
\shortstack{\textbf{Opt. \&}\\\textbf{Clip}} & & & \\
\midrule
\multicolumn{12}{l}{\textit{\textbf{Production Latent Pretraining: Kernel-Fused Tensor Saturation ($256 \times 256$, In-RAM Latents)}}} \\
NVIDIA A40         & $256^2$ & 64 (16,384)          & 13,254.5 MB          & 1.44 ms           & --                & 99.98 ms          & 469.88 ms             & 36.33 ms             & 607.65 ms          & 44.43 TFLOPS        & 26.90\% / 31.71\% \\
\textbf{RTX 4090}  & $256^2$ & \textbf{64 (16,384)} & \textbf{15,249.5 MB} & \textbf{0.27 ms}  & \textbf{0.28 ms}  & \textbf{38.94 ms} & \textbf{247.91 ms}    & \textbf{20.87 ms}    & \textbf{308.27 ms} & \textbf{84.22 TFLOPS} & \textbf{50.98\% / 63.25\%} \\
\midrule
\multicolumn{12}{l}{\textit{\textbf{Online Autoencoder Wall ($512 \times 512$, FLUX.2 VAE On-the-Fly Encoding, Dense Tokens)}}} \\
NVIDIA A40         & $512^2$ & 64 (65,536)          & 35,201.0 MB          & 0.33 ms           & 2,263.52 ms       & 111.34 ms         & 1,396.95 ms           & 36.28 ms             & 3,808.42 ms        & 51.39 TFLOPS        & 31.11\% / \textbf{15.27\%} \\
\bottomrule
\end{tabular}%
}
\end{table*}

\subsection{Analytical Derivation of the Amortization Asymmetry}
\label{sec:appendix_amortization_math}

To formally model the sampling latency trade-off between latent diffusion (with a terminal autoencoder) and pixel diffusion (with a per-step convolutional detailer head), let $N$ denote the number of ODE sampling steps, $t_{\text{step, latent}}$ the per-step latent DiT evaluation latency, $t_{\text{fixed, VAE}}$ the one-time terminal VAE decoding latency, $\Delta t_{\text{head}}$ the marginal per-step latency added by evaluating the Patch Detailer Head (PDH), and $t_{\text{fixed, pixel}} \approx 0$ the negligible fixed pixel tensor formatting overhead. Total generation latencies $T_{\text{latent}}(N)$ and $T_{\text{pixel}}(N)$ for a batch of $B$ images over $N$ sampling steps are then expressed as:
\begin{align}
    T_{\text{latent}}(N) &= N \cdot t_{\text{step, latent}} + t_{\text{fixed, VAE}}, \label{eq:t_latent_unique} \\
    T_{\text{pixel}}(N)  &= N \cdot \left( t_{\text{step, latent}} + \Delta t_{\text{head}} \right) + t_{\text{fixed, pixel}}. \label{eq:t_pixel_unique}
\end{align}
Defining the per-step pixel evaluation latency as $t_{\text{step, pixel}} = t_{\text{step, latent}} + \Delta t_{\text{head}}$, setting $T_{\text{latent}}(N^*) = T_{\text{pixel}}(N^*)$ yields the analytical break-even step threshold $N^*$:
\begin{equation}
    N^* = \frac{t_{\text{fixed, VAE}} - t_{\text{fixed, pixel}}}{\Delta t_{\text{head}}} = \frac{t_{\text{fixed, VAE}} - t_{\text{fixed, pixel}}}{t_{\text{step, pixel}} - t_{\text{step, latent}}}.
    \label{eq:break_even_crossover_formal}
\end{equation}

Substituting the empirical latency parameters benchmarked on an NVIDIA A40 at $1024 \times 1024$ resolution with batch size $B=16$ (Table~\ref{tab:sampling_profiling_appendix}):
\begin{align*}
    t_{\text{fixed, VAE}} &= 4.82\,\text{s}, \quad t_{\text{fixed, pixel}} = 0.09\,\text{s}, \\
    t_{\text{step, latent}} &= 1.387\,\text{s}, \quad t_{\text{step, pixel}} = 1.773\,\text{s} \implies \Delta t_{\text{head}} = 0.386\,\text{s}.
\end{align*}
Evaluating Eq.~\eqref{eq:break_even_crossover_formal} yields:
\begin{equation}
    N^* = \frac{4.82 - 0.09}{1.773 - 1.387} = \frac{4.73}{0.386} \approx 12.25\text{ steps}.
\end{equation}
When $N > N^*$, the accumulated per-step penalty $N \cdot \Delta t_{\text{head}}$ exceeds the one-time VAE decoding cost $t_{\text{fixed, VAE}}$, proving that pixel-space diffusion with recurrent convolutional heads is mathematically sub-optimal for multi-step trajectories without architectural modifications such as patch enlargement (effective pixel patch size $P \ge 32$).

\begin{table*}[!t]
\centering
\small
\caption{\textbf{Sampling execution profile across latent and pixel-space adaptation on NVIDIA A40} ($B=16$ samples, TorchInductor compilation, FlashAttention-2, bfloat16). Latent mode evaluates 206M DiT followed by a terminal VAE decode. Pixel mode evaluates 216M DiT + PDH without an autoencoder.}
\label{tab:sampling_profiling_appendix}
\setlength{\tabcolsep}{4.5pt}
\renewcommand{\arraystretch}{1.08}
\resizebox{\textwidth}{!}{%
\begin{tabular}{lcccccccc}
\toprule
\textbf{Resolution} & \textbf{Mode} & \textbf{Steps} & 
\shortstack{\textbf{Total}\\\textbf{Time}} & 
\shortstack{\textbf{Latency /}\\\textbf{Image}} & 
\shortstack{\textbf{Step Cost}\\\textbf{($t_{\text{step}}$)}} & 
\shortstack{\textbf{Fixed Overhead}\\\textbf{($t_{\text{fixed}}$)}} & 
\shortstack{\textbf{Peak}\\\textbf{VRAM}} & \textbf{Rate} \\
\midrule
$1024 \times 1024$ & Latent + VAE & 30 & 46.43 s & 2.902 s / img & $1.387\text{ s / step}$ & $4.82\text{ s (VAE Decode)}$ & 6.75 GB & \textbf{0.34 img/s} \\
$1024 \times 1024$ & Latent + VAE & 20 & 32.56 s & 2.035 s / img & $1.387\text{ s / step}$ & $4.82\text{ s (VAE Decode)}$ & 6.75 GB & \textbf{0.49 img/s} \\
$1024 \times 1024$ & Pixel (DiP)  & 30 & 53.28 s & 3.330 s / img & $1.773\text{ s / step}$ & $0.09\text{ s (Zero VAE)}$    & 13.22 GB & 0.30 img/s \\
$1024 \times 1024$ & Pixel (DiP)  & 20 & 35.55 s & 2.222 s / img & $1.773\text{ s / step}$ & $0.09\text{ s (Zero VAE)}$    & 13.22 GB & 0.45 img/s \\
\midrule
$512 \times 512$   & Latent + VAE & 30 & 5.99 s  & 0.374 s / img & $0.181\text{ s / step}$ & $0.56\text{ s (VAE Decode)}$ & 2.78 GB & \textbf{2.67 img/s} \\
$512 \times 512$   & Pixel (DiP)  & 30 & 7.89 s  & 0.493 s / img & $0.260\text{ s / step}$ & $0.09\text{ s (Zero VAE)}$    & 4.40 GB & 2.03 img/s \\
\bottomrule
\end{tabular}%
}
\end{table*}

\section{Toy-Diffusion Architectural Ablation Study}
\label{sec:toy_diffusion}

Prior to full-scale foundation pretraining on the 625K dataset, we established the 
controlled \textit{toy-diffusion} \footnote{\url{https://github.com/deeplearningcafe/toy-diffusion}} benchmark on 87,000 image crops ($256 \times 256$) from the \textit{Anime Faces dataset}\footnote{\url{https://huggingface.co/datasets/aipracticecafe/anime-faces-256px-v2}} 
to isolate structural components (Table~\ref{tab:toy_diffusion_ablations}).

\paragraph{Dual-Stream vs.\ Single-Stream Efficiency.} 
Dual-stream cross-attention decisively outperformed single-stream concatenation. 
Concatenating disparate text and visual embeddings into shared attention matrices 
induced modality interference, capping convergence at $\text{FID} = 3.89$. Decoupling 
modalities into dedicated projections achieved higher sustained hardware MFU 
($70.4\%$ vs.\ $49.8\%$) and improved distributional fidelity to $\text{FID} = 3.06$. 
Furthermore, eliminating per-block AdaLN modulation in favor of prepending continuous 
sinusoidal time tokens directly to the context stream reduced backbone parameters by 
$28\%$ without degrading generative quality.

\paragraph{Bidirectional Text Adapters.}
Replacing a static linear/MLP projection with a 2-layer bidirectional Transformer 
Text Adapter (6.09M parameters) reduced FID from $3.42$ to $3.12$. The adapter aligns 
unidirectional causal representations emitted by the frozen \texttt{LFM2.5-230M}~\cite{liquidAI2026230M} LLM 
into an expressive spatial conditioning subspace.

\begin{figure}[t]
\centering
\includegraphics[width=0.85\textwidth]{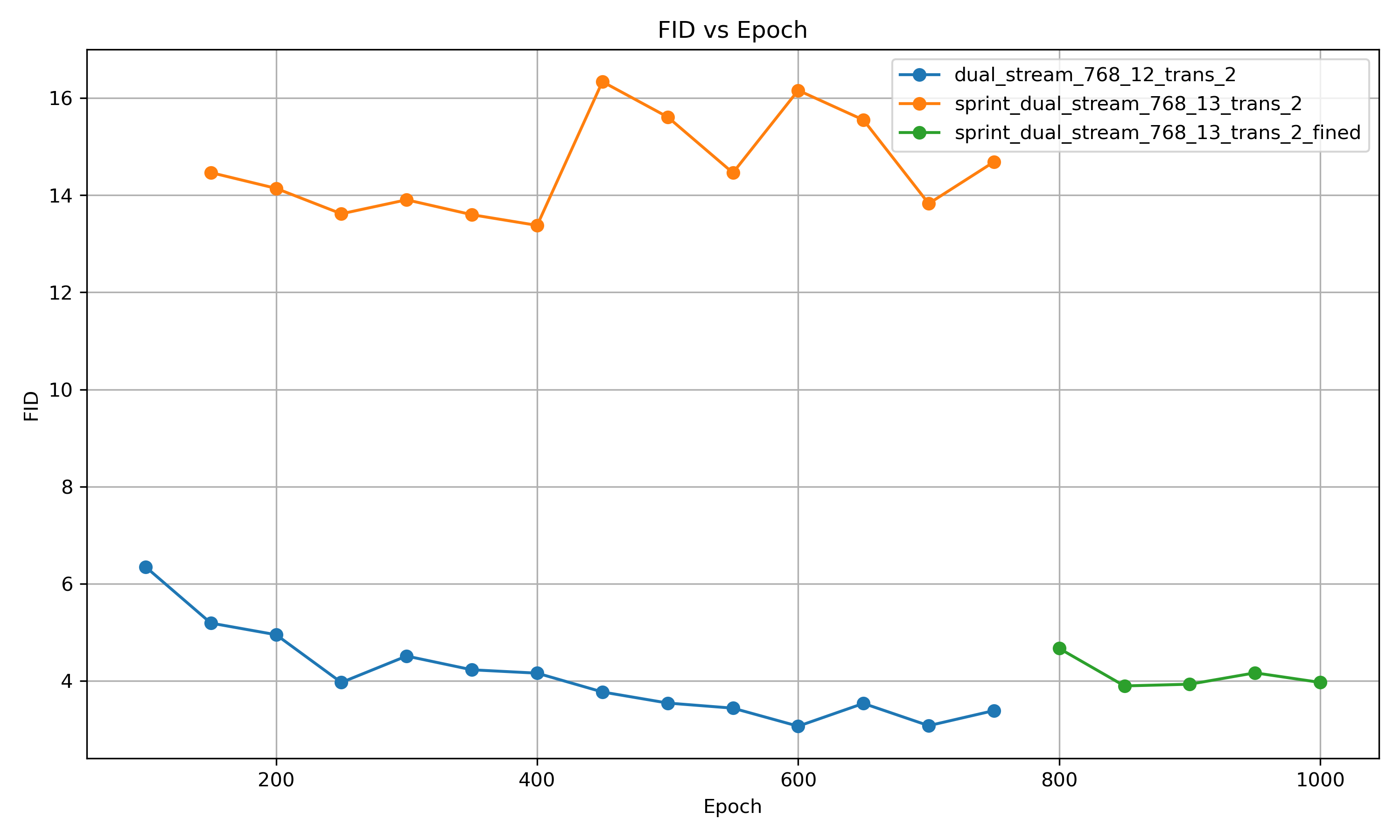}
\caption{\textbf{Quantitative convergence on the \texttt{toy-diffusion} benchmark.} 
Evaluation of SPRINT token drop schedules ($r = 0.75 \to 0.00$) on FID.}
\label{fig:toy_diffusion_curves}
\end{figure}

\paragraph{SPRINT Token-Dropping Dynamics.}
Under structured $2 \times 2$ patch dropping ($r=0.75$), SPRINT reduced training 
iteration latency by $52\%$. While sparse pretraining alone plateaued at 
$\text{FID} = 5.56$, executing a terminal dense refinement phase ($r=0.00$) dropped 
the metric to $\mathbf{3.96}$, matching dense baselines within $0.5$ FID while saving 
over half the cumulative pretraining FLOPs (Figure~\ref{fig:toy_diffusion_curves}).

\paragraph{Parameter Cooldown via WSD Scheduling.}
Diffusion transformers are predominantly trained with static learning rates. We 
evaluated Warmup-Stable-Decay (WSD), transitioning from a flat learning rate to 
linear decay in the terminal $20\%$ of epochs. For the $D=1024$ backbone, parameter 
cooldown dropped FID from $3.27$ to $\mathbf{2.85}$; on the canonical $D=768$ model, 
FID dropped from $3.27$ to $\mathbf{3.06}$, proving that scheduled learning rate 
decay is essential for optimal flow trajectory convergence.

\begin{table}[t]
\centering
\small
\caption{\textbf{Architectural and conditioning ablations on the \texttt{toy-diffusion} 
framework} (evaluated on 87,000 facial portraits at $256 \times 256$, patch size $P=16$).}
\label{tab:toy_diffusion_ablations}
\setlength{\tabcolsep}{7pt}
\renewcommand{\arraystretch}{1.05}
\begin{tabular}{lcccc}
\toprule
\textbf{Backbone / Architecture Variant} & \textbf{Params} & \textbf{BS} & 
\textbf{Epochs} & \textbf{FID $\downarrow$} \\
\midrule
\multicolumn{5}{l}{\textit{\textbf{Part I: Latent Architecture \& Conditioning Ablations}}} \\
Single-Stream DiT ($D=768, L=14$)        & 218M & 256 & 600  & 3.89 \\
Efficient U-Net Baseline                 & 185M & 256 & 400  & 3.45 \\
Dual-Stream DiT + MLP Text Adapter       & 200M & 256 & 600  & 3.42 \\
Dual-Stream DiT + LLM (Frozen LFM2.5)    & 206M & 512 & 1000 & 3.12 \\
Dual-Stream Baseline (Constant LR)       & 206M & 512 & 600  & 3.06 \\
Dual-Stream Baseline (Large Batch)       & 206M & 1024& 1400 & \textbf{2.42} \\
Dual-Stream Wide ($D=1024, L=10$)        & 352M & 512 & 750  & 2.85 \\
SPRINT Dual-Stream ($r=0.75$, Sparse Only) & 206M & 256 & 700 & 5.56 \\
SPRINT Dual-Stream ($r=0.75 \to$ Dense FT) & 206M & 256 & 1000 & \textbf{3.96} \\
\midrule
\multicolumn{5}{l}{\textit{\textbf{Part II: Latent-to-Pixel DiP Adaptation (Toy Scale)}}} \\
Stage 1: PDH Only (Frozen DiT Backbone)  & +10M & 256 & 15   & 28.40 \\
Stage 2: Joint FT ($P=16$, Hard Clamp)   & 216M & 256 & 50   & 19.85 \\
Stage 2: Joint FT ($P=16$, Dynamic Clip) & 216M & 256 & 20   & 18.20 \\
Stage 2: Joint FT ($P=16$, Dynamic Clip) & 216M & 256 & 50   & 16.00 \\
Stage 2: Joint FT ($P=16$, Dynamic Clip) & 216M & 256 & 85   & \textbf{14.39} \\
Stage 2: Staged Patch Widening ($P=16 \to 32$) & 216M & 256 & 50 & \textbf{33.80} \\
\bottomrule
\end{tabular}
\end{table}

\section{Comprehensive Latent-to-Pixel Adaptation Telemetry and Breakdowns}
\label{sec:appendix_pixel_benchmarks}

This section provides the complete, unabridged numerical tables corresponding to the condensed Latent-to-Pixel adaptation summaries in Section~\ref{sec:pixel_adaptation_experiments}.

\subsection{Complete Pixel Adaptation Lifecycle and Hyperparameter Matrix}
Table~\ref{tab:unabridged_pixel_lifecycle_appendix} details the complete optimization progression across all pixel adaptation stages, specifying the trained modules, effective pixel patch size $P$, SPRINT token drop ratio $r$, per-GPU micro-batch size (`Micro-BS`), peak learning rate $\eta$, warmup/decay schedule, epoch and sample counts, and physical silicon throughputs.

\begin{table*}[h!]
\centering
\small
\caption{\textbf{Complete pretraining and adaptation progression for Latent-to-Pixel transfer at $512 \times 512$ on NVIDIA RTX 4090 (24GB).} Hardware evaluated at \$0.33/GPU-hour (\$0.40/h for A40 in Phase 0). Effective batch size is maintained at 128 across all stages using gradient accumulation.}
\label{tab:unabridged_pixel_lifecycle_appendix}
\setlength{\tabcolsep}{2.8pt}
\renewcommand{\arraystretch}{1.08}
\resizebox{\textwidth}{!}{%
\begin{tabular}{lcccccccccccr}
\toprule
\shortstack{\textbf{Stage /}\\\textbf{Phase}} & 
\shortstack{\textbf{Modules}\\\textbf{Trained}} & 
\shortstack{\textbf{Patch}\\\textbf{Size ($P$)}} & 
\shortstack{\textbf{Drop}\\\textbf{($r$)}} & 
\shortstack{\textbf{Micro-BS}\\\textbf{(Per GPU)}} & 
\shortstack{\textbf{Peak LR}\\\textbf{($\eta$)}} & 
\shortstack{\textbf{Warmup /}\\\textbf{Schedule}} & 
\textbf{Epochs} & 
\textbf{Samples} & 
\shortstack{\textbf{Throughput}\\\textbf{(img/s)}} & 
\shortstack{\textbf{MFU (\%)}\\\textbf{Theo / Act}} & 
\shortstack{\textbf{Wall}\\\textbf{Time}} & 
\shortstack{\textbf{Cost}\\\textbf{(\$)}} \\
\midrule
Phase 0: Synth Gen (A40) & -- (Sampling) & -- & -- & 32 & -- & -- & -- & 0.10M & 1.85 & -- & 15.0 h & \$6.00 \\
Stage 1: PDH Warmup      & Embed + PDH   & 16 & 0.75 & 16 & $1.0\times 10^{-4}$ & 25\% Const & 5  & 0.50M & 61.8 & 62.4\% / 72.8\% & 2.25 h & \$0.74 \\
Stage 2A: Joint SPRINT   & Full Model    & 16 & 0.75 & 16 & $6.0\times 10^{-5}$ & 10\% Const & 25 & 2.50M & 57.7 & 58.1\% / 67.8\% & 12.0 h & \$3.98 \\
Stage 2B: Dense Refine   & Full Model    & 16 & 0.00 & 8  & $6.0\times 10^{-5}$ & 10\% WSD    & 15 & 1.50M & 44.4 & 59.8\% / 69.8\% & 9.4 h  & \$3.10 \\
Stage 3: Real Data SFT   & Full Model    & 16 & 0.00 & 8  & $6.0\times 10^{-5}$ & 10\% WSD    & 13 & 5.16M & 42.1 & 56.7\% / 66.2\% & 34.0 h & \$11.22 \\
\midrule
\multicolumn{7}{l}{\textbf{Cumulative Pixel DiT Adaptation Total (Stages 1--3)}} & \textbf{58} & \textbf{9.66M} & \textbf{46.5} & \textbf{57.8\% / 67.5\%} & \textbf{57.65 h} & \textbf{\$19.04} \\
\multicolumn{7}{l}{\textbf{Cumulative Grand Total (Including Offline Synth Generation)}} & -- & -- & -- & -- & \textbf{72.65 h} & \textbf{\$25.04} \\
\midrule
\multicolumn{13}{l}{\textit{\textbf{Progressive Token Compression via Large Patch Adaptation ($P=16 \to 32$)}}} \\
ps32-adapt16 (Large Patch) & Full Model  & 32 & 0.00 & 16 & $4.0\times 10^{-5}$ & 10\% WSD    & 25 & 9.92M & 38.3 & 54.2\% / 63.3\% & 72.0 h & \$23.76 \\
\bottomrule
\end{tabular}%
}
\end{table*}

\subsection{Fine-Grained GenEval and DPG-Bench Task Breakdowns}
Table~\ref{tab:pixel_adaptation_benchmarks_extended} isolates the specific compositional sub-metrics across training checkpoints, comparing native velocity regression ($v$-prediction) against clean-data prediction ($x_0$-prediction) under identical compute durations across Stage~2 (synthetic anchor adaptation) and Stage~3 (real-data SFT refinement). 

\begin{figure*}[h!]
\centering
\begin{subfigure}[b]{0.48\textwidth}
    \centering
    \includegraphics[width=0.92\linewidth]{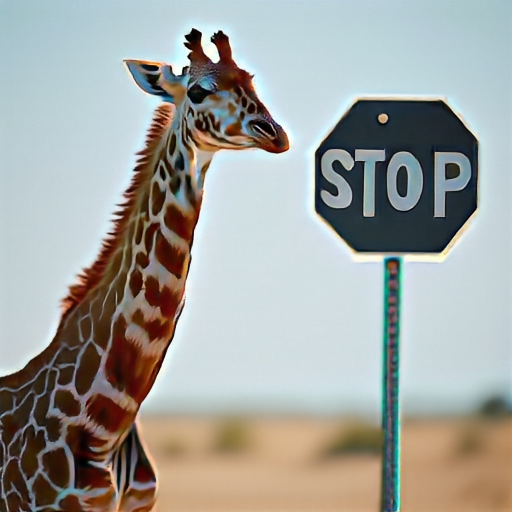}
    \caption{\textbf{Stage 3 SFT ($x_0$-Prediction Ablation, $512^2$):} Converting $\hat{x}_0$ into velocity targets $v_t = (x_t - \hat{x}_0)/t$ triggers $t \to 0$ singularities in \texttt{bfloat16}, forcing dynamic percentile clipping that flattens dynamic range and desaturates colors}
    \label{fig:pixel_x0_pred_giraffe}
\end{subfigure}\hfill
\begin{subfigure}[b]{0.48\textwidth}
    \centering
    \includegraphics[width=0.92\linewidth]{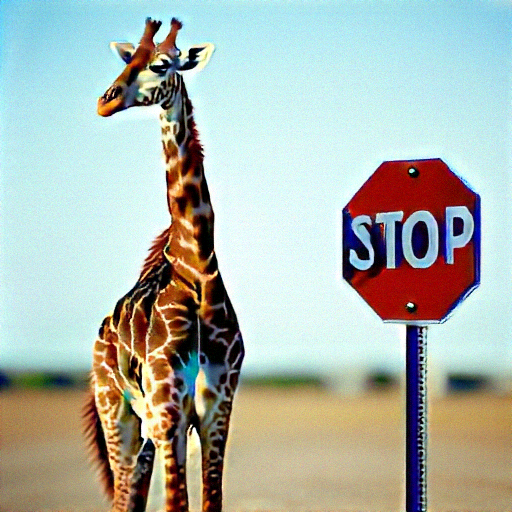}
    \caption{\textbf{Stage 3 SFT (Native $v$-Prediction, Final Pixel DiT):} Direct velocity regression maintains smooth probability paths with \textbf{zero activation clipping}, recovering $>95\%$ of the latent anchor while boosting Two-Object ($0.780$ vs.\ $0.699$) and Color Attribution.}
    \label{fig:pixel_v_pred_giraffe}
\end{subfigure}
\vspace{-4pt}
\caption{\textbf{Qualitative and quantitative comparison of $x_0$-prediction vs.\ native $v$-prediction in $512^2$ Latent-to-Pixel adaptation under identical random seeds.} Native $v$-prediction eliminates dynamic clipping artifacts and color blowout.}
\label{fig:pixel_v_vs_x0_comparison}
\end{figure*}

Across Stage~2B, switching from sparse ($r=0.75$) to dense ($r=0.00$) tokens under $v$-prediction improves two-object layout ($0.543 \to 0.614$), color accuracy ($0.662 \to 0.726$), and color attribution ($0.215 \to 0.273$), consistently outperforming $x_0$-prediction ($0.505$ vs.\ $0.477$ overall GenEval). As illustrated qualitatively in Figure~\ref{fig:pixel_v_vs_x0_comparison}, $x_0$-prediction requires dynamic percentile clipping to suppress $t \to 0$ singularities in \texttt{bfloat16} when recovering velocities $v_t = (x_t - \hat{x}_0)/t$, which flattens dynamic range and desaturates colors, whereas native $v$-prediction preserves vibrant chromatic fidelity with zero activation clipping.

\begin{table*}[h!]
\centering
\small
\caption{\textbf{Comprehensive benchmark evaluation of latent-to-pixel adaptation configurations on GenEval and DPG-Bench across training stages.} All models evaluate at $512^2$ canvas resolution using a 100-step Euler ODE solver with CFG scale $w_{\text{cfg}}=6.0$. Sampling throughput and per-image latency profiled on a single NVIDIA A40 GPU at batch size 32. Intermediate stages demonstrate that $v$-prediction achieves faster initial convergence and prevents color desaturation.}
\label{tab:pixel_adaptation_benchmarks_extended}
\setlength{\tabcolsep}{3.0pt}
\renewcommand{\arraystretch}{1.08}
\resizebox{\textwidth}{!}{%
\begin{tabular}{lcccccccccccc}
\toprule
\shortstack{\textbf{Configuration /}\\\textbf{Stage}} & 
\shortstack{\textbf{Patch}\\\textbf{Size ($P$)}} & 
\shortstack{\textbf{Target}\\\textbf{Obj.}} & 
\shortstack{\textbf{Steps /}\\\textbf{Epochs}} & 
\multicolumn{6}{c}{\textbf{GenEval Task Breakdown}} & 
\shortstack{\textbf{GenEval}\\\textbf{Overall $\uparrow$}} & 
\shortstack{\textbf{DPG-Bench}\\\textbf{Total $\uparrow$}} & 
\shortstack{\textbf{Latency / Img}\\\textbf{(20-step / 30-step)}} \\
\cmidrule(lr){5-10}
& & & & \textbf{Single} & \textbf{Two-Obj} & \textbf{Count} & \textbf{Colors} & \textbf{Position} & \shortstack{\textbf{Color-}\\\textbf{Attr}} & & & \\
\midrule
\multicolumn{13}{l}{\textit{\textbf{Baseline Latent Foundation Model (Pretraining Anchor)}}} \\
Asuka-512 (Base Latent)         & $P=16$ & $v$-pred   & 139 Ep       & 0.950 & 0.811 & 0.347 & 0.817 & 0.278 & 0.450 & \textbf{0.609} & \textbf{78.51} & 0.250 s / 0.374 s \\
\midrule
\multicolumn{13}{l}{\textit{\textbf{Stage 2: Synthetic Data SPRINT Adaptation Progression (100k Synthetic Images)}}} \\
Stage 2A: SPRINT Sparse ($r=0.75$) & $P=16$ & $v$-pred   & 25 Ep (12.0h)& 0.928 & 0.543 & 0.259 & 0.662 & 0.183 & 0.215 & 0.465          & 69.63          & 0.329 s / 0.493 s \\
Stage 2B: Dense Refine ($r=0.00$)  & $P=16$ & $x_0$-pred & 15 Ep (9.4h) & 0.928 & 0.586 & \textbf{0.291} & 0.628 & \textbf{0.195} & 0.233 & 0.477          & 71.49          & 0.329 s / 0.493 s \\
Stage 2B: Dense Refine ($r=0.00$)  & $P=16$ & $v$-pred   & 15 Ep (9.4h) & \textbf{0.941} & \textbf{0.614} & 0.284 & \textbf{0.726} & 0.193 & \textbf{0.273} & \textbf{0.505} & \textbf{72.35} & 0.329 s / 0.493 s \\
\midrule
\multicolumn{13}{l}{\textit{\textbf{Stage 3: Real-Data SFT Convergence (396k Curated Images, $\min(H, W) \ge 1024$px)}}} \\
Stage 3 SFT ($x_0$-pred Ablation)  & $P=16$ & $x_0$-pred & 70k (13 Ep)  & 0.950 & 0.699 & 0.297 & 0.707 & 0.270 & 0.348 & 0.545          & \textbf{76.32} & 0.329 s / 0.493 s \\
\textbf{Stage 3 SFT ($v$-pred, Final Ours)} & $P=16$ & $v$-pred & 70k (13 Ep) & 0.950 & \textbf{0.780} & \textbf{0.335} & \textbf{0.750} & \textbf{0.285} & \textbf{0.420} & \textbf{0.568} & 75.31 & 0.329 s / 0.493 s \\
\midrule
\multicolumn{13}{l}{\textit{\textbf{Progressive Token Compression via Large Patch Adaptation ($P=16 \to 32$)}}} \\
ps32-adapt16 (Large Patch)         & $P=32$ & $v$-pred   & +75k (25 Ep) & 0.938 & 0.576 & 0.325 & 0.739 & 0.195 & 0.250 & 0.504          & 69.87          & \textbf{0.187 s / 0.277 s} \\
\bottomrule
\end{tabular}%
}
\end{table*}

\section{Camera Dynamics, Spatial Disentanglement, and Failure Modes}
\label{sec:camera_dynamics}

Our continuous aspect-normalized RoPE coordinate manifold (Eq.~\eqref{eq:continuous_coords}) and shifted square crop pretraining confer zero-shot virtual camera controllability at inference time without requiring auxiliary camera tokens, fine-tuning, or structural modifications.

\subsection{Zero-Shot Virtual Viewport Manipulation}

\begin{figure*}[!t]
\centering
\begin{minipage}{0.32\textwidth}
    \centering
    \includegraphics[width=\linewidth]{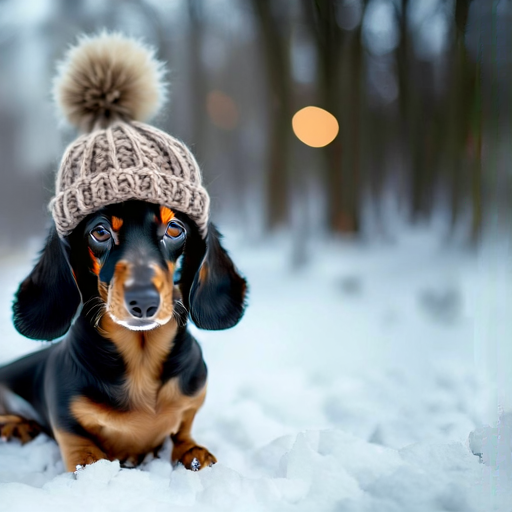}
    \vspace{2pt}
    \centerline{\footnotesize (a) $x_{\text{shift}} = -0.25$}
\end{minipage}\hfill
\begin{minipage}{0.32\textwidth}
    \centering
    \includegraphics[width=\linewidth]{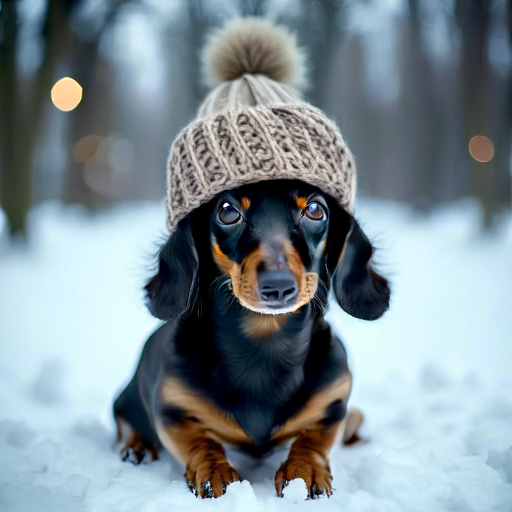}
    \vspace{2pt}
    \centerline{\footnotesize (b) $x_{\text{shift}} = 0.00$}
\end{minipage}\hfill
\begin{minipage}{0.32\textwidth}
    \centering
    \includegraphics[width=\linewidth]{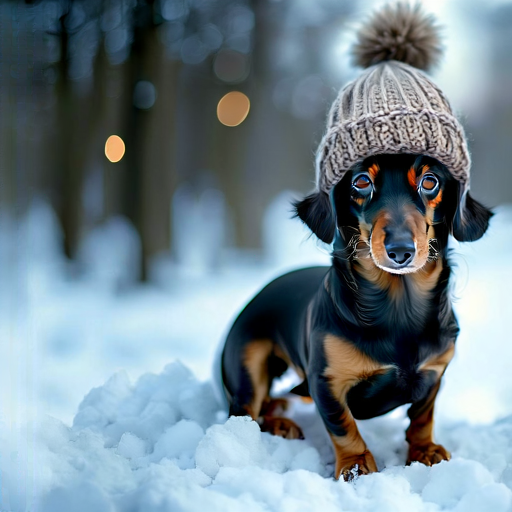}
    \vspace{2pt}
    \centerline{\footnotesize (c) $x_{\text{shift}} = +0.25$}
\end{minipage}
\vspace{-4pt}
\caption{\textbf{Horizontal ($X$-axis) viewport panning under identical noise seeds.} Modulating the horizontal offset $x_{\text{shift}}$ of the continuous coordinate mesh translates the virtual viewport across the manifold, shifting the dachshund horizontally relative to the centered baseline ($x_{\text{shift}} = 0.00$) while preserving knit cap details and snow texture geometry.}
\label{fig:camera_x_shift}
\end{figure*}

\begin{figure*}[!t]
\centering
\begin{minipage}{0.32\textwidth}
    \centering
    \includegraphics[width=\linewidth]{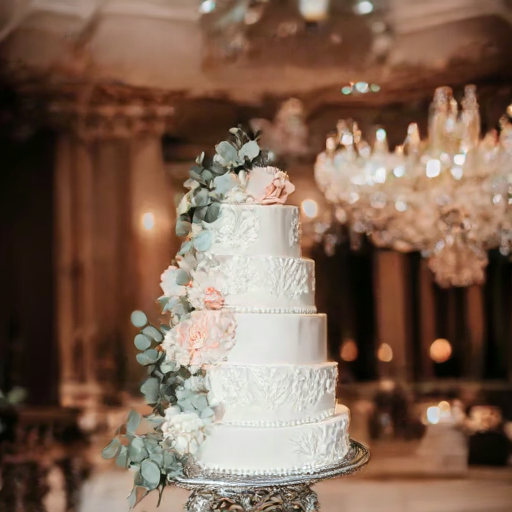}
    \vspace{2pt}
    \centerline{\footnotesize (a) $y_{\text{shift}} = -0.25$}
\end{minipage}\hfill
\begin{minipage}{0.32\textwidth}
    \centering
    \includegraphics[width=\linewidth]{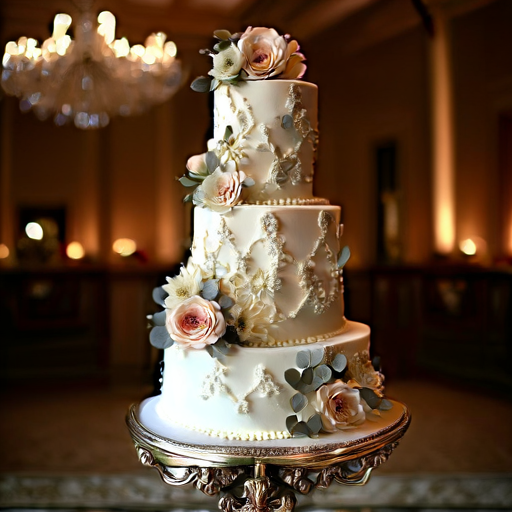}
    \vspace{2pt}
    \centerline{\footnotesize (b) $y_{\text{shift}} = 0.00$}
\end{minipage}\hfill
\begin{minipage}{0.32\textwidth}
    \centering
    \includegraphics[width=\linewidth]{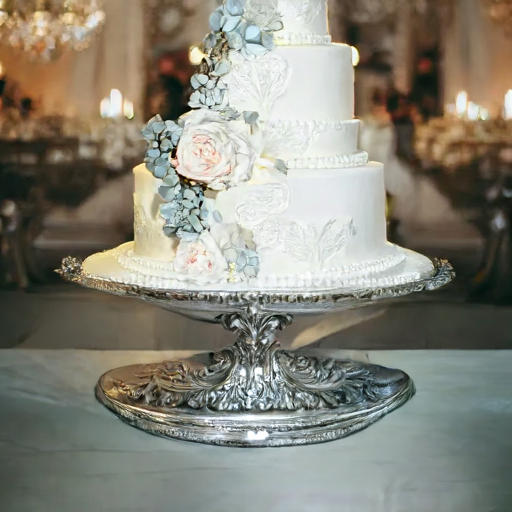}
    \vspace{2pt}
    \centerline{\footnotesize (c) $y_{\text{shift}} = +0.25$}
\end{minipage}
\vspace{-4pt}
\caption{\textbf{Vertical ($Y$-axis) viewport framing under identical noise seeds.} Adjusting the vertical offset $y_{\text{shift}}$ alters the virtual camera elevation: $y_{\text{shift}} = -0.25$ elevates the framing to resolve upper floral tiers and the ballroom chandelier, whereas $y_{\text{shift}} = +0.25$ lowers the viewpoint to ground the ornate metallic pedestal stand.}
\label{fig:camera_y_shift}
\end{figure*}

\begin{figure*}[!t]
\centering
\begin{minipage}{0.32\textwidth}
    \centering
    \includegraphics[width=\linewidth]{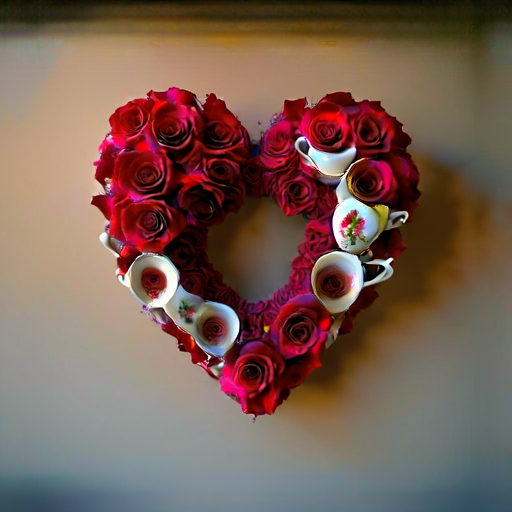}
    \vspace{2pt}
    \centerline{\footnotesize (a) $\text{Zoom} = 0.75\times$}
\end{minipage}\hfill
\begin{minipage}{0.32\textwidth}
    \centering
    \includegraphics[width=\linewidth]{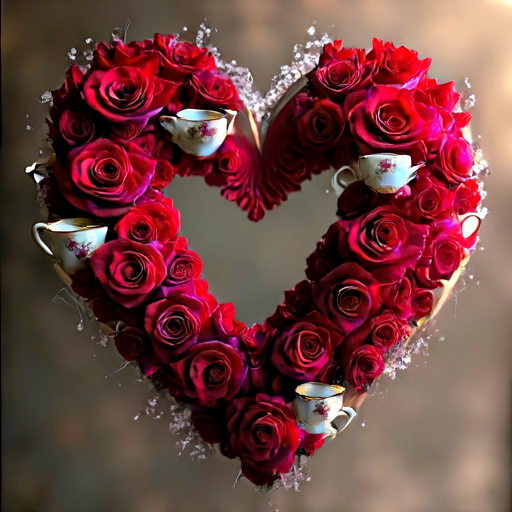}
    \vspace{2pt}
    \centerline{\footnotesize (b) $\text{Zoom} = 1.00\times$}
\end{minipage}\hfill
\begin{minipage}{0.32\textwidth}
    \centering
    \includegraphics[width=\linewidth]{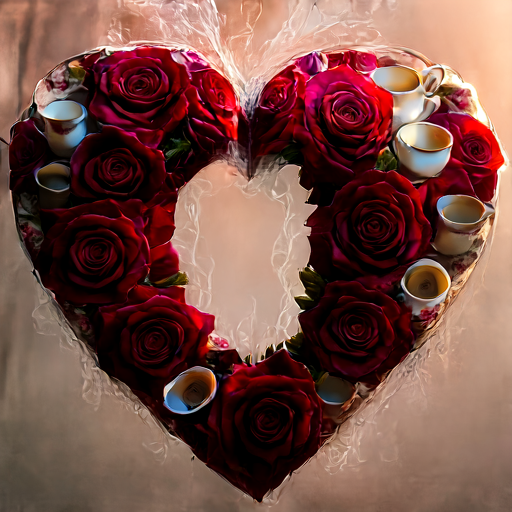}
    \vspace{2pt}
    \centerline{\footnotesize (c) $\text{Zoom} = 1.25\times$}
\end{minipage}
\vspace{-4pt}
\caption{\textbf{Continuous focal zoom scaling via RoPE coordinate grid compression.} Rescaling the continuous evaluation mesh by optical magnification factor $z \in \{0.75\times, 1.00\times, 1.25\times\}$ simulates continuous wide-angle expansion and focal zoom under identical latent noise initializations. At $z = 1.25\times$, transformer attention capacity concentrates on local topological features, resolving fine rose petal contours and porcelain teacup glazes while strictly maintaining global structural alignment.}
\label{fig:camera_zoom}
\end{figure*}

\begin{figure*}[!t]
\centering
\begin{subfigure}[b]{0.32\textwidth}
    \centering
    \includegraphics[width=\linewidth]{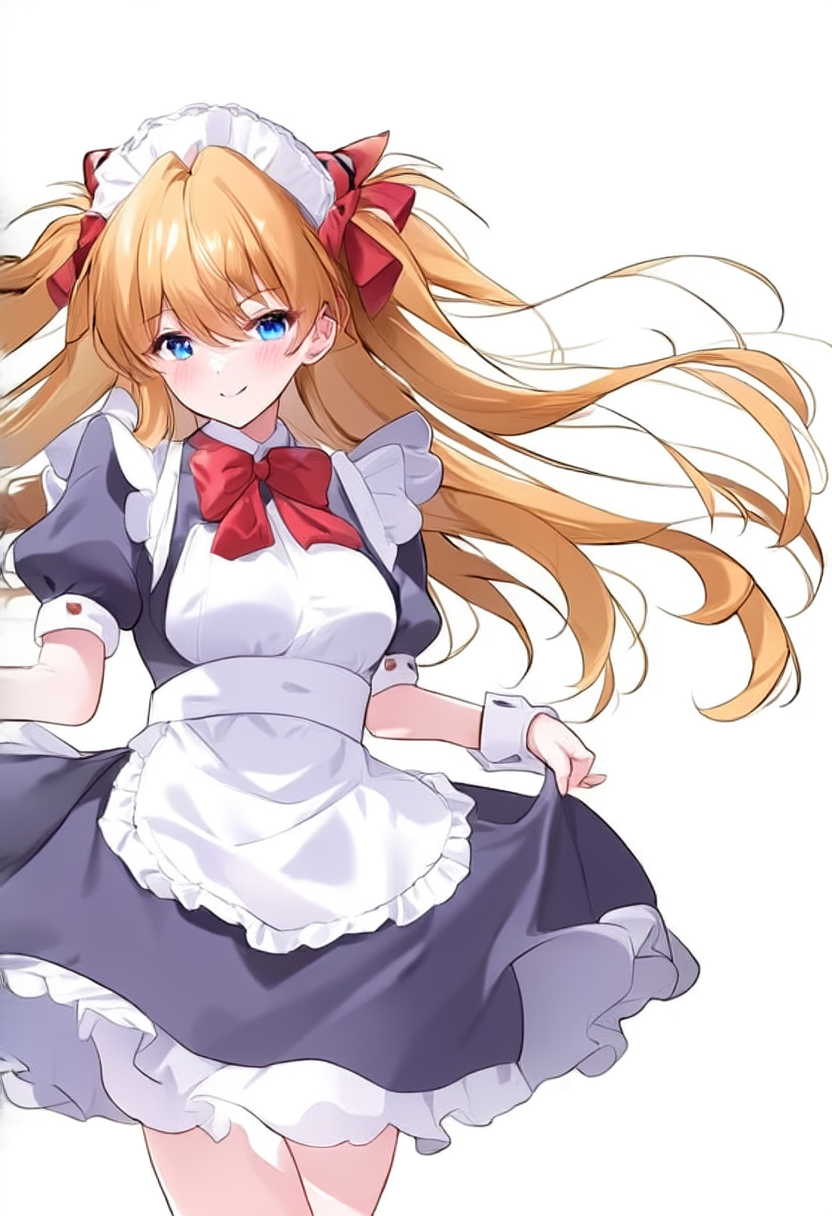}
    \caption{$x_{\text{shift}} = -0.25$ (Camera Left)}
\end{subfigure}\hfill
\begin{subfigure}[b]{0.32\textwidth}
    \centering
    \includegraphics[width=\linewidth]{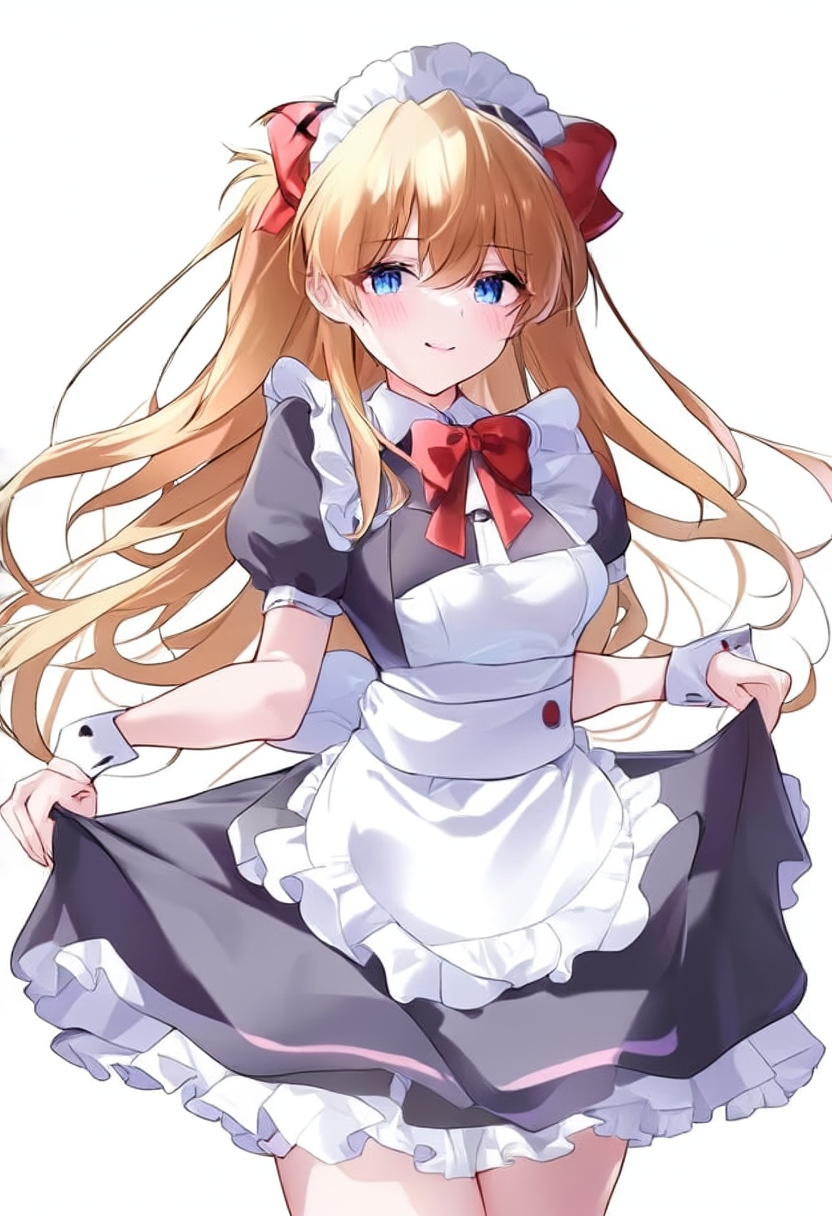}
    \caption{$x_{\text{shift}} = 0.00$ (Center Viewport)}
\end{subfigure}\hfill
\begin{subfigure}[b]{0.32\textwidth}
    \centering
    \includegraphics[width=\linewidth]{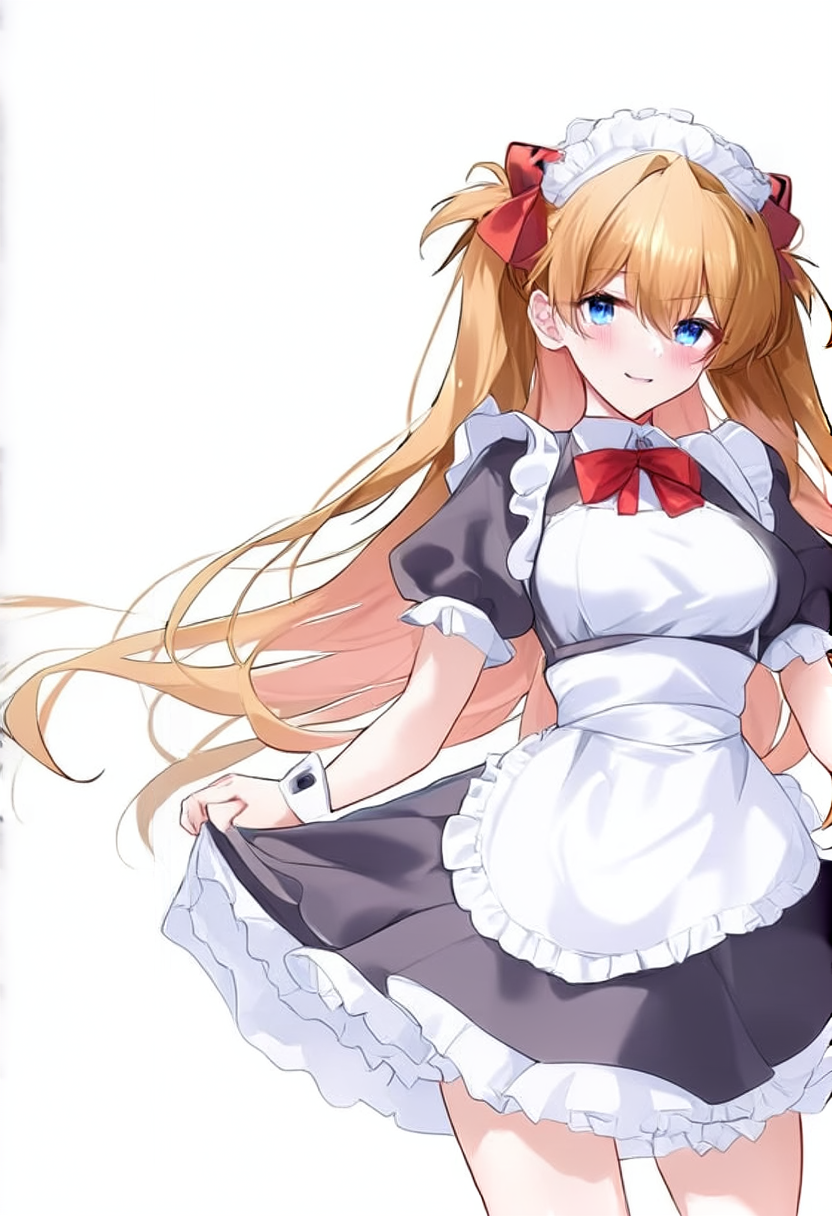}
    \caption{$x_{\text{shift}} = +0.25$ (Camera Right)}
\end{subfigure}

\vspace{6pt}
\begin{subfigure}[b]{0.32\textwidth}
    \centering
    \includegraphics[width=\linewidth]{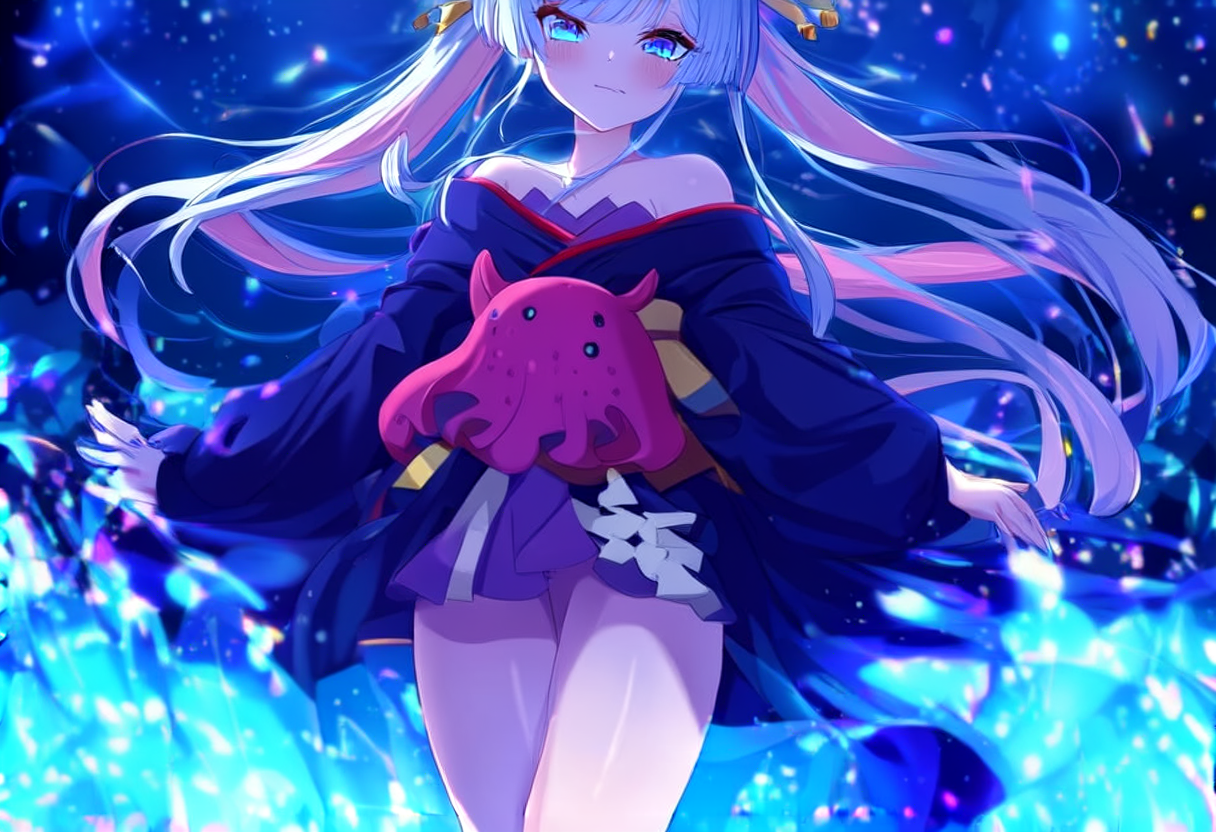}
    \caption{$y_{\text{shift}} = -0.25$ (Camera Up)}
\end{subfigure}\hfill
\begin{subfigure}[b]{0.32\textwidth}
    \centering
    \includegraphics[width=\linewidth]{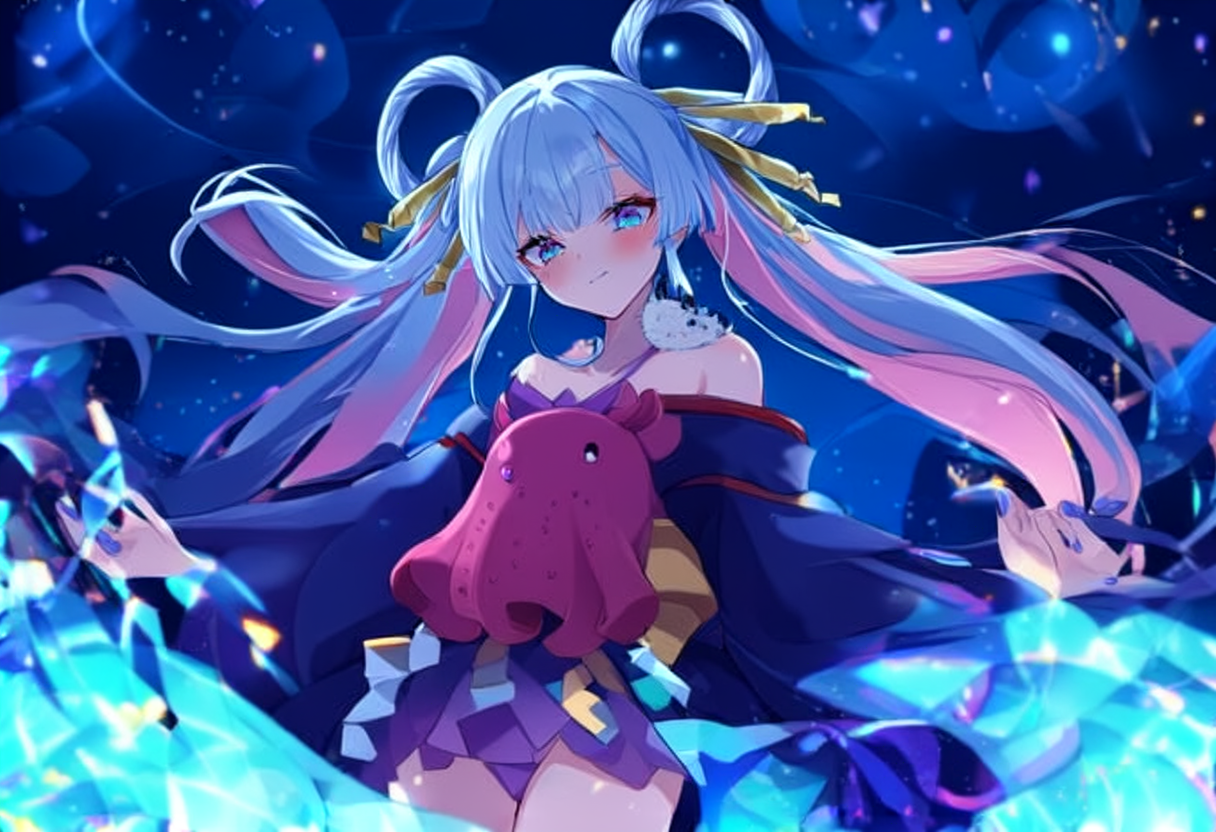}
    \caption{$y_{\text{shift}} = 0.00$ (Center Viewport)}
\end{subfigure}\hfill
\begin{subfigure}[b]{0.32\textwidth}
    \centering
    \includegraphics[width=\linewidth]{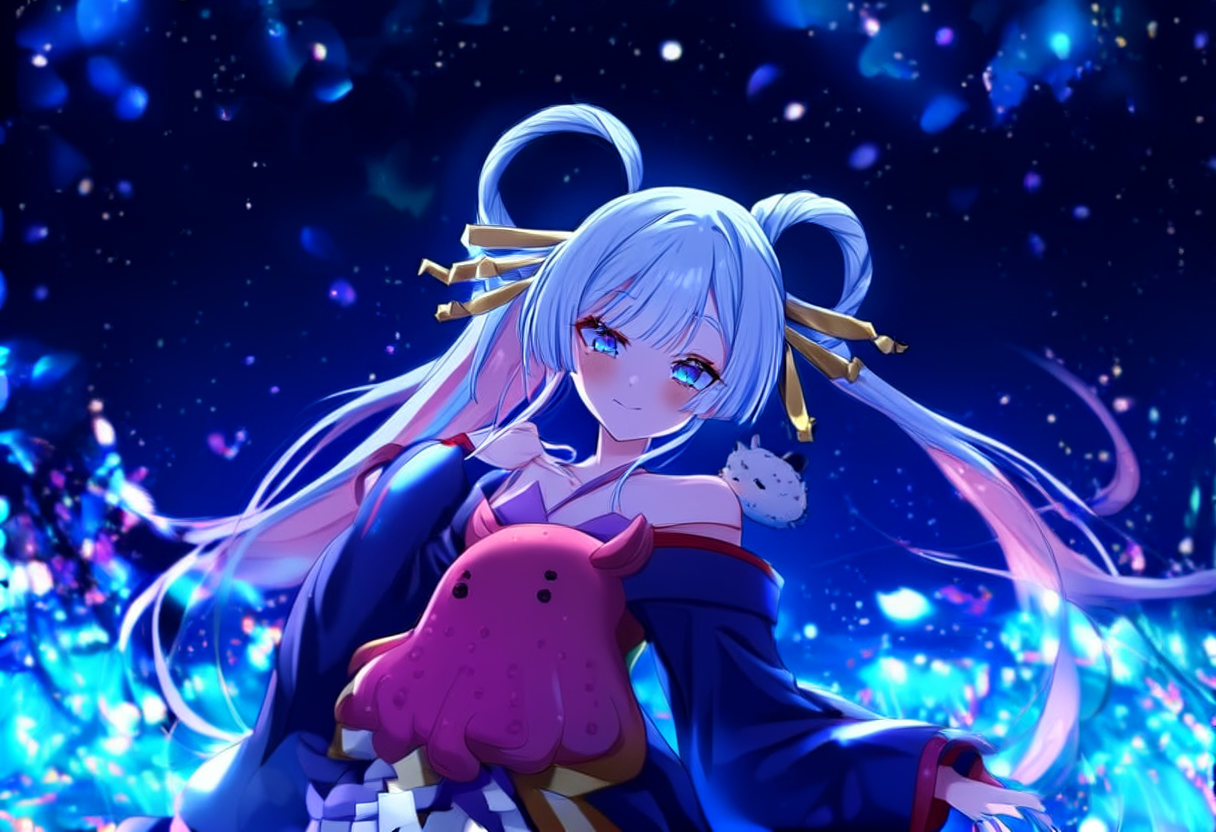}
    \caption{$y_{\text{shift}} = +0.25$ (Camera Down)}
\end{subfigure}

\vspace{6pt}
\begin{subfigure}[b]{0.32\textwidth}
    \centering
    \includegraphics[width=\linewidth]{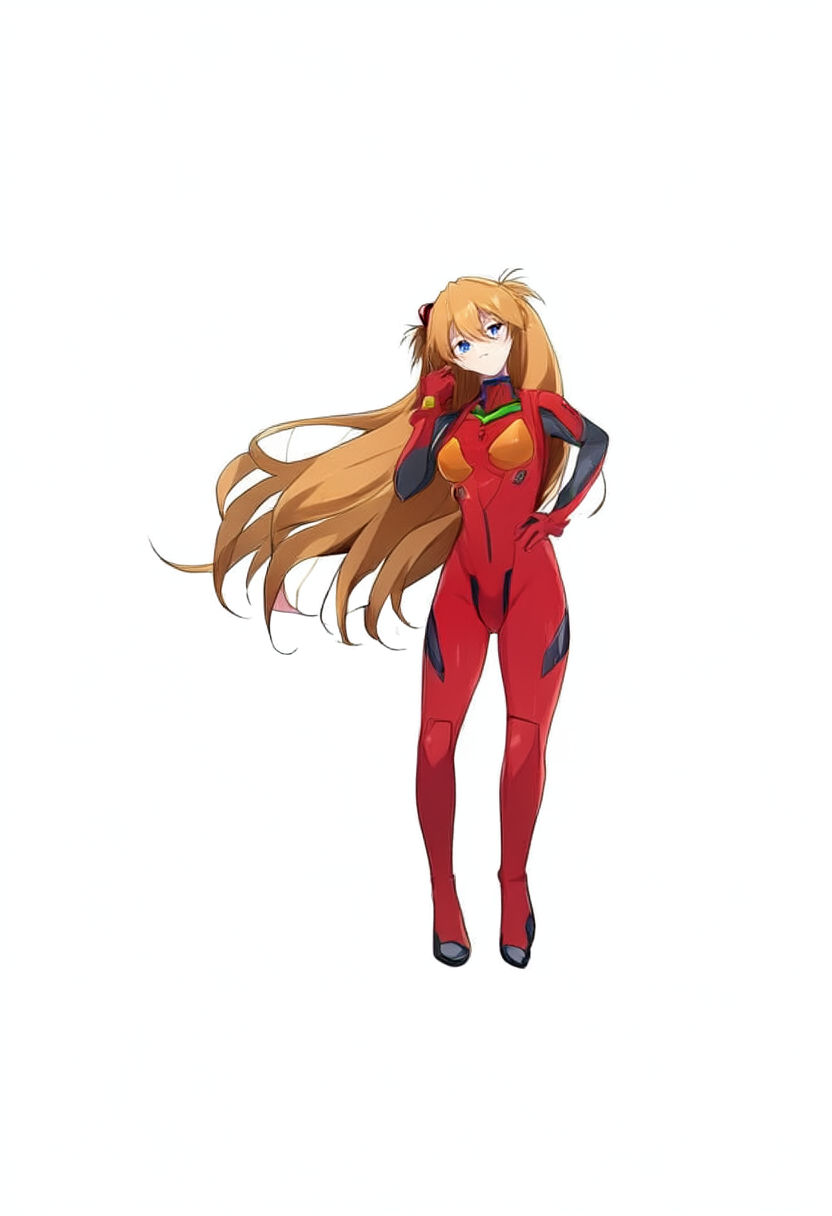}
    \caption{Zoom $= 0.75\times$}
\end{subfigure}\hfill
\begin{subfigure}[b]{0.32\textwidth}
    \centering
    \includegraphics[width=\linewidth]{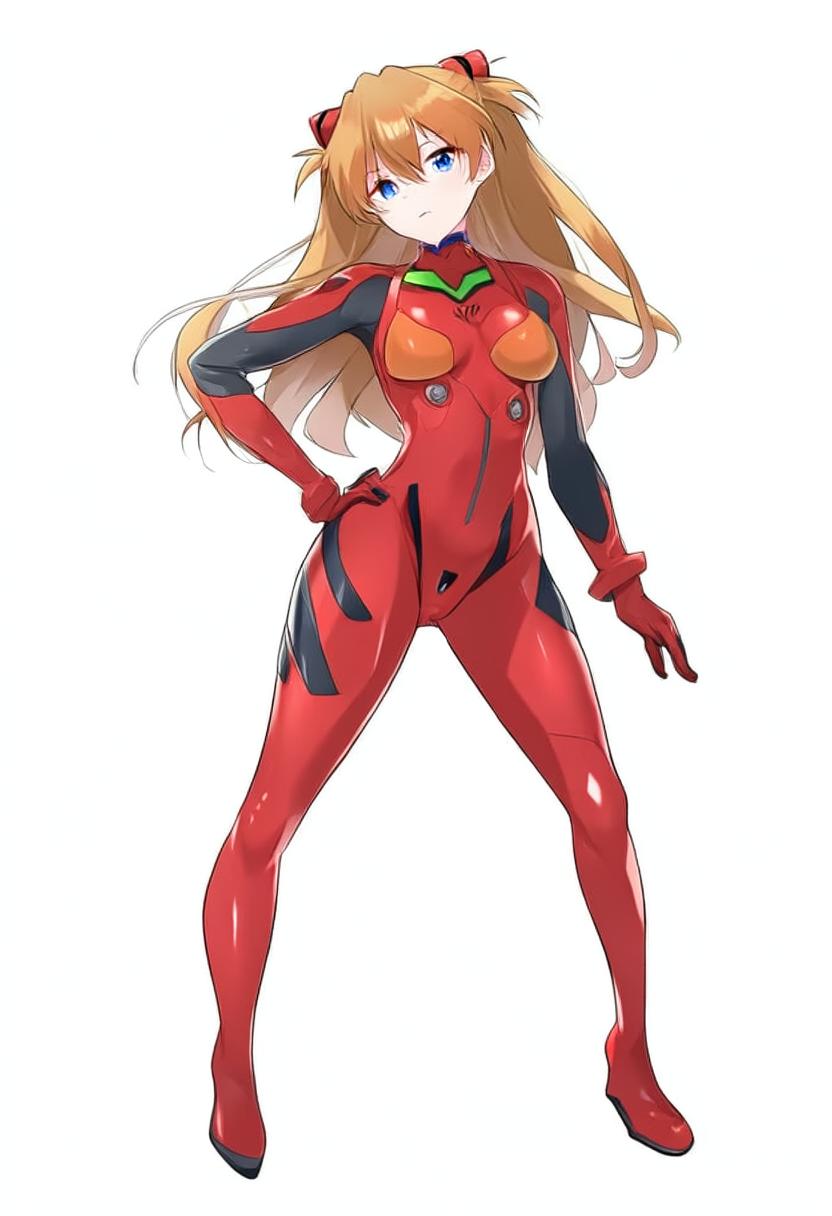}
    \caption{Zoom $= 1.00\times$}
\end{subfigure}\hfill
\begin{subfigure}[b]{0.32\textwidth}
    \centering
    \includegraphics[width=\linewidth]{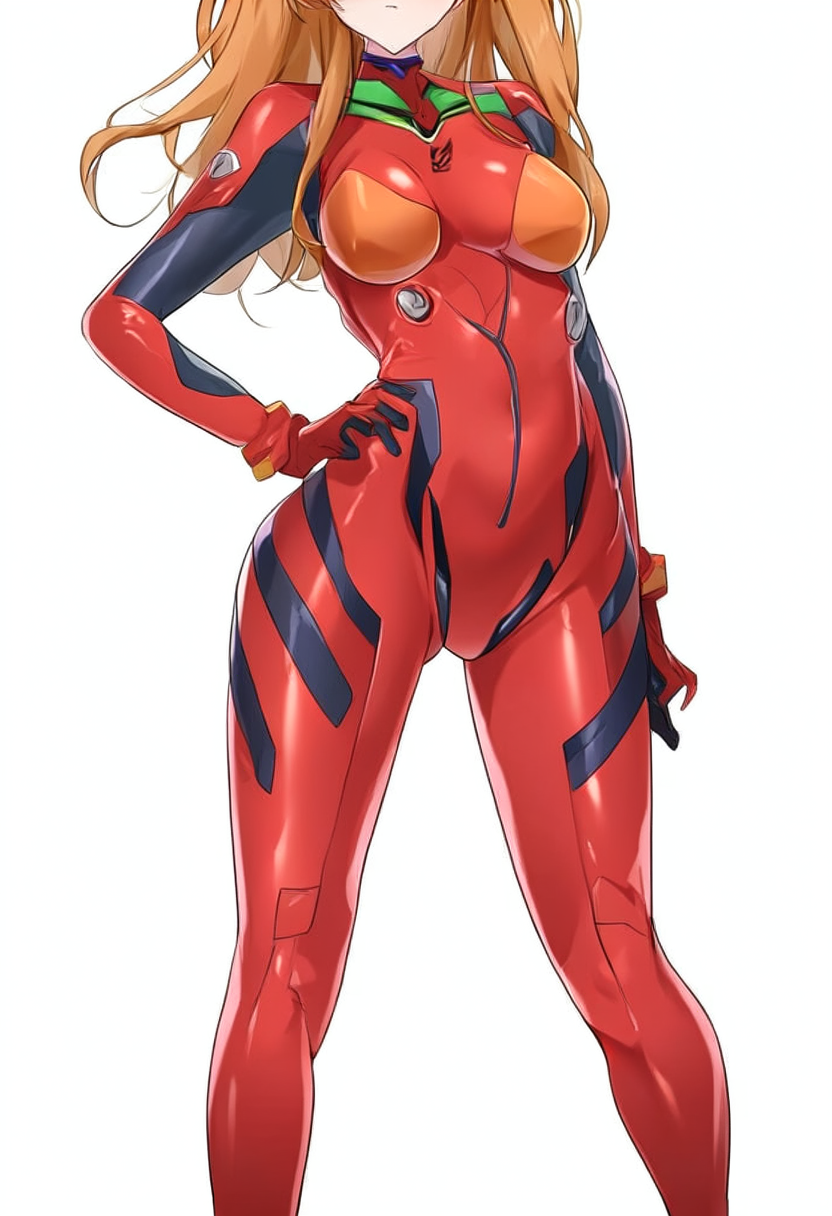}
    \caption{Zoom $= 1.25\times$}
\end{subfigure}
\vspace{-4pt}
\caption{\textbf{Zero-shot continuous virtual camera manipulation on anime illustrations under identical random seeds.} Manipulating the continuous coordinate mesh allows continuous horizontal panning (Row~1), vertical elevation adjustment (Row~2), and optical zoom (Row~3) purely via position map manipulation, without retraining, camera adapters, or prompt modifications.}
\label{fig:anime_camera_controls}
\vspace{-4pt}
\end{figure*}

Applying affine transformations (Eq.~\eqref{eq:camera_transforms}) directly to the continuous evaluation coordinate mesh $(x_{\text{pos}}, y_{\text{pos}})$ parameterized by normalized horizontal and vertical pan offsets $(x_{\text{shift}}, y_{\text{shift}})$ and an optical magnification factor $z \equiv \text{zoom} \in \mathbb{R}^+$ enables three independent, continuous spatial degrees of freedom under identical random seeds:

\begin{itemize}
    \item \textbf{Horizontal Panning ($X$-Shift):} Figures~\ref{fig:camera_x_shift} and~\ref{fig:anime_camera_controls} (Row~1) illustrate horizontal camera panning. Applying non-zero offsets $x_{\text{shift}} \in \{-0.25, +0.25\}$ translates the virtual viewport horizontally across the scene manifold relative to the centered baseline ($x_{\text{shift}} = 0.00$), shifting subject framing and revealing peripheral background context while preserving fine anatomical structures, eye catchlights, and woven fabric textures without artifacting.
    \item \textbf{Vertical Panning ($Y$-Shift):} Figures~\ref{fig:camera_y_shift} and~\ref{fig:anime_camera_controls} (Row~2) demonstrate vertical framing control. Injecting a positive offset ($y_{\text{shift}} = +0.25$) lowers the virtual camera viewpoint, grounding the composition by resolving the ornate metallic pedestal stand, whereas a negative offset ($y_{\text{shift}} = -0.25$) elevates the viewpoint to capture the upper floral tiers and the ballroom chandelier.
    \item \textbf{Continuous Optical Zoom:} Figures~\ref{fig:camera_zoom} and~\ref{fig:anime_camera_controls} (Row~3) demonstrate continuous focal magnification. Compressing the evaluation grid via $z = 1.25\times$ concentrates the DiT's spatial attention on the central wreath cavity, magnifying individual porcelain teacup rims and petal folds without geometric distortion, while $z = 0.75\times$ expands the field of view for wide-angle framing.
\end{itemize}

\subsection{Failure Modes and Generation Artifacts}
\label{sec:failure_modes}

\begin{figure}[!t]
\centering
\begin{minipage}{0.48\columnwidth}
    \centering
    \includegraphics[width=\linewidth]{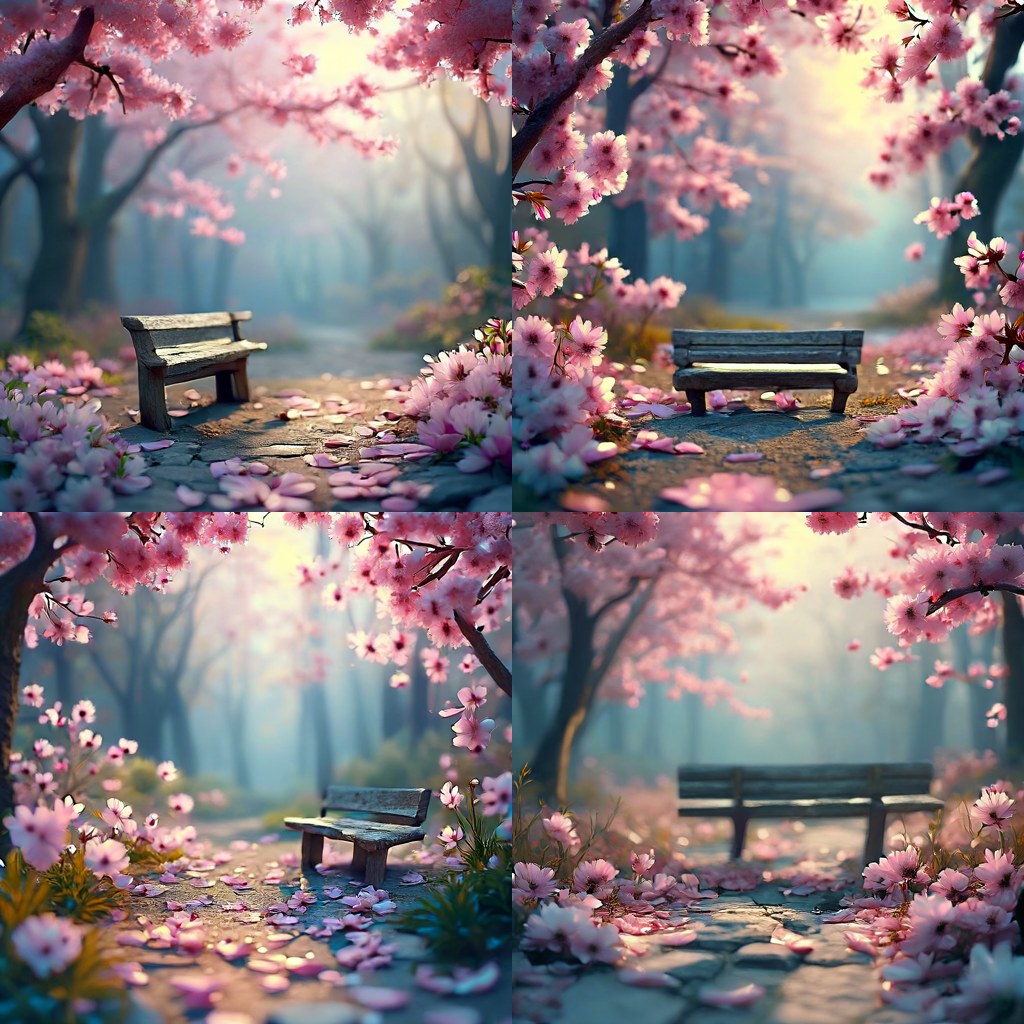}
    \vspace{2pt}
    \centerline{\footnotesize (a) Unzoomed ($z = 1.0\times$)}
\end{minipage}\hfill
\begin{minipage}{0.48\columnwidth}
    \centering
    \includegraphics[width=\linewidth]{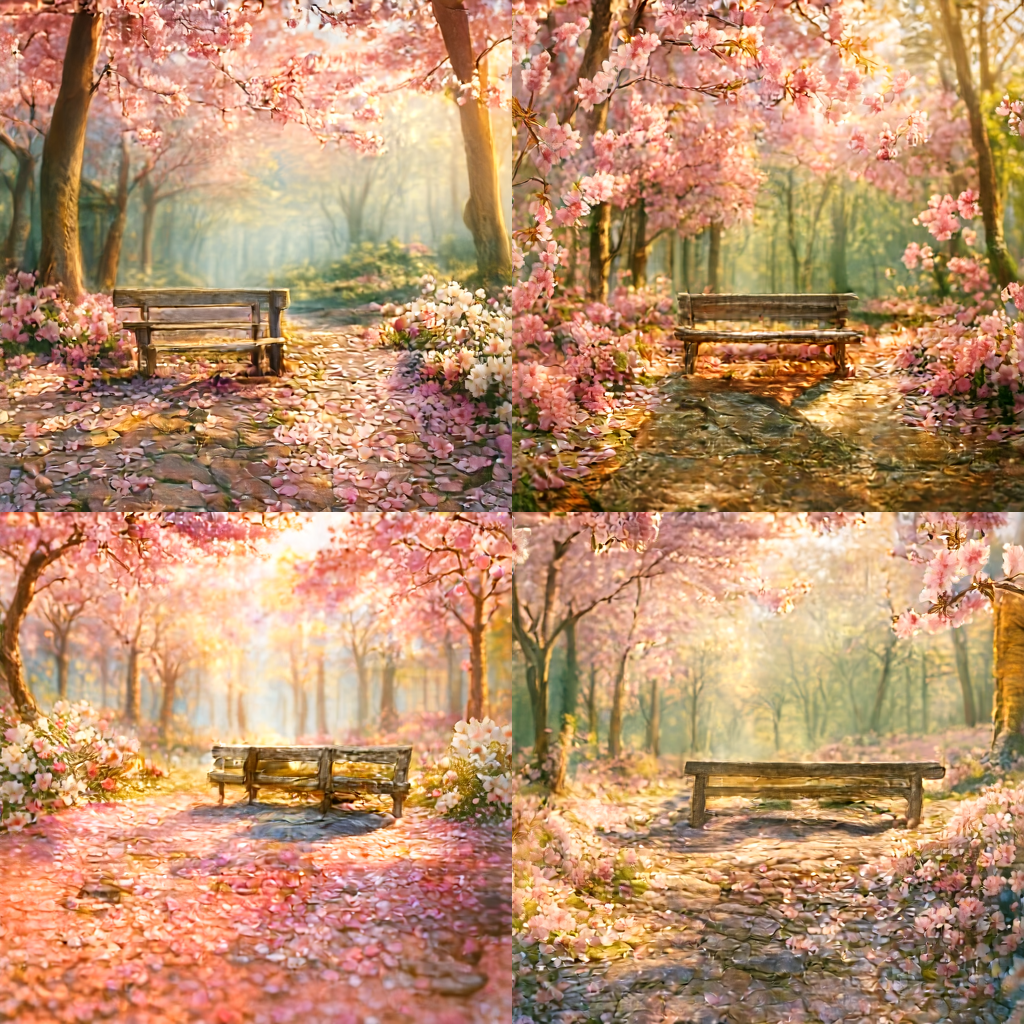}
    \vspace{2pt}
    \centerline{\footnotesize (b) RoPE Zoom ($z > 1.0\times$)}
\end{minipage}
\vspace{-4pt}
\caption{\textbf{Failure mode comparison under continuous RoPE zoom on a complex organic composition.} \textbf{(a)} At canonical scale ($z = 1.0\times$), the 206M dual-stream backbone synthesizes a coherent scene with sharp lighting, structured perspective, and clean floral silhouettes. \textbf{(b)} Applying continuous RoPE focal zoom ($z > 1.0\times$) compresses the evaluation grid, inducing: (i) painterly high-frequency texture smearing on the stone pathway, (ii) semantic attribute bleeding (pink pigment leaking into ground cobblestones), and (iii) spatial non-rigidity and warping along linear wooden bench slats.}
\label{fig:failure_case_bench}
\end{figure}

While Asuka-FM achieves competitive open-domain synthesis (0.609 GenEval, 78.51 DPG-Bench) within a limited compute envelope ($\sim$160 GPU hours), the model exhibits localized failure modes when subjected to out-of-distribution prompts. Figure~\ref{fig:failure_case_bench} provides an explicit comparative ablation.

A systematic analysis reveals that these artifacts are not structural flaws of the base generator, but rather emerge from the physical limits of continuous coordinate scaling:

\paragraph{1. High-Frequency Texture Smearing via Semantic Dilation.}
As demonstrated in Figure~\ref{fig:failure_case_bench}(a), the base model readily generates coherent cobblestones, realistic foliage distributions, and consistent depth-of-field. However, when focal zoom ($z > 1.0\times$) is applied (Figure~\ref{fig:failure_case_bench}(b)), the coordinate grid contracts, forcing each patch of effective pixel stride $P=16$ to cover a smaller physical subregion of the scene. Because the compact 206M backbone lacks multi-scale texture detailers or sub-patch supervision, it cannot hallucinate micro-surface details (e.g., individual stone crevices or fine wood grain). The model consequently defaults to smooth velocity trajectories, producing the waxy, painterly swirls visible in the foreground path.

\paragraph{2. Spatial Non-Rigidity via Rotary Phase Dampening.}
Recall that $r_W$ denotes the aspect-ratio width scale factor, $N_W$ the horizontal patch-grid count, and $\omega_k \in [\pi/4, 2\pi]$ the spatial frequency for rotation channel $k$. While the calibrated spectrum guarantees isotropic decay at unit scale ($z = 1.0\times$), compressing the coordinate mesh by zoom factor $z$ reduces the effective inter-patch phase difference to $\Delta \theta_k = (2 r_W / (N_W \cdot z)) \omega_k$. This phase dampening slightly weakens relative positional discriminability between adjacent visual tokens in the spatial attention kernel ($K(\mathbf{p}_i, \mathbf{p}_{i+1}) \to 1.0$, Eq.~\eqref{eq:rope_kernel}), resulting in geometric relaxation. While organic structures (foliage, clouds) tolerate this relaxation, man-made rigid objects (such as the linear wooden slats of the bench) exhibit noticeable curvature and structural warping.

\paragraph{3. Semantic Attribute Bleeding under Compressed Receptive Fields.}
In dense compositional scenes containing strong complementary colors (vibrant pink petals against neutral grey stones and brown wood), joint self-attention in the shallow encoder $f_\theta$ tends to disperse dominant color tokens. Under focal zoom, where coordinate density is artificially compressed, the effective receptive field of self-attention spans a broader semantic neighborhood in normalized space. This causes the pink floral pigment to bleed into adjacent latents, tinting the cobblestone path and staining the bench wood.

\subsection{Positional Representation Robustness}
\label{sec:appendix_rope_plasticity}

\begin{figure*}[t!]
\centering
\begin{subfigure}[b]{0.48\textwidth}
    \centering
    \includegraphics[width=0.88\linewidth]{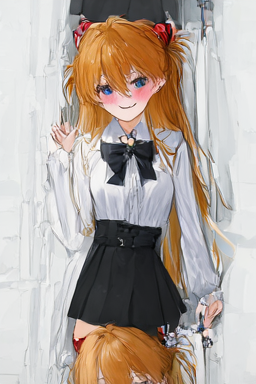}
    \caption{\textbf{Discrete RoPE Collapse:} Unseen integer positions outside the square grid induce harmonic repetition, causing duplicated heads and fragmented torsos along the vertical canvas.}
    \label{fig:rope_uncalibrated}
\end{subfigure}\hfill
\begin{subfigure}[b]{0.48\textwidth}
    \centering
    \includegraphics[width=0.88\linewidth]{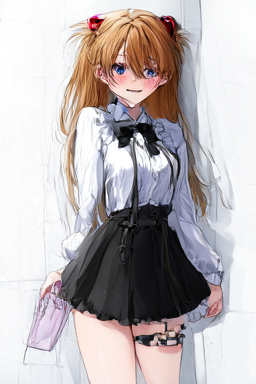}
    \caption{\textbf{Continuous Calibrated RoPE:} Aspect-normalized continuous coordinates bound the manifold within $[-r_H, r_H] \times [-r_W, r_W]$, yielding single-subject anatomical consistency.}
    \label{fig:rope_calibrated}
\end{subfigure}
\vspace{-4pt}
\caption{\textbf{Ablation of RoPE positional formulations across non-square aspect ratios under identical random seeds.} Switching to continuous aspect-normalized coordinates with frequency spectrum $\omega \in [\pi/4, 2\pi]$ completely resolves anatomical tiling without requiring pretraining restarts.}
\label{fig:rope_curtain_ablation}
\end{figure*}

During early exploratory pretraining on the 337K Danbooru corpus, Asuka-FM was initially optimized with discrete integer position IDs ($\text{pos} \in \{0, \dots, N-1\}$, where $N$ is the image token count). Deviating from the square aspect ratio during inference exposed an architectural failure: because inter-patch indices outside the training grid are unseen, the model interpreted the expanded canvas as additional spatial instances, replicating heads and torsos (Figure~\ref{fig:rope_curtain_ablation}(a)).

Rather than restarting pretraining from scratch, we investigated \textit{positional representation plasticity}: whether a partially trained diffusion transformer could adapt mid-flight to continuous aspect-normalized RoPE without catastrophic forgetting of visual and character priors.

\begin{table}[h]
\centering
\small
\caption{\textbf{Stage 1 RoPE swap progression on the 337K Danbooru corpus:} Successful coordinate adaptation from discrete integer grid to continuous aspect-normalized RoPE within 20 epochs ($\approx$6.74M samples) on a single RTX 4090.}
\label{tab:rope_swap_progression}
\resizebox{\columnwidth}{!}{%
\begin{tabular}{lccccccr}
\toprule
\textbf{Phase} & \textbf{Epochs} & \textbf{Drop} & \textbf{RoPE Mode} 
& \textbf{Samples} & \textbf{Patches} & \textbf{PFLOPs$_{\text{exec}}$} & \textbf{Time} \\
\midrule
1A: Base Sparse     & 0--100   & 0.75 & Discrete   & 33.70M & 8.63B & 7.2 & 28.0 h \\
1B: RoPE Adaptation & 100--120 & 0.75 & Continuous & 6.74M  & 1.73B & 1.4 & 5.6 h \\
1C: Dense FT        & 120--145 & 0.00 & Continuous & 8.43M  & 2.16B & 4.2 & 10.0 h \\
1D: Cosine Decay    & 145--170 & 0.00 & Continuous & 8.43M  & 2.16B & 4.2 & 10.0 h \\
\midrule
\multicolumn{4}{l}{\textbf{Total Stage 1 (Swapped Run)}} & 57.30M & 14.68B & \textbf{17.0} & \textbf{53.6 h} \\
\bottomrule
\end{tabular}%
}
\end{table}

As detailed in Table~\ref{tab:rope_swap_progression}, the RoPE adaptation phase (Phase~1B of the swapped run) realigned the model to continuous coordinates within only 20 epochs ($5.6$ GPU hours). The attention query-key projection matrices smoothly adapted to continuous phase rotations without loss of text-image alignment or character identities, and subsequent dense fine-tuning (Phases~1C and~1D) completely eradicated replication artifacts across arbitrary aspect ratios (Figure~\ref{fig:rope_curtain_ablation}(b)).

\subsection{Semantic Attribute Disentanglement}
\label{sec:appendix_disentanglement}

A hallmark of a well-conditioned multi-modal transformer is the mathematical decoupling of semantic attributes (e.g., color, apparel) from spatial and geometric priors (e.g., body pose, lighting, perspective).

\begin{figure*}[t!]
\centering
\begin{subfigure}[b]{0.48\textwidth}
    \centering
    \includegraphics[width=0.88\linewidth]{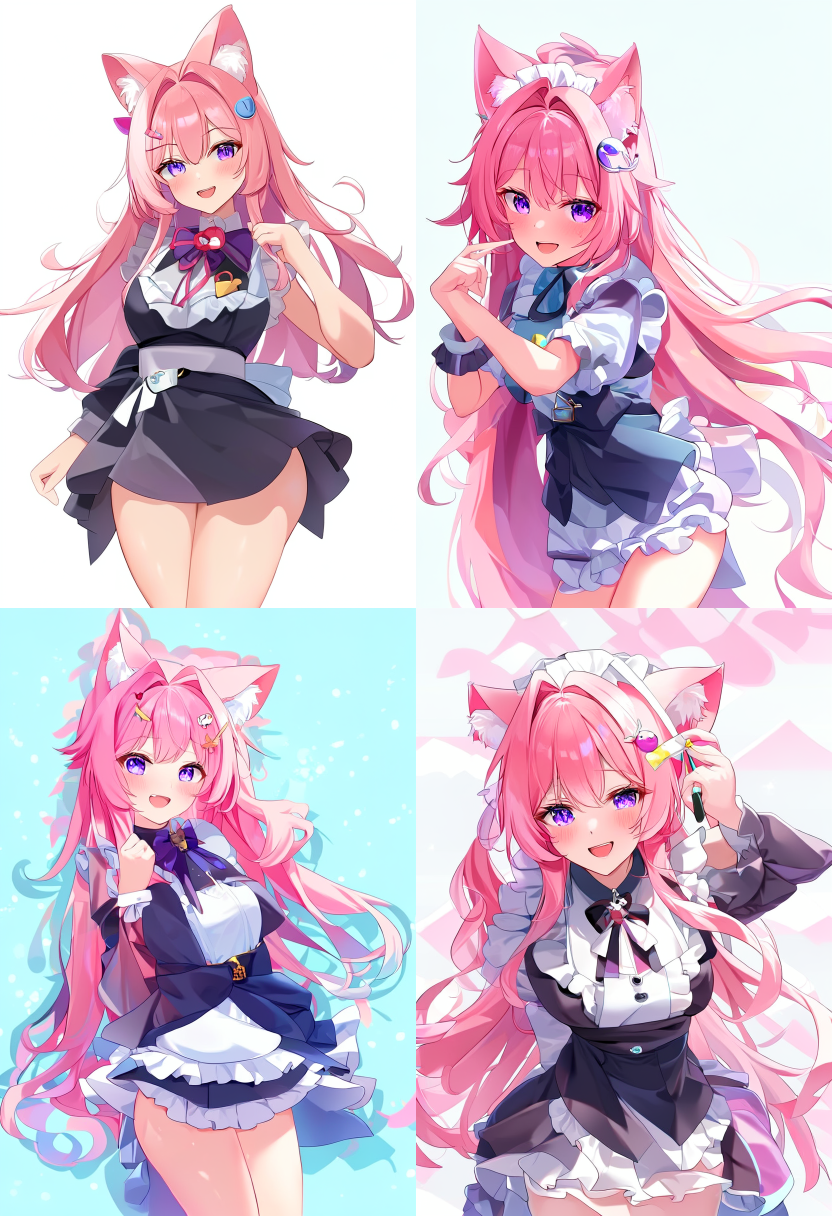}
    \caption{\textbf{Prompt conditioning:} \texttt{pink\_hair, cat\_ears, maid\_outfit}}
\end{subfigure}\hfill
\begin{subfigure}[b]{0.48\textwidth}
    \centering
    \includegraphics[width=0.88\linewidth]{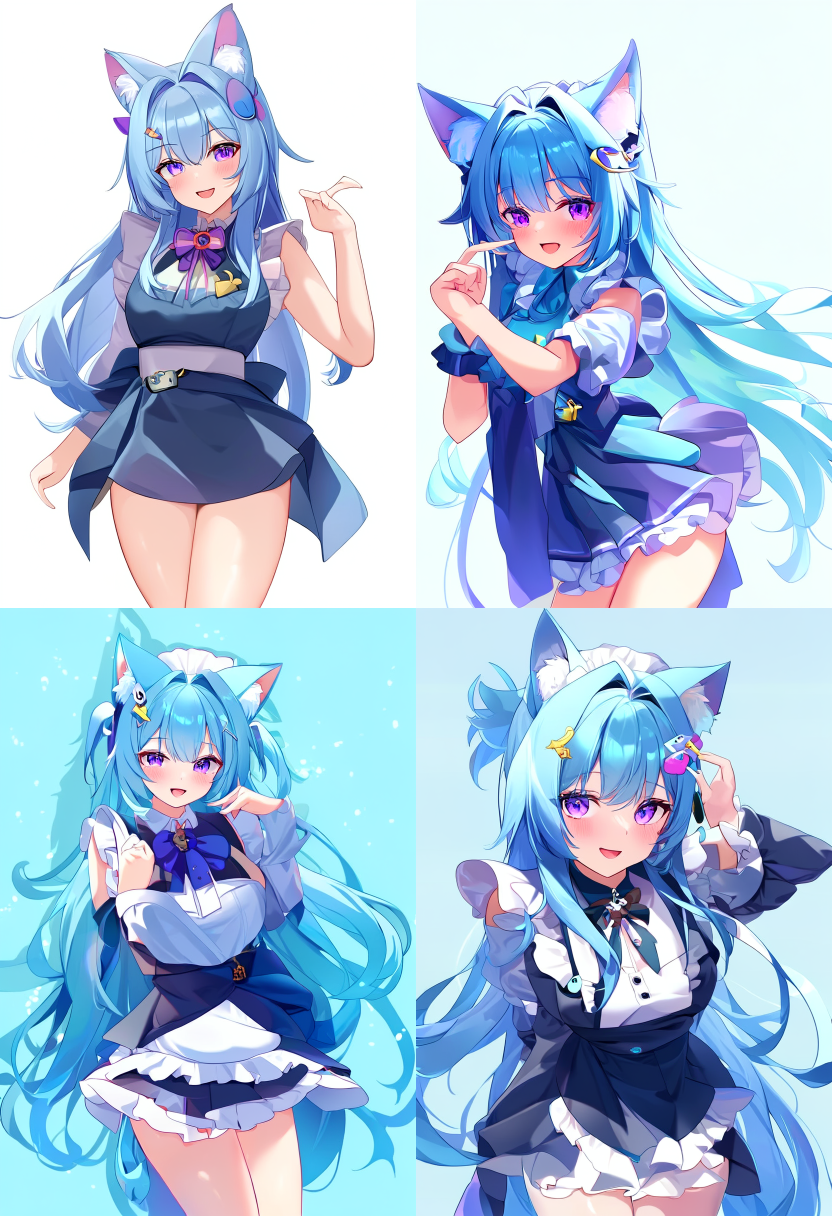}
    \caption{\textbf{Prompt conditioning:} \texttt{blue\_hair, cat\_ears, maid\_outfit}}
\end{subfigure}
\vspace{-4pt}
\caption{\textbf{Strict semantic color disentanglement under identical latent initializations.} Modifying exclusively the hair color prompt token updates the pigment channels without perturbing facial geometry, eye catchlights, clothing ruffles, hand gestures, or background sparkle lighting.}
\label{fig:disentanglement_comparison}
\end{figure*}

To evaluate whether Asuka-FM achieves clean semantic disentanglement, we sample images under identical standard Gaussian prior noise initializations $x_1 \sim \mathcal{N}(\mathbf{0}, \mathbf{I})$ while altering a single descriptive attribute in the prompt context. As demonstrated in Figure~\ref{fig:disentanglement_comparison}, toggling the tag from \texttt{pink\_hair} to \texttt{blue\_hair} selectively alters only the chromatic values of the hair strands. The complex finger pose, ribbon folds, facial morphology, and specular lighting remain invariant down to the pixel level. This confirms that the Dual-Stream DiT architecture effectively prevents cross-attention attribute leakage across distinct token subspaces.

\section{Visual Dynamics of SPRINT: Sparse Attenuation and Dense Recovery}
\label{sec:appendix_sprint_dynamics}

To understand how sequence sparsification affects visual representation learning, we examine the qualitative representations produced during the 75\% sparse phase (token drop ratio $r = 0.75$) versus the final dense refinement phase ($r = 0.00$).

\begin{figure*}[t!]
\centering
\begin{subfigure}[b]{0.48\textwidth}
    \centering
    \includegraphics[width=0.85\linewidth]{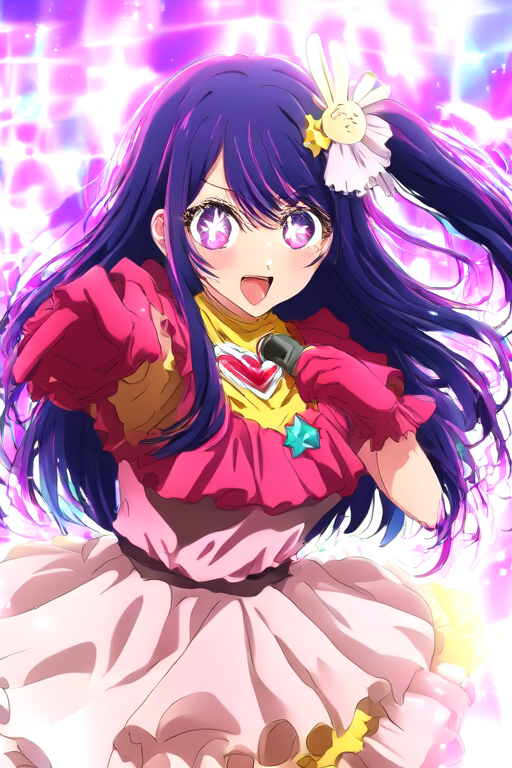}
    \caption{\textbf{SPRINT Sparse Phase ($r = 0.75$):} Macro-composition, character silhouette, and color blocking converge rapidly, but high-frequency lineart, eye highlights, and fringe hair details exhibit transient attenuation.}
\end{subfigure}\hfill
\begin{subfigure}[b]{0.48\textwidth}
    \centering
    \includegraphics[width=0.85\linewidth]{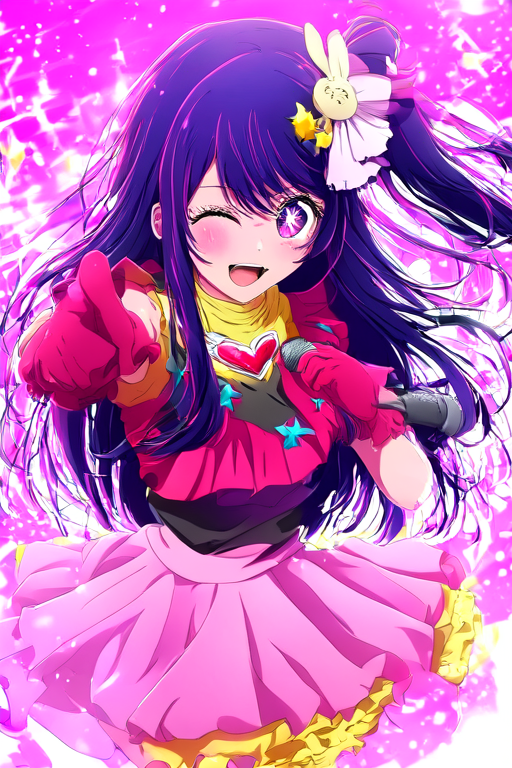}
    \caption{\textbf{Dense Refined Phase ($r = 0.00$):} Full token interaction restores high-frequency lineart, resolves complex eye iris stars, sharpens finger contours, and stabilizes intricate stage lighting.}
\end{subfigure}
\vspace{-4pt}
\caption{\textbf{Visual impact of SPRINT sequence sparsification during training under identical noise seeds.} Sparse pretraining establishes robust macro-semantic layout at $>50\%$ compute savings, after which dense refinement recovers razor-sharp textural boundaries.}
\label{fig:sprint_token_recovery}
\end{figure*}

\paragraph{Structured Group-Wise vs. Uniform Random Token Dropping.}
Uniform random token dropping produces irregular spatial holes that destabilize continuous RoPE phase shifts, as adjacent patch distances vary stochastically across the sequence. For an input canvas of resolution $H_0 \times W_0$ and effective patch size $P = 16$, SPRINT avoids irregular holes by partitioning the $(H_0/P) \times (W_0/P)$ patch grid into non-overlapping $n \times n = 2 \times 2$ windows ($n=2$), deterministically sampling $k=1$ active token per group. This preserves a regular, stride-2 downsampled grid that acts as a coherent, coarser canvas for the sparse deep middle path $g_\theta$.

\paragraph{High-Frequency Attenuation in Sparse Regimes.}
Because $75\%$ of visual tokens are replaced with the learnable mask vector $\mathbf{M} \in \mathbb{R}^D$ inside the middle blocks $g_\theta$, the model relies on the shallow dense encoder features $f_t$ (produced by $f_\theta$) and the linear skip fusion $h_t = W_{\text{fuse}}[f_t \,\|\, g_t^{\text{pad}}] + b_{\text{fuse}}$ (where $g_t^{\text{pad}}$ denotes the mask-padded deep representation, and $W_{\text{fuse}}, b_{\text{fuse}}$ are the fusion projection parameters) to reconstruct fine edges. As seen in Figure~\ref{fig:sprint_token_recovery}(a), sparse checkpoints capture overall pose and vibrant coloring but lack micro-structural crispness. Transitioning to dense fine-tuning (Figure~\ref{fig:sprint_token_recovery}(b)) allows attention weights across all 1,024 patches (at $512^2$) to re-engage, completely recovering intricate lineart.

\begin{figure*}[t!]
\centering
\begin{subfigure}[b]{0.48\textwidth}
    \centering
    \includegraphics[width=\linewidth]{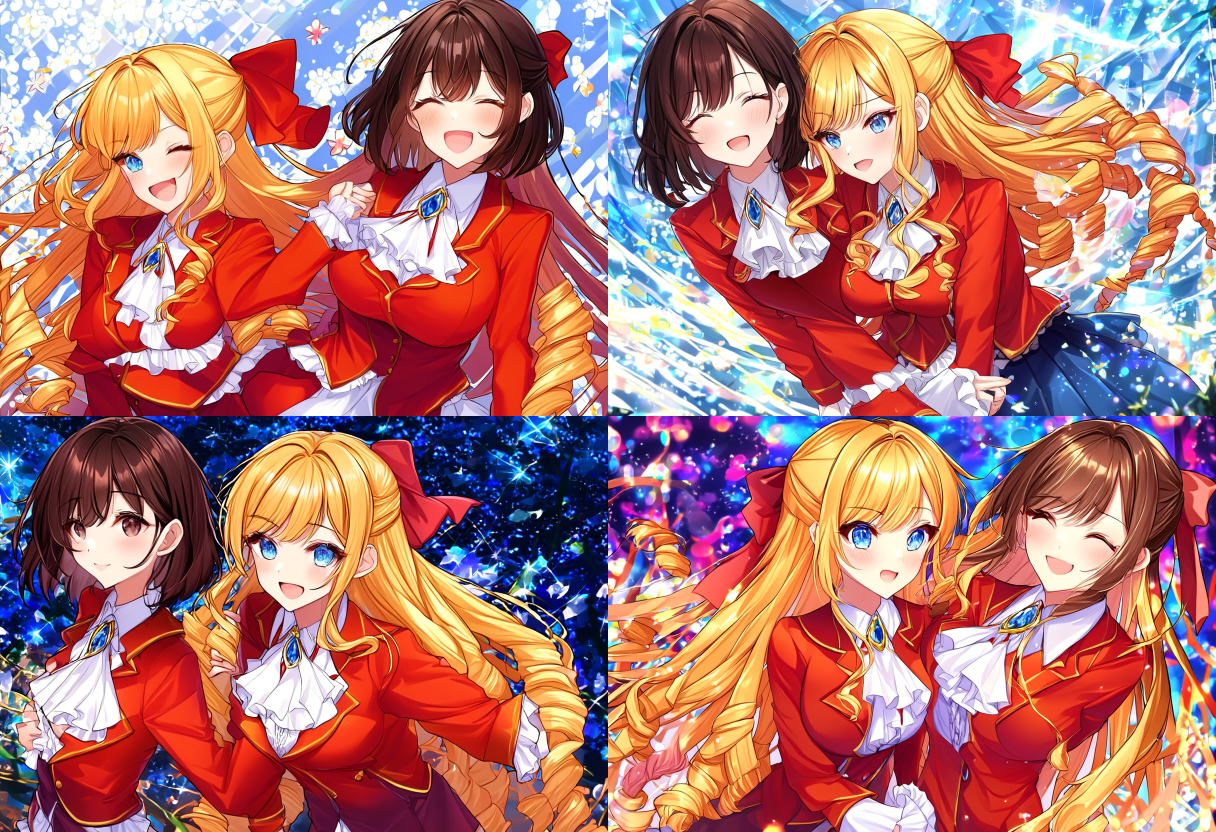}
    \caption{\textbf{1MP SPRINT Sparse ($r = 0.75$, 3 Epochs):} Spatial stride of 32px induces multi-character anatomical fracturing and blurred limb interactions.}
\end{subfigure}\hfill
\begin{subfigure}[b]{0.48\textwidth}
    \centering
    \includegraphics[width=\linewidth]{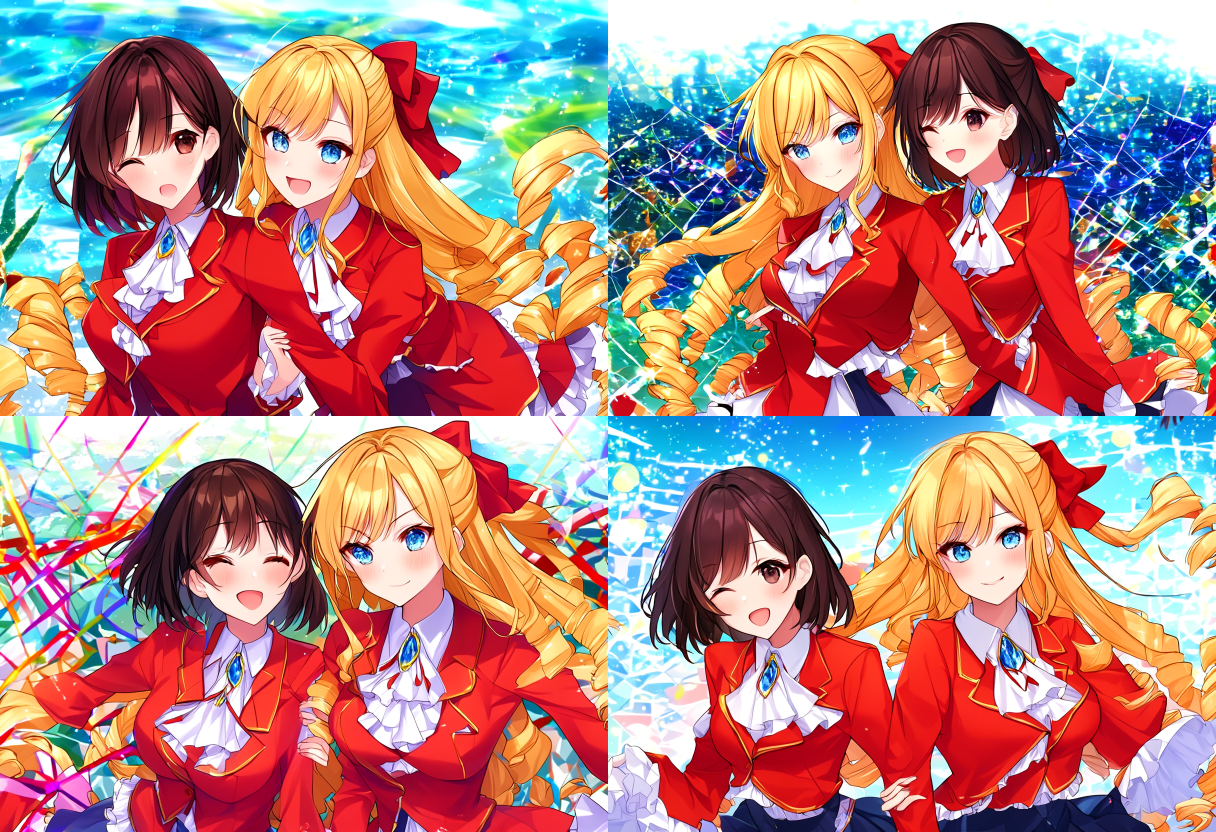}
    \caption{\textbf{1MP Dense Finetuned ($r = 0.00$, 4 Epochs):} Restores complete multi-subject anatomical separation, fine ribbon details, and environmental clarity.}
\end{subfigure}
\vspace{-4pt}
\caption{\textbf{Resolution scaling dynamics at native 1MP ($1024^2$, $N=4096$) across sparse and dense pretraining phases.}}
\label{fig:sprint_1mp_recovery}
\end{figure*}

\paragraph{Multi-Character Scaling at 1MP Resolution.}
At native 1-megapixel resolution ($1024 \times 1024$, total image token count $N = 4096$), $75\%$ token dropping discards 3,072 patches per step. In complex multi-character compositions, this creates spatial blind spots wider than individual faces or hands. As illustrated in Figure~\ref{fig:sprint_1mp_recovery}(a), early sparse epochs on multi-character compositions induced anatomical fractures and color leakage across character boundaries. Executing subsequent dense fine-tuning (Phase~3B in Table~\ref{tab:anime_full_pretraining_progression_appendix}) re-established global semantic coordination (Figure~\ref{fig:sprint_1mp_recovery}(b)), proving that dense refinement is mandatory when scaling sequence sparsification to megapixel regimes.

\section{Curated Anime Dataset and Multi-Stage Pretraining Progression}
\label{sec:appendix_anime_pretraining}

To evaluate domain specialization, extreme aspect-ratio scaling, and continuous coordinate geometry, we established the \textit{curated-danbooru-2026}\footnote{\url{https://huggingface.co/datasets/aipracticecafe/curated-danbooru-2026}} testbed. This dataset comprises $N_{\text{total}} = 337{,}000$ high-density anime-style illustrations sampled from Danbooru2023~\cite{danbooru2023}.

Table~\ref{tab:anime_full_pretraining_progression_appendix} documents the complete computational expenditure, physical throughput, and financial accounting across the three resolution tiers ($256^2 \to 512^2 \to 1024^2$) under the same $7PD$ / $6PD$ accounting as Table~\ref{tab:training_stages} (for $P = 206.12\text{M}$ active parameters and $D$ tokens per sample). Reflecting the bimodal prompt-length distribution of Danbooru tags, we evaluate sequence compute at the representative medium context tier ($152\text{ text} + 1\text{ time} = 153$ context tokens, vs.\ $228$ tokens in the 625K corpus). Consequently, Stages~1 and~2B/2C achieve higher image throughput ($234.0\text{--}334.3\text{ img/s}$ at $256^2$ and $72.0\text{--}78.0\text{ img/s}$ at $512^2$ dense) while sustaining $68.4\%$ cumulative $\text{MFU}_{\text{act}}$ across Stages~1--2. Note that Stage~2A ($70.9\text{ img/s}$, $\text{MFU}_{\text{act}} = 40.0\%$) did not realize the full SPRINT speedup achieved in the 625K pipeline ($108.5\text{ img/s}$) due to network speed during latent streaming from HuggingFace, while Stage~3 ($1024^2$) drops to $14.5\%\text{--}26.2\%$ $\text{MFU}_{\text{act}}$ due to the online VAE bottleneck (Section~\ref{sec:appendix_1mp_scaling}).

\begin{table*}[htbp]
\centering
\small
\caption{\textbf{Pretraining progression and compute metrics of Asuka-FM on the curated anime dataset ($N_{\text{total}} = 337{,}000$) on a single consumer NVIDIA RTX 4090 GPU (24GB).} Formatted symmetrically with Table~\ref{tab:training_stages}. Pretraining executed using an effective batch size of 256 for Stages~1 and~2 (micro-batch 64 with 4 gradient accumulation steps), and effective batch size 128 for Stage~3 (micro-batch 16 with 8 accumulation steps). Throughput, compute, and MFU are derived directly from per-stage epochs and wall-clock hours using the medium context tier ($152\text{ text} + 1\text{ time} = 153$ tokens). Hardware costs evaluated at \$0.33/GPU-hour.}
\label{tab:anime_full_pretraining_progression_appendix}
\resizebox{\textwidth}{!}{%
\begin{tabular}{lcccccccccr}
\toprule
\textbf{Stage / Phase} & \textbf{Res. ($X^2$)} & \textbf{Drop ($r$)} & \textbf{Epochs} & \textbf{Samples} 
& \textbf{Throughput (img/s)} & \textbf{MFU (\%) Theo / Act$^\dagger$} 
& \textbf{Compute (EFLOPs)$^*$} & \textbf{Wall Time} & \textbf{Cost (\$)} \\
\midrule
\multicolumn{10}{l}{\textit{\textbf{Stage 1: Base Semantic Pretraining ($256 \times 256$, Latent Patch Grid $16 \times 16 = 256$ Image Tokens)}}} \\
1A: SPRINT Sparse & $256^2$ & 0.75 & 100 & 33.70M & 334.3 & 69.1\% / 80.6\% & 13.42 & 28.0 h & \$9.24 \\
1B: Dense Refine  & $256^2$ & 0.00 & 50  & 16.85M & 234.0 & 71.7\% / 83.6\% & 9.94  & 20.0 h & \$6.60 \\
1C: WSD Cooldown  & $256^2$ & 0.00 & 25  & 8.43M  & 234.0 & 71.7\% / 83.6\% & 4.97  & 10.0 h & \$3.30 \\
\midrule
\multicolumn{10}{l}{\textit{\textbf{Stage 2: High-Resolution Adaptation ($512 \times 512$, Latent Patch Grid $32 \times 32 = 1,024$ Image Tokens)}}} \\
2A: High-Res Sparse & $512^2$ & 0.75 & 25 & 8.43M & 70.9 & 34.3\% / 40.0\% & 7.84 & 33.0 h & \$10.89 \\
2B: Dense Finetune  & $512^2$ & 0.00 & 10 & 3.37M & 72.0 & 63.4\% / 74.0\% & 5.72 & 13.0 h & \$4.29 \\
2C: WSD Cooldown    & $512^2$ & 0.00 & 5  & 1.69M & 78.0 & 68.7\% / 80.2\% & 2.86 & 6.0 h  & \$1.98 \\
\midrule
\multicolumn{10}{l}{\textit{\textbf{Stage 3: Native Megapixel Scaling ($1024 \times 1024$, Latent Patch Grid $64 \times 64 = 4,096$ Image Tokens)}}} \\
3A: SPRINT Sparse   & $1024^2$ & 0.75 & 3 & 1.01M & 7.80 & 12.4\% / 14.5\% & 3.10 & 36.0 h & \$11.88 \\
3B: Dense Finetune  & $1024^2$ & 0.00 & 4 & 1.35M & 7.07 & 22.5\% / 26.2\% & 8.26 & 53.0 h & \$17.49 \\
\midrule
\multicolumn{3}{l}{\textbf{Cumulative 512$^2$ Subtotal (Stages 1 + 2)}} & \textbf{215} & \textbf{72.46M} & -- & \textbf{58.7\% / 68.4\%} & \textbf{44.77} & \textbf{110.0 h} & \textbf{\$36.30} \\
\multicolumn{3}{l}{\textbf{Cumulative Grand Total (Stages 1 + 2 + 3)}} & \textbf{222} & \textbf{74.81M} & -- & \textbf{40.6\% / 47.4\%} & \textbf{56.13} & \textbf{199.0 h} & \textbf{\$65.67} \\
\bottomrule
\multicolumn{10}{l}{\footnotesize $^\dagger\text{MFU}_{\text{act}}$ evaluates actual tensor ops ($7PD$ per sample, for active parameter count $P=206.12\text{M}$ and tokens per sample $D$) under SwiGLU recomputation; $\text{MFU}_{\text{theo}}$ denotes standard algorithmic utilization ($6PD$).} \\
\multicolumn{10}{l}{\footnotesize $^*1\text{ EFLOP} = 10^{18}\text{ FLOPs} = 10^3\text{ PFLOPs} = 10^{-3}\text{ ZFLOPs}$. Cumulative theoretical algorithmic compute ($6PD$) is $38.37\text{ EFLOPs}$ (Stages 1--2) and $48.11\text{ EFLOPs}$ (Stages 1--3).}
\end{tabular}%
}
\end{table*}

\paragraph{SwinV2 Prompt Upsampling and Dynamic Prefix Formatting.}
Because crowdsourced tags frequently suffer from omissions, every training image is tagged via a \texttt{SwinV2-Transformer} tagger~\cite{liu2022swinv2}, extracting fine-grained descriptors across apparel, hair styling, accessories, and atmospheric lighting. As introduced in Section~\ref{sec:systems_engineering}, tags are assembled into the canonical prefix format:
\begin{displayquote}
\texttt{[character count], [character name], [series], [artist], [rating], [aesthetic tags], [year], [shuffled general tags]}
\end{displayquote}
General tags are shuffled per epoch to eliminate sequence order biases.

\paragraph{Tag-Rarity Loss Weighting.}
To prevent high-frequency tags (\texttt{1girl}, \texttt{solo}) from drowning out gradient signals for rare characters and artist signatures, we apply the inverse tag-frequency weighting of Eq.~\eqref{eq:rarity_loss_weight}. Given the active prompt tag set $\mathcal{T}$ of cardinality $|\mathcal{T}|$, dataset size $N_{\text{total}} = 337{,}000$, and corpus frequency $N_k$ of tag $k \in \mathcal{T}$, we compute the per-sample prompt weight $w_{\text{prompt}}$ and scale the base flow-matching objective $\mathcal{L}_{\text{CFM}}$ to obtain the weighted loss $\mathcal{L}_{\text{weighted}}$:
\begin{equation}
    w_{\text{prompt}} = 1.0 + \frac{1}{|\mathcal{T}|} \sum_{k \in \mathcal{T}} \ln\left( \frac{N_{\text{total}}}{N_k + 1} \right), \quad \mathcal{L}_{\text{weighted}} = w_{\text{prompt}} \cdot \mathcal{L}_{\text{CFM}},
    \label{eq:appendix_rarity_loss_weight}
\end{equation}
where $\mathcal{T}$ represents the prompt tag set, $N_{\text{total}} = 337{,}000$, and $N_k$ is the corpus tag frequency.

\paragraph{Hierarchical Aesthetic Tiering.}
Samples are classified into four quality tiers based on community scores and favorite counts: \texttt{masterpiece}, \texttt{good score}, \texttt{bad score}, and \texttt{worse score}. To preserve distinctive low-visibility styles, curated artist works are capped against falling into lower aesthetic classes.

\section{Native 1MP Scaling and Hardware Bottleneck Profiling}
\label{sec:appendix_1mp_scaling}

While prior evaluations of SPRINT were restricted to ImageNet-1K at $512 \times 512$, we investigated extending Asuka-FM directly to native 1-megapixel ($1024 \times 1024$, total image token count $N=4096$) generation on single GPUs. Pretraining under 75\% SPRINT token dropping (drop ratio $r=0.75$, Phase~3A) followed by dense fine-tuning ($r=0.00$, Phase~3B) revealed severe systems barriers:

\begin{table}[h]
\centering
\small
\caption{\textbf{Execution profiling breakdown of Asuka-FM at native $1024 \times 1024$ resolution} on a single NVIDIA A40 GPU across variable micro-batch sizes (FLUX.2 VAE, $N=4096$ tokens, TorchInductor compilation, bfloat16 precision).}
\label{tab:profiling_1mp_breakdown}
\resizebox{\columnwidth}{!}{%
\begin{tabular}{lcccc}
\toprule
\textbf{Pipeline Component} & \multicolumn{2}{c}{\textbf{Micro-BS 16 (Accum 8)}} & \multicolumn{2}{c}{\textbf{Micro-BS 32 (Accum 4)}} \\
\cmidrule(lr){2-3} \cmidrule(lr){4-5}
& \textbf{Latency (ms)} & \textbf{Share (\%)} & \textbf{Latency (ms)} & \textbf{Share (\%)} \\
\midrule
Data Loading / IO (CPU)       & 0.16 ms   & 0.0\%  & 0.18 ms   & 0.0\% \\
Online Frozen VAE Encoding    & 2,548.10 ms & \textbf{72.6\%} & 4,906.91 ms & \textbf{72.5\%} \\
Text Encoder Forward Pass     & 31.84 ms   & 0.9\%  & 63.23 ms   & 0.9\% \\
Backbone Forward Pass (Sparse)& 229.31 ms  & 6.5\%  & 457.20 ms  & 6.8\% \\
Backbone Backward Pass (Sparse)& 659.65 ms & 18.8\% & 1,299.48 ms & 19.2\% \\
Optimizer Step \& Norm Clip   & 38.49 ms   & 1.1\%  & 37.83 ms   & 0.6\% \\
\midrule
\textbf{Total Iteration Latency}& \textbf{3,507.55 ms} & \textbf{100.0\%} & \textbf{6,764.83 ms} & \textbf{100.0\%} \\
\midrule
Backbone Isolated Throughput  & \multicolumn{2}{c}{46.42 TFLOPS ($\text{MFU}_{\text{act}} = 28.10\%$)} & \multicolumn{2}{c}{46.98 TFLOPS ($\text{MFU}_{\text{act}} = 28.44\%$)} \\
End-to-End System MFU         & \multicolumn{2}{c}{\textbf{3.66\%}} & \multicolumn{2}{c}{\textbf{3.79\%}} \\
\bottomrule
\end{tabular}%
}
\end{table}

\paragraph{The Online VAE Wall at Scale.}
At $1024 \times 1024$, storing uncompressed 32-channel float16 latents for hundreds of thousands of samples requires $>1.2\,\text{TB}$ of high-speed NVMe storage, exceeding consumer workstation capacity. Consequently, latents must be encoded from raw pixels on-the-fly. 

As detailed in Table~\ref{tab:profiling_1mp_breakdown}, profiling on the NVIDIA A40 demonstrates that across both micro-batch regimes, the frozen in-place VAE forward pass consumes an overwhelming \textbf{$72.5\%\text{--}72.6\%$} of total step latency ($2{,}548.10\,\text{ms}$ at micro-batch 16, $4{,}906.91\,\text{ms}$ at micro-batch 32). In contrast, the entire 206M generative backbone forward-backward pass executes in only $888.96\,\text{ms}$ at micro-batch 16 and $1{,}756.68\,\text{ms}$ at micro-batch 32. Consistent with our RTX 4090 profiling at micro-batch 8 (Section~\ref{sec:hardware_profiling}, where the frozen VAE accounts for $62.0\%$ of step latency and runs $2.42\times$ slower than the DiT backbone), the frozen autoencoder on the A40 runs nearly \textbf{$3\times$ slower} than the trainable generative backbone, causing tensor cores to idle for almost three-quarters of execution time and collapsing end-to-end system MFU to \textbf{$3.66\%\text{--}3.79\%$}. This finding reinforces that consumer-scale generative modeling cannot rely on online autoencoding at high resolutions, providing decisive systems justification for our decoupled latent-to-pixel adaptation strategy.

The frozen autoencoder runs nearly \textbf{$3\times$ slower} than the trainable generative backbone, causing tensor cores to idle for almost three-quarters of execution time and collapsing end-to-end system MFU to an unacceptable \textbf{$3.66\%\text{--}3.79\%$}. This finding reinforces that consumer-scale generative modeling cannot rely on online autoencoding at high resolutions, providing decisive systems justification for our decoupled latent-to-pixel adaptation strategy.

\paragraph{Shallow Receptive-Field Starvation in 1MP SPRINT.}
SPRINT extracts dense spatial representations in the initial shallow encoder ($L_{\text{enc}} = 2$ layers) before subsampling. At $256 \times 256$, two self-attention layers with effective patch size $P=16$ span a local receptive field covering $12.5\%$ of the linear canvas span (capturing full facial landmarks). At $1024 \times 1024$, however, that same two-layer receptive field covers only \textbf{$3.1\%$} of the linear canvas span.

Subsampling $75\%$ of tokens ($r = 0.75$) immediately after Layer~2 forces the sparse middle blocks $g_\theta$ to route global semantic composition from ungrounded micro-edges with an effective spatial stride of $32\,\text{px}$. During Phase~3A, this receptive-field starvation induced anatomical fracturing in complex multi-character scenes (Figure~\ref{fig:sprint_1mp_recovery}(a)). While subsequent dense fine-tuning (Phase~3B, Figure~\ref{fig:sprint_1mp_recovery}(b)) restored anatomical coherence, this degradation demonstrates that sequence sparsification at 1MP could require a density curriculum (starting at $r=0.50$) or a deeper shallow encoder ($L_{\text{enc}} \ge 4$).

These empirical findings validate our decision to focus consumer-scale latent-to-pixel adaptation within the $512 \times 512$ regime, where the autoencoder throughput wall is permanently bypassed.

\section{Curated Anime Qualitative Generation Gallery}
\label{sec:qualitative_gallery}

\begin{figure*}[htbp]
\centering
\includegraphics[width=\textwidth, height=0.86\textheight, keepaspectratio]{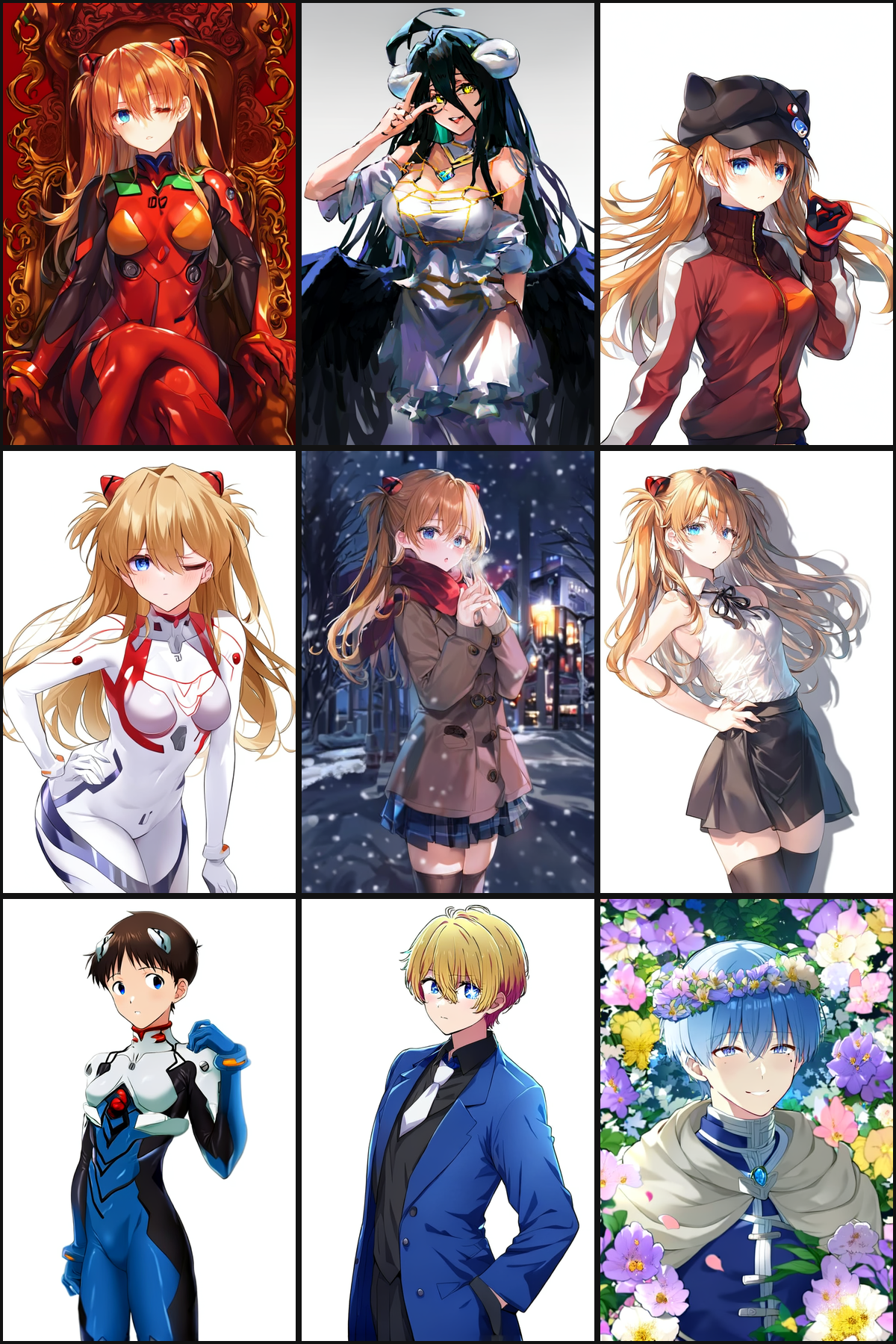}
\caption{\textbf{High-quality zero-shot samples from the anime-specialized Asuka-FM model.} Our 206M dual-stream backbone demonstrates sharp lineart, intricate anatomical rendering, and character-concept binding pretrained from scratch on a single consumer RTX 4090 despite a compact open dataset.}
\label{fig:anime_teaser_showcase}
\end{figure*}

In Figure~\ref{fig:anime_teaser_showcase}, we provide high-resolution zero-shot samples synthesized by our domain-specialized 206M Asuka-FM model pretrained from scratch on the curated 337K Danbooru corpus on a single consumer RTX 4090 GPU, demonstrating sharp lineart, anatomical rendering, and character-concept binding.

\end{document}